\newcommand{\final}{1}
\newcommand{\arxiv}{1}
\documentclass[sigconf,nonacm]{acmart} % for final publication on arxiv
\acmSubmissionID{247}

\usepackage[ruled]{algorithm2e} % For algorithms

\SetAlFnt{\small}
\SetAlCapFnt{\small}
\SetAlCapNameFnt{\small}
\SetAlCapHSkip{0pt}

\usepackage{amsthm}

\DeclareMathOperator*{\argmin}{arg\,min}
\usepackage{amssymb}
\usepackage{pifont}
\usepackage{makecell}
\usepackage{textcomp}
\usepackage{wrapfig}
\usepackage{color}
\usepackage{xspace}
\usepackage{enumitem}
\usepackage{multirow}
\usepackage{soul}
\usepackage{ifthen}
\usepackage{microtype}

\usepackage{subfig}
\usepackage{hyperref}
\newcommand{\onethirdfigurewidth}{0.32}
\newcommand{\onefourthfigurewidth}{0.24}

\newcommand{\oneninthfigurewidth}{0.1111111111111}
\newcommand{\onetenthfigurewidth}{0.098}

\newcommand{\revision}[1]{{\color{orange} #1}}

\newcommand{\yl}[1]{{\color{blue}#1}}

\ifthenelse{\equal{\final}{1}}
{
\renewcommand{\revision}[1]{#1}
\renewcommand{\yl}[1]{#1}

}
{}

\newcommand{\sysname}{VRMM\xspace} % Volumetric Relightable Morphable Model

\newcommand{\light}{l}
\newcommand{\direction}{\mathbf{d}}

\newcommand{\etal}{\emph{et al.}}
\newcommand{\ie}{\emph{i.e.}}

\newcommand{\loss}{\mathcal{L}}
\newcommand{\totalloss}{\loss_{total}}
\newcommand{\dataloss}{\loss_{img}}

\newcommand{\regularizationloss}{\loss_{reg}}
\newcommand{\expressionloss}{\loss_{exp}}

\newcommand{\encoder}{\mathcal{E}}

\newcommand{\expressionencoder}{\encoder_{e}}
\newcommand{\transformencoder}{\encoder_{T}}
\newcommand{\decoder}{\mathcal{D}}
\newcommand{\meshdecoder}{\decoder_{mesh}}
\newcommand{\transformdecoder}{\decoder_{T}}
\newcommand{\opacitydecoder}{\decoder_{\alpha}}
\newcommand{\colordecoder}{\decoder_{rgb}}
\newcommand{\mvpdecoder}{\decoder_{MVP}}
\newcommand{\identitydecoder}{\decoder_{id}}

\newcommand{\latentcode}{z}
\newcommand{\expressioncode}{\latentcode_{e}}
\newcommand{\identitycode}{\latentcode_{id}}

\newcommand{\vertex}{\mathbf{v}}
\newcommand{\pixel}{p}
\newcommand{\blendweight}{\mathbf{w}}

\newcommand{\featuremap}{\mathcal{F}}

\newcommand{\appearancefeaturemap}{\featuremap_\mathrm{{appe}}}
\newcommand{\alphafeaturemap}{\featuremap_\mathrm{{alpha}}}

\newcommand{\voxelcolor}{V_{rgb}}
\newcommand{\voxelopacity}{V_{\alpha}}

\newcommand{\renderedimage}{I_{rgb}}

\newcommand{\cameraparams}{\phi}
\newcommand{\projection}{\Pi}

\copyrightyear{2024}
\acmYear{2024}
\setcopyright{acmlicensed}\acmConference[SIGGRAPH Conference Papers '24]{Special Interest Group on Computer Graphics and Interactive Techniques Conference Conference Papers '24}{July 27-August 1, 2024}{Denver, CO, USA}
\acmBooktitle{Special Interest Group on Computer Graphics and Interactive Techniques Conference Conference Papers '24 (SIGGRAPH Conference Papers '24), July 27-August 1, 2024, Denver, CO, USA}
\acmDOI{10.1145/3641519.3657406}
\acmISBN{979-8-4007-0525-0/24/07}

\begin{document}

%%%%%%%%% TITLE
%\title{Towards Practical Creation of High-Fidelity Relightable Avatars}
%\title{A Practical Approach to Capturing and Building High-Fidelity Relightable Avatars}
\title{VRMM: A Volumetric Relightable Morphable Head Model}
%  for Animatable Avatar Reconstruction

\ifthenelse{\equal{\arxiv}{1}}
{
\author{Haotian Yang$^{1}$\hspace{0.08in} Mingwu Zheng$^{1}$\hspace{0.08in} Chongyang Ma$^{1}$\hspace{0.08in} Yu-Kun Lai$^2$\hspace{0.08in} Pengfei Wan$^1$\hspace{0.08in} Haibin Huang$^{1*}$ \vspace{4pt}\\
\institution{$^1$Kuaishou Technology \hspace{0.3in} $^2$Cardiff University}
}
\renewcommand{\shortauthors}{Yang et al.}
\renewcommand{\authors}{Haotian Yang, Mingwu Zheng, Chongyang Ma, Yu-Kun Lai, Pengfei Wan, and Chongyang Ma}
}
{
\author{Haotian Yang}
\affiliation{
\institution{Kuaishou Technology}
\country{China}
}
\email{yanght321@gmail.com}

\author{Mingwu Zheng}
\affiliation{
\institution{Kuaishou Technology}
\country{China}
}
\email{zhengmingwu@kuaishou.com}

\author{Chongyang Ma}
\affiliation{
\institution{Kuaishou Technology}
\country{China}
}
\email{chongyangm@gmail.com}

\author{Yu-Kun Lai}
\affiliation{
\institution{Cardiff University}
\country{United Kingdom}
}
\email{laiy4@cardiff.ac.uk}

\author{Pengfei Wan}
\affiliation{
\institution{Kuaishou Technology}
\country{China}
}
\email{wanpengfei@kuaishou.com}

\author{Haibin Huang}
\authornote{Corresponding author.}
\affiliation{
\institution{Kuaishou Technology}
\country{China}
}
\email{jackiehuanghaibin@gmail.com}

}

%\title{High-Fidelity Avatar via Self-Organized Relightable Mixture of Volumetric Primitives}
% \title{Relightable Avatar via Tracking-Free Mixture of Volumetric Primitives}

\begin{teaserfigure}
\centering
    \includegraphics[width=\linewidth]{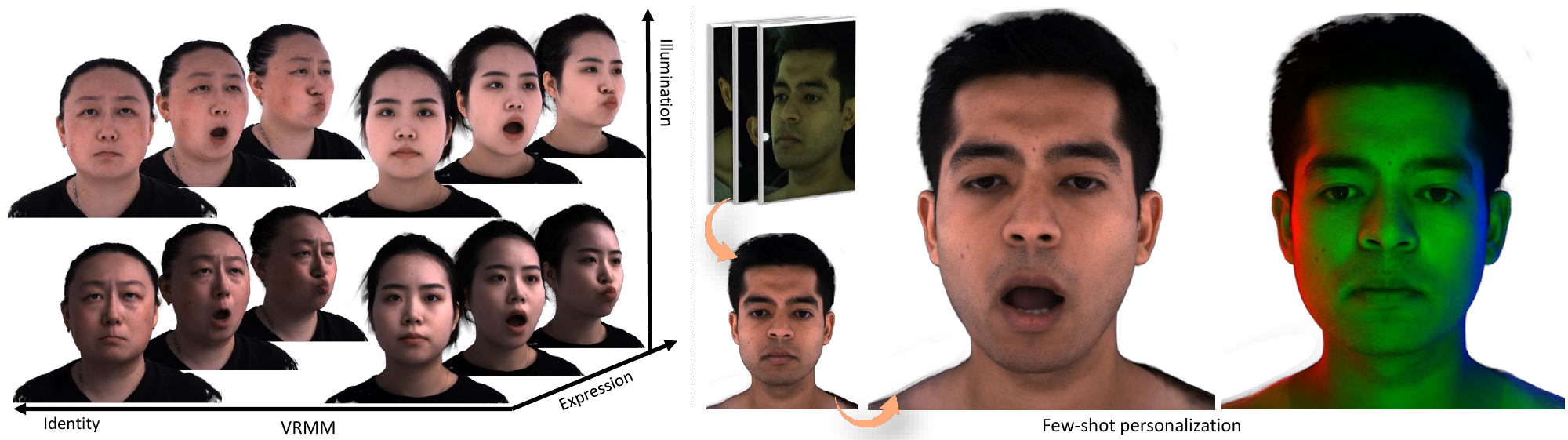}
    \captionof{figure}{We present \sysname, a novel volumetric head prior with fully disentangled low-dimensional parametric space for identity, expression, and illumination. Trained on dynamic expressions of hundreds of people captured in a LightStage with controllable illumination, our \sysname enables high-quality animatable and relightable avatar reconstruction from few-shot observations.}
\label{fig:teaser}
\end{teaserfigure}

% \renewcommand{\shortauthors}{H. Yang, M. Zheng, W. Feng, H. Huang, Y.-K. Lai, P. Wan, Z. Wang, and C. Ma}
% \renewcommand{\authors}{Haotian Yang, Mingwu Zheng, Wanquan Feng, Haibin Huang, Yu-Kun Lai, Pengfei Wan, Zhongyuan Wang, and Chongyang Ma}

%%%%%%%%% ABSTRACT
\begin{abstract}
\ifthenelse{\equal{\arxiv}{1}}
{
\renewcommand{\thefootnote}{}
\footnotetext{* Corresponding author.}
}
{}

In this paper, we introduce the Volumetric Relightable Morphable Model (\sysname), a novel volumetric and parametric facial prior for 3D face modeling. While recent volumetric prior models offer improvements over traditional methods like 3D Morphable Models (3DMMs), they face challenges in model learning and personalized reconstructions. Our \sysname overcomes these by employing a novel training framework that efficiently disentangles and encodes latent spaces of identity, expression, and lighting into low-dimensional representations. This framework, designed with self-supervised learning, significantly reduces the \yl{constraints} for training data, making it more feasible in practice. The learned \sysname offers relighting capabilities and encompasses a comprehensive range of expressions. We demonstrate the versatility and effectiveness of \sysname through various applications like avatar generation, facial reconstruction, and animation. Additionally, we address the common issue of overfitting in generative volumetric models with a novel prior-preserving personalization framework based on \sysname. Such \yl{an} approach enables high-quality 3D face reconstruction from even a single portrait input.  Our experiments showcase the potential of \sysname to significantly enhance the field of 3D face modeling.

\end{abstract}



% 30-word summary: We propose a novel framework, Tracking-free Relightable Avatar (TRAvatar), for capturing and reconstructing high-fidelity 3D avatars. Compared to existing methods, TRAvatar works in a more practical and efficient setting.


\maketitle

%%%%%%%%% BODY TEXT
\section{Introduction}

%The quest for realistic 3D facial representation has been a cornerstone of both academic research and practical applications, ranging from digital entertainment to telepresence and biometric identification. Among the various approaches for 3D facial modeling, 3D Morphable Models (3DMMs)~\cite{blanz1999morphable} have set a benchmark due to their ability to encode faces in a parametric form. This parametric nature allows for the manipulation of face identity, expressions and other attributes with relative ease, making 3DMMs a powerful tool in the field.

%Historically, mesh-based 3DMMs have been the standard. However, they often struggle to achieve a truly lifelike fidelity, particularly in the subtleties of complex head components such as hair and the interior of the mouth. To address these shortcomings, research has shifted toward \yl{volumetric}
%-based 
%models. These models offer a better representation of the 3D space occupied by facial structures and promise improved realism in facial modeling. 

%Despite the advancements offered by volumetric approaches, there remains a significant gap in their capability. Due to the lack of both data and training method, existing data-driven volumetric priors are yet to adequately model dynamic facial expressions or to simulate the effects of variable lighting conditions on the face, which is critical for relightable appearance. Such limitations have impeded the broader application of these models in scenarios where expressive and responsive faces are paramount.

In this study, we explore 3D face modeling from a representation learning perspective. The pursuit of realistic 3D facial representations is pivotal in both academic research and practical applications, encompassing digital entertainment, telepresence, and biometric identification. As a critical task in computer graphics and computer vision, this area has garnered sustained attention. Early works like 3D Morphable Models (3DMMs)~\cite{blanz1999morphable, vlasic2005face, yang2020facescape, cao2013facewarehouse, jiang2019disentangled, li2017learning, booth20163d, paysan20093d} have set a benchmark due to their ability to encode faces in a parametric form. Their parametric nature allows for the manipulation of face identity, expressions and other attributes with relative ease, making 3DMMs a powerful tool in the field. However, they often struggle to achieve a truly lifelike fidelity, particularly in the subtleties of complex head components such as hair and the interior of the mouth. To overcome these limitations, research has shifted towards volumetric models~\cite{cao2022authentic, zhuang2022mofanerf, buhler2023preface, wang2022morf, hong2022headnerf}, which offer a more comprehensive representation of facial structures and promise enhanced realism.

%To address these shortcomings, research has shifted toward volumetric models~\cite{cao2022authentic, zhuang2022mofanerf, buhler2023preface, wang2022morf, hong2022headnerf}, which offer a better representation of the 3D space occupied by facial structures and promise improved realism in facial modeling.    

%In this study, we explore 3D face modeling from a representation learning perspective. The pursuit of realistic 3D facial representations is pivotal in both academic research and practical applications, encompassing digital entertainment, telepresence, and biometric identification. As a critical task in computer graphics and computer vision, this area has garnered sustained attention. Early efforts, such as 3D Morphable Models (3DMMs)~\cite{blanz1999morphable, vlasic2005face, yang2020facescape, cao2013facewarehouse, jiang2019disentangled, li2017learning, booth20163d, paysan20093d}, have established benchmarks with their parametric form, enabling manipulation of face identity, expressions, and other attributes. However, they often fall short in achieving lifelike fidelity, particularly in complex areas like hair and mouth interiors.

Despite the advancements offered by volumetric models, there remains a significant gap in their capability. Our main observations are two-fold: first of all, existing data-driven volumetric models are not able to adequately model dynamic facial expressions or to simulate the effects of variable lighting conditions on the face. 
%Due to the current modeling design of most volumetric prior models, huge amounts of data with various expressions, lighting conditions as well as face identities would be required to train and improve volumetric models, which is tedious and not practical. 
Existing animatable volumetric prior models either adopt supervised learning with decoupled input data of discrete identities and expressions~\cite{zhuang2022mofanerf, hong2022headnerf}, or require expensive and brittle preprocessing steps~\cite{cao2022authentic}, which is tedious and not practical to scale up. These drawbacks also limit the further %the 
query of a continuous relightable model.
Subsequently, there will be common overfitting issues when using generative models for downstream reconstruction tasks~\cite{tewari2020stylerig, tov2021designing, zhu2020improved, abdal2019image2stylegan}, which is also observed in volumetric prior personalization~\cite{buhler2023preface}. This drawback is that due to the limited \yl{training} data, the inversion of input facial data into the volumetric modeling space is often problematic and leading to reconstruction below satisfactory, which also deteriorates the editability, i.e., relightable and animatable \yl{properties} learned in the prior model. 

To bridge this gap, we propose \sysname, a novel volumetric and parametric 3D face prior.  \sysname is built upon volumetric primitives linked to the UV space of a base mesh~\cite{lombardi2021mixture}, and adapt a physically-inspired appearance decoder \cite{yang2023towards} for relighting . It uses multi-identity mapping and an expression encoder to handle various expressions and lighting conditions. By explicitly encoding identity, expression and lighting into low-dimensional representations,  \sysname learns to disentangle the associated latent spaces and can be trained in a self-supervised manner. Our design of \sysname effectively \yl{reduces the constraints} of training data required by previous volumetric models and \yl{enables} training with more \yl{flexible data collections}. In \yl{practice}, the training of our \sysname is based on a dataset comprising high-quality multi-view image sequences of fewer than 300 individuals. These individuals were captured exhibiting dynamic expressions within a LightStage under controllable lighting conditions.

We further adapt our \sysname model for various fitting and reconstruction tasks. As discussed above, fine-tuning is necessary for volumetric prior models to overcome the inversion issue~\cite{buhler2023preface, wang2022morf, cao2022authentic} at the cost of over-fitting. Hence we propose a prior-preserving framework for model fitting. Specifically, we use an identity related regularization term to balance the \sysname\yl{'s} generation capacity and fitting accuracy. Our %deliberately 
\yl{specifically}
designed framework allows high-quality avatars that \yl{are} animatable and relightable to be created from few-shot captures, significantly \yl{outperforming} existing \yl{baselines}.

%\sysname is  designed to capture a full range of facial expressions and to support relighting, thereby addressing the two major limitations of existing volumetric models. To be more specific, \sysname explicitly encodes identity, expression and lighting into low-dimensional representations and learns to disentangle the associated latent spaces. Moreover, \sysname can be trained in a self-supervised manner, \textbf{\color{red}{DETAILS}}

%The relightable feature of our model unlocks the potential for realistic face modeling under various lighting conditions, an aspect pivotal in creating lifelike digital humans.

 %Through this dataset, Furthermore, we will highlight a wide range of applications for our model, including avatar generation, facial reconstruction and animation, demonstrating the \sysname's potential to boost the field of 3D face modeling.

In summary, our contributions are:
\begin{itemize}[leftmargin=*]
\item We present \sysname, the first 3D volumetric facial prior that is both \yl{continuously} relightable and encompasses full range of expressions to the best of our knowledge. 

\item We propose a novel training framework to learn the disentangled parametric space of expression, identity and lighting for \sysname from dynamic multi-view image sequences captured under controllable light conditions.

\item We propose a novel personalization method that \yl{is} elaborately designed to keep the animatable and relightable \yl{properties} provided by the prior, which enables high-fidelity avatar reconstruction from several or even one image.

\item Extensive experiments demonstrate that \sysname can be used in various applications and outperforms previous methods.

\end{itemize}
\section{Related Work}
\label{sec:related}

In this section, we review closely related parametric 3D face models based on both traditional mesh-based representation and volumetric representation. We also discuss related methods of neural avatar reconstruction.

\paragraph{Parametric head models}
Accurate modeling of 3D head geometry and appearance remains a significant challenge in both computer graphics and computer vision. Among various methodologies, 3D Morphable Models (3DMMs) emerged as a pioneering and systematic approach. Originally introduced by Blanz and Vetter \shortcite{blanz1999morphable} and later evolved into multi-linear models \cite{vlasic2005face}, 3DMMs have been fundamental in modeling facial meshes and textures. They serve as a universal facial prior in diverse applications like face reconstruction and tracking from single-view images \cite{zhu2017face, thies2016face2face, dou2017end}. However, 3DMMs are limited by their linear and mesh-based nature, restricting the scope of shape and appearance modeling.

Recent advances in deep learning have sought to address these limitations, employing nonlinear methods for more refined facial modeling \cite{bagautdinov2018modeling, tran2018nonlinear, tran2019learning, ploumpis2020towards, zhang2022video, zheng2022imface}. Despite these improvements, surface-based representations, including both meshes and implicit functions, still struggle to capture the complete complexity of human head components such as teeth, facial hair, eyes, and so on.

A significant shift in this domain is evident with the advent of neural volumetric rendering, particularly NeRF \yl{(Neural Radiance Field)}-based techniques \cite{mildenhall2020nerf, muller2022instant, kerbl2023gaussian}. These methods holistically represent the 3D objects, achieving both photorealism and 3D consistency from multi-view images. Early attempts in volumetric head modeling primarily focused on the geometry and appearance of a single scene or identity \cite{lombardi2019neural, ma2021pixel, yang2023towards, lombardi2021mixture}. More recent work extends to multi-identity scenarios through generative modeling \cite{wang2022morf, hong2022headnerf, zhuang2022mofanerf, buhler2023preface, cao2022authentic}. For instance, MoRF \cite{wang2022morf} employs an auto-decoder framework \cite{park2019deepsdf} to learn a conditional NeRF in a \yl{polarization}-based studio setup, separating the diffuse and specular shading components. However, its applications are confined to studio environments and lack generalization to in-the-wild scenes. Preface \cite{buhler2023preface} expands on this, allowing high-resolution rendering in more casual settings. Nonetheless, both MoRF and Preface are limited to static head models, missing out on dynamic and relightable features which are crucial for real-world applications. MoFaNeRF \cite{zhuang2022mofanerf} and Cao~\etal~\shortcite{cao2022authentic} integrate dynamic expressions, enabling the creation of animatable avatars. HeadNeRF \cite{hong2022headnerf} also introduces basic relighting capabilities but is constrained to certain lighting conditions. Currently, no previous method achieves \yl{full} creation of expressive, dynamically relightable heads under various physically accurate lighting conditions. For a more comprehensive comparison, please refer to Tab.~\ref{tab:cmp_ability}. To our knowledge, VRMM is the first model to achieve this, presenting a significant advancement in the field of volumetric and relightable morphable head models.

\begin{table}[]
\centering
\caption{Comparison of different methods for volumetric head models across four aspects. \revision{VRMM is the only model that enables a physically relightable and animatable model with real-time rendering capabilities. Single-stage reconstruction indicates that the morphable model does not require reconstructed meshes or other 3DMMs when fitting new data.} (* Zhuang~\etal~\shortcite{zhuang2022mofanerf} and Hong~\etal~\shortcite{hong2022headnerf} only support limited expressions. Hong~\etal~\shortcite{hong2022headnerf} is limited to relighting under preset lighting conditions.)}
\label{tab:cmp_ability}
\resizebox{\columnwidth}{!}{%
    \begin{tabular}{l|cccc} % Adjust the widths as needed
        \hline
        {Method} & \thead{Animatable} & \thead{Relightable} & \thead{Single-stage \\ reconstruction} & \thead{Real-time \\ rendering} \\ \hline
        Zhuang~\etal~\shortcite{zhuang2022mofanerf} & \checkmark* & \ding{55} & \checkmark & \ding{55} \\ 
        Cao~\etal~\shortcite{cao2022authentic} & \checkmark & \ding{55} & \ding{55} & \checkmark\\ 
        Hong~\etal~\shortcite{hong2022headnerf} & \checkmark* & \checkmark* & \ding{55} & \checkmark \\ 
        Wang~\etal~\shortcite{wang2022morf} & \ding{55} & \ding{55} & \checkmark & \ding{55} \\ 
        Buhler~\etal~\shortcite{buhler2023preface} & \ding{55} & \ding{55} & \checkmark & \ding{55} \\ 
        VRMM (Ours) & \checkmark & \checkmark & \checkmark & \checkmark \\ \hline
    \end{tabular}
}
\end{table}

\paragraph{Avatar reconstruction}
The field of 3D face reconstruction and performance capture has seen extensive research efforts over decades, leading to the development of sophisticated 3D scanning systems. These systems, focusing on static geometry reconstruction \cite{beeler2010high, ghosh2011multiview} and dynamic performance capture \cite{beeler2011high, bradley2010high, huang2011leveraging, collet2015high, dou2017motion2fusion, guo2019relightables}, predominantly employ multi-view stereo (MVS) and structured light techniques for point cloud acquisition. Subsequent steps involve estimating deforming geometry to maintain temporal mesh consistency. However, such tracking process often requires labor-intensive MVS reconstruction for numerous frames and complex optical-flow optimization, while real-time face tracking algorithms still fall short in accuracy.

\begin{figure*}%[t]
    \centering
    \includegraphics[width=\linewidth]{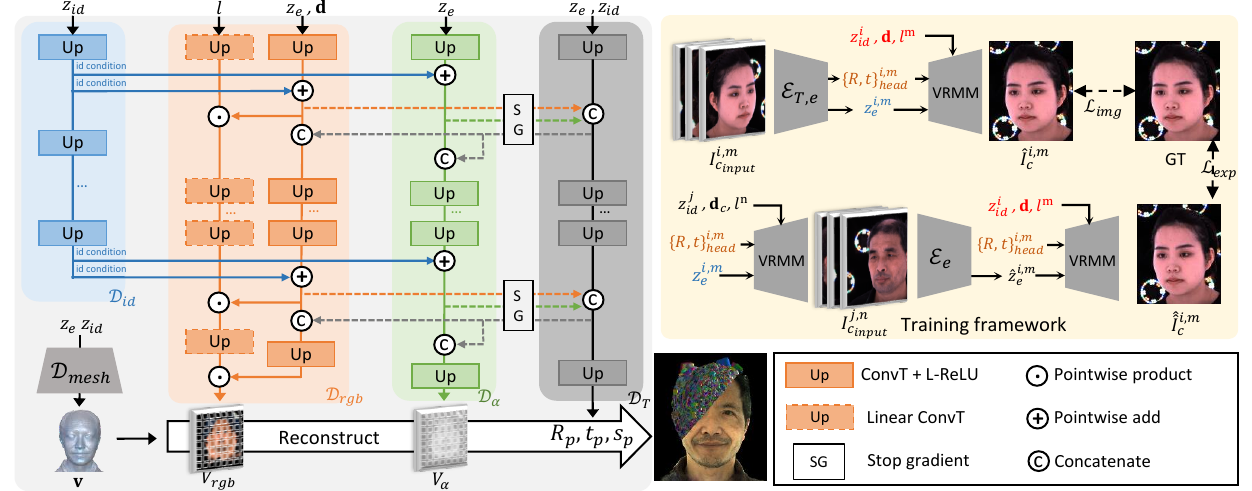}
    % \caption{The pipeline of \sysname. Network Architecture (left): \sysname takes identity code $\identitycode$, expression code $\expressioncode$, view direction $\direction$ and environment light $\light$ as input. The model output, including base mesh and volumetric primitives, can then be generated by corresponding decoders and rendered into an image in real time. Especially, the transformation decoder $\transformdecoder$, opacity decoder $\opacitydecoder$ and the non-linear branch of the relightable appearance decoder $\colordecoder$ are connected by detach-concatenation between blocks, which we found is critical to obtain a stable result. Training Framework (right): We follow TRavatar~\cite{yang2023towards} and DeepSDF~\cite{park2019deepsdf} to jointly train expression encoder $\expressionencoder$, transformation encoder $\transformencoder$, per-person identity codes $\identitycode$ and the decoders. Moreover, we introduce a novel cycle-consistency loss $\expressionloss$ to further align the semantic meaning for the expression codes.
    % }
    \caption{The \sysname pipeline. Network architecture (left): \sysname accepts inputs of identity code $\identitycode$, expression code $\expressioncode$, view direction $\direction$, and environmental light $\light$. The output, comprising a base mesh and volumetric primitives, is generated by respective decoders and rendered into an image in real-time. Notably, the transformation decoder $\transformdecoder$, opacity decoder $\opacitydecoder$, and the non-linear branch of the relightable appearance decoder $\colordecoder$ are interconnected through a detach-concatenation process between blocks, a key factor we found for achieving stable results. Training Framework (right): Our framework jointly trains the expression encoder $\expressionencoder$, transformation encoder $\transformencoder$, per-person identity codes $\identitycode$, and the decoders in \sysname. Additionally, we incorporate a novel expression consistency loss $\expressionloss$ to enhance the semantic alignment of expression codes.}
%%%YKL Some symbols can be better defined. Element wise multiplication symbol should be consistent with the paper.
\label{fig:method}
\end{figure*}

Another vital component for creating realistic and relightable avatars is the estimation of light interaction with the subject, particularly the reflectance properties. Traditional methods typically model this interaction using bidirectional reflectance distribution functions (BRDF) \cite{schlick1994inexpensive}, determined by observing appearance changes under various lighting conditions. Active lighting approaches, such as those using LightStage setups \cite{debevec2000acquiring}, involve data collection with complex arrangements like one-light-at-a-time (OLAT) capture or polarized and color gradient illuminations \cite{ma2007rapid, ghosh2011multiview, zhang2022video, fyffe2015single, guo2019relightables}. Conversely, passive capture methods \revision{~\cite{riviere2020single, zheng2023neuface, li2022eyenerf}} significantly reduce the need for elaborate setups.

Despite these advancements, these methods still require high hardware costs and considerable effort for data acquisition. The advent of deep learning and large-scale datasets has enabled the estimation of geometry and reflectance properties from single-view input\revision{~\cite{lattas2021avatarme++, lattas2020avatarme, li2020dynamic, li2020learning, Papantoniou2023Relightify, yamaguchi2018high, caselles2023sira}}. Nevertheless, these methods typically focus only on skin regions, as the complexity of hair and eye structures makes single-view inverse rendering difficult. Moreover, most reconstructed results are static and not animatable. \revision{Recent endeavors have focused on creating dynamic avatars from monocular videos \cite{zielonka2022instant, gao2022reconstructing, chen2022relighting4d, zheng2023pointavatar} or RGB-D inputs \cite{cao2022authentic}. However, these methods typically fall short in providing relightable attributes, or their quality is constrained by the training video's limitations, due to the lack of a powerful prior model.} The challenge remains to photorealistically model complete human heads, considering all the highly complex and diverse compositions of material, geometry, and expression.

To our best knowledge, our \sysname is the first capable of reconstructing high-quality animatable and relightable volumetric avatars from few-shot captures, representing a significant leap in the field of 3D avatar reconstruction.

\section{Method}
\label{sec:method}

In this section, we first introduce the training framework of our volumetric relightable morphable model using high-quality data captured in the studio, as shown in Figure~\ref{fig:method}. Then we present a specially designed pipeline to fit the learned \sysname model to in the studio or in-the-wild captures for consumer-grade authentic 3D avatar reconstruction.

\subsection{Preliminaries}
Our \sysname is built upon the representation of Mixture of Volumetric Primitives (MVP)~\cite{lombardi2021mixture}, which represents the scene as volumetric primitives attached to the UV space of a base mesh. We adopt the physically-inspired appearance decoder~\cite{yang2023towards} to support relighting, where the lighting condition $\light$ is represented as the incoming light field of $N_l$ densely sampled directions on the sphere. Specifically, given the expression code $\expressioncode$, the view direction $\direction$, and the lighting condition $\light$, a series of decoders $\mvpdecoder$ predict the position and color of  $N_{prim}$ volumetric primitives for rendering:
\begin{equation}
\begin{aligned}
\{\vertex, R_{p}, t_{p}, s_{p}, \voxelopacity, \voxelcolor\} = \mvpdecoder (\expressioncode, \direction, \light),
\end{aligned}
\end{equation}
where $\vertex$ is the position of the vertices of the base mesh, $\{R_{p}, t_{p}, s_{p}\}$ are the rotation, translation and non-uniform \yl{scaling} of $N_{prim}$ primitives relative to the base mesh, $\voxelopacity$ and $\voxelcolor$ are the opacity and color of each voxel in the primitives, respectively. The color of the \yl{rendered} image $\renderedimage$ at pixel $\pixel$ can be obtained by integrating the radiance of the voxels along the direction $\mathbf{d}_\pixel$ of the ray starting from the position $\mathbf{o}_\pixel$ of pixel in 3D space:
\begin{equation}
\begin{aligned}
\renderedimage(\pixel) = \int_{t_{\min}}^{t_{\max}}\voxelcolor(\mathbf{o}_\pixel + t \mathbf{d}_\pixel)\frac{dT(\pixel, t)}{dt},
\end{aligned}
\end{equation}
\begin{equation}
\begin{aligned}
T(\pixel, t)=\min\Big(1, \int_{t_{\min}}^{t}\voxelopacity\big(\mathbf{o}_\pixel + t \mathbf{d}_\pixel\big)\Big),
\end{aligned}
\end{equation}
where $t_{\min}$ and $t_{\max}$ are the predetermined near and far range of the camera plane. 

The volumetric primitives in MVP have consistent structure in the UV space of the base mesh, which improves the quality when extending to \yl{multiple identities} as it is easier for the network to learn the shared features across identities \yl{compared} to other alternatives based on the NeRF representation~\cite{hong2022headnerf, wang2022morf, zhuang2022mofanerf, buhler2023preface}.

% Comparing to other representations, TRAvatar is trained with topology consistent tracking end-to-end from dynamic image sequences, which avoids the brittle preprocessing step of surface-tracking and enables scalable training, making it suitable for building our morphable model from more than 400K frames of 250 different identities.
% \chongyang{It is too early to mention our dataset here.}

\subsection{\sysname}

We start by extending the existing person-specific relightable MVP with a low-dimensional identity code $\identitycode$ for our generative volumetric morphable head model. Then we discuss our modification to the network architecture and training process for disentanglement and better reconstruction quality.

\paragraph{Multi-identity model}

Formally, our \sysname maps the identity code $\identitycode$, the expression code $\expressioncode$, the view direction $\direction$, and the lighting condition $\light$ to the base mesh and corresponding volumetric primitives:
\begin{equation}
\begin{aligned}
\{\vertex, R_{p}, t_{p}, s_{p}, \voxelopacity, \voxelcolor\} = {\rm VRMM} (\identitycode, \expressioncode, \direction, \light).
\end{aligned}
\end{equation}
Specifically, \sysname is composed of five decoders:
\begin{equation}
\begin{aligned}
{\rm VRMM} :=\{ \meshdecoder, \identitydecoder, \transformdecoder, \opacitydecoder, \colordecoder \}.
\end{aligned}
\end{equation}
The mesh decoder $\meshdecoder$ is a multi-layer perceptron that predicts the vertex positions $\vertex$ of the base mesh given the identity code $\identitycode$ and expression code $\expressioncode$. The convolutional identity decoder $\identitydecoder$ maps the low-dimensional identity code $\identitycode$ to hierarchical feature maps $\alphafeaturemap$ and $\appearancefeaturemap$ that are injected to the opacity decoder $\opacitydecoder$ and appearance decoder $\colordecoder$, respectively. The transformation decoder $\transformdecoder$ maps the identity code $\identitycode$ and expression code $\expressioncode$ to the rotation $R_{p}$, translation $t_{p}$, and scale $s_{p}$ of the primitives for identity-related expression decoding. The opacity decoder $\opacitydecoder$ predicts the voxel opacity $\voxelopacity$ of the primitives conditioned on the expression code $\expressioncode$ and the feature maps $\alphafeaturemap$ from the identity encoder. The relightable appearance decoder $\colordecoder$ adopts \yl{an} architecture similar to \cite{yang2023towards} that takes view direction $\direction$, lighting condition $\light$, and expression code $\expressioncode$ as input to predict the voxel color $\voxelcolor$ with a linear branch for lighting related decoding and $\appearancefeaturemap$ hierarchically injected to the non-linear branch.
%: $\vertex = \meshdecoder(\identitycode, \expressioncode)$. 

Considering the fact that the transformation and appearance of the volumetric primitives are closely related, we further concatenate the feature maps from the intermediate layers of the convolutional transformation decoder $\transformdecoder$ to corresponding layers of the opacity decoder $\opacitydecoder$ and the non-linear branch of the relightable appearance decoder $\colordecoder$, and vice-versa. However, direct concatenation leads to unstable convergence during training, which we believe is due to the scale discrepancy of gradient in different branches. So we stop the gradient back-propagation through the concatenated feature maps. We empirically find that the detach-concatenate operation significantly alleviate the jittering of the volumetric primitives when the \yl{numbers} of \yl{identities} and training \yl{frames} increase. 

Different from \cite{cao2022authentic} where the meshes and textures are directly fed into the prior model for identity encoding, \sysname learns to generate novel relightable identities from the low-dimensional identity code, enabling \sysname to be used for avatar reconstruction by directly fitting to images without depending on traditional mesh-based parametric face models. 

%\subsubsection{Unified Expression Space}
\paragraph{Disentangled latent space training}

Different from previous volumetric head priors~\cite{zhuang2022mofanerf, hong2022headnerf} that are trained on limited predefined expressions, our \sysname is designed to \yl{encompass a} full range of dynamic expressions for animatable avatar reconstruction. Besides, the relightable appearance means capturing under varying lighting conditions, which leads to a significant \yl{challenge},
%difficulty, 
as it is not feasible to capture many people performing a large number of identical expressions under different lighting \yl{conditions} even in the studio.

Inspired by recent cross-identity neural retargeting methods~\cite{zhang2022video, xu2023latentavatar}, we propose a novel framework for our \sysname to learn shared expression latent space across identities without explicit supervision. We also adopt the tracking-free training pipeline~\cite{yang2023towards} that jointly learns topology consistent tracking end-to-end from dynamic image sequences, which avoids the brittle preprocessing step of surface-tracking \yl{(especially with changing lightings)} and enables scalable training for \yl{a} large number of different identities.

Specifically, given the training dataset with image $I_c^{i, m}$ of dynamic performance of subject $i$ captured by camera $c$ at \yl{frame} $m$ under known illumination $\light^m$, we jointly train an expression encoder $\expressionencoder$ that predicts the mean and variance of a multi-variant Gaussian distribution for expression code as well as a transformation encoder $\transformencoder$ that predicts the rigid rotation $R^{i,m}$ and translation $t^{i,m}$ of the head in that frame from a subset of camera views $I_{c_{input}}^{i, m}$. The expression code $\expressioncode$ is \yl{then} sampled from the Gaussian distribution. The encoding process can be represented as:
\begin{equation}
\begin{aligned}
\expressioncode^{i, m}=\expressionencoder(I_{c_{input}}^{i, m}),
\end{aligned}
\end{equation}
\begin{equation}
\begin{aligned}
R^{i,m}, t^{i,m}=\transformencoder(I_{c_{input}}^{i, m}).
\end{aligned}
\end{equation}
As for the identity code, instead of using the full auto-encoder framework, we adopt the decoder-only method~\cite{buhler2023preface, wang2022morf} where the identity code $\identitycode^{i}$ for each subject is initialized with Gaussian noise and jointly optimized with the networks during training. Note that the auto-decoder framework learns a generative model though there is not an explicit sampling process during training as discussed in \cite{park2019deepsdf}.
Then the synthesized image $\hat{I}_c^{i, m}$ of a camera $c$ with view direction $\direction_c$ and camera parameters $\cameraparams_c$ is given by:
\begin{equation}
\begin{aligned}
\hat{I}_c^{i, m}=\projection({\rm VRMM}(\identitycode^{i}, \expressioncode^{i, m}, \direction_c, \light^m), R^{i,m}, t^{i,m}, \cameraparams_c),
\end{aligned}
\end{equation}
where $\projection$ represents the aforementioned differentiable ray-marching process for rendering.

The loss $\totalloss$ of the training objective function consists of three parts:
\begin{equation}
\label{eq:trainloss}
\begin{aligned}
\totalloss=\dataloss+\regularizationloss+\expressionloss,
\end{aligned}
\end{equation}
where $\dataloss=\loss_1+\lambda_{VGG}\loss_{VGG}$ comparing the reconstruction $\hat{I}_c^{i, m}$ and the observed image $I_c^{i, m}$ consists of the $L_1$ loss term $\loss_1$ and the perceptual loss term $\loss_{VGG}$ with weight $\lambda_{VGG}$. The regularization loss is given by:
\begin{equation}
\begin{aligned}
% \ifthenelse{\equal{\arxiv}{0}}
% {
\regularizationloss=\lambda_{KLD}\loss_{KLD}+\lambda_{vol}\loss_{vol}+\lambda_{scale}\loss_{scale}+\lambda_{id}||\identitycode||^2_2,
% }
% {
% \regularizationloss=&\lambda_{KLD}\loss_{KLD}+\lambda_{vol}\loss_{vol}\\&+\lambda_{scale}\loss_{scale}+\lambda_{id}||\identitycode||^2_2,
%}
\end{aligned}
\end{equation}
where $\loss_{KLD}$ is the KL-divergence loss between the distribution of the expression code $\expressioncode$ and multi-variant Gaussian prior. $\loss_{vol}$ is the volume minimization prior proposed in MVP~\cite{lombardi2021mixture}. \revision{The scale regularization term $\loss_{scale}$ penalizes the $k$ sides with the shortest length in the $N_{prim}$ cubic primitives to prevent the volumetric primitives from squeezing without affecting normal primitives: 
\begin{equation}
\begin{aligned}
%\loss_{scale}=\sum_{s \in k\ {\rm shortest\ sides\ of\ s_{p}}  }\log(1/s).
%\loss_{scale}=\sum_{n} \log(1/{s_p})_n,
\loss_{scale}=\sum_{n\in \mathcal{S}} \log(1/{s_p^n}),
\end{aligned}
\end{equation}
where $\mathcal{S}$ is the set of indices of the $k$ shortest sides among all the sides $s_p$, and $s_p^n$ is the predicted scale of the $n$-th side. }
%$n$ is the index of the shortest $k$ sides in $s_p$.} 
The $L2$ regularization term on identity code $\identitycode$ is derived by assuming the prior distribution of $\identitycode$ to be multi-variant Gaussian distribution. $\lambda_{KLD}$, $\lambda_{vol}$, $\lambda_{scale}$, and $\lambda_{id}$ are balancing weights.

We find that the KL-divergence regularization $\loss_{KLD}$ on the expression code plays an important role for our \sysname to learn a disentangled representation for identity and expression. By applying a much \yl{larger} regularization weight $\lambda_{KLD}$ on expression code \yl{compared} to $\lambda_{id}$, the information in $\expressioncode$ is limited so that the expression encoder $\expressionencoder$ learns to extract only expression-related information while lighting and identity are injected from other branches. We also experiment with the adversarial training for disentanglement~\cite{zhang2022video, schwartz2020eyes} but observe no improvement.

The system tends to learn a shared expression space even without \yl{a} specific constraint as the decoders are shared across identities. To further align the semantic meaning for the expression code $\expressioncode$ for different identities, we propose to incorporate a novel expression consistency loss $\expressionloss$. Specifically, given the expression code $\expressioncode^{i, m}$ predicted by expression encoder $\expressionencoder$ from $I_{c_{input}}^{i, m}$, we randomly choose an identity code $\identitycode^j$ from another identity $j$ and a different lighting condition $\light^n$. Then we render images corresponding to $\expressioncode^{i, m}$, $\identitycode^j$, and $\light^n$:
\begin{equation}
\begin{aligned}
% \ifthenelse{\equal{\arxiv}{0}}
% {
\hat{I}_{c_{input}}^{j, n}=\projection({\rm VRMM}(\identitycode^{j}, \expressioncode^{i, m}, \direction_{c_{input}}, \light^n), R^{i,m}, t^{i,m}, \cameraparams_{c_{input}}).
% }
% {
% \hat{I}_{c_{input}}^{j, n}=\projection({\rm VRMM}(&\identitycode^{j}, \expressioncode^{i, m}, \direction_{c_{input}}, \light^n), \\&R^{i,m}, t^{i,m}, \cameraparams_{c_{input}}).
% }
\end{aligned}
\end{equation}
$\hat{I}_{c_{input}}^{j, n}$ should have %save 
\yl{the same}
expression as in $I_{c_{input}}^{i, m}$ but different identity and lighting condition. We feed $\hat{I}_{c_{input}}^{j, n}$ to the expression encoder $\expressionencoder$ to extract the corresponding expression code $\hat{\expressioncode}^{i, m}$. We also render the image $\hat{\hat{I}}_c^{i, m}$ with $\hat{\expressioncode}^{i, m}$, the original identity $\identitycode^i$, and lighting $\light^m$, as directly measuring the distance in the parameter space is shown to be inefficient~\cite{tewari2017mofa, tewari2020stylerig}. The expression consistency loss $\expressionloss$ is then given by:
\begin{equation}
\begin{aligned}
\expressionloss=\lambda_{imgexp}\loss_{imgexp}+\lambda_{parexp}||\hat{\expressioncode}^{i, m} - \expressioncode^{i, m}||^2_2,
\end{aligned}
\end{equation}
where $\loss_{imgexp}$ measures the image space difference similar to $\dataloss$, $\lambda_{imgexp}$ and $\lambda_{parexp}$ are the weights of different terms.

\subsection{Model Fitting}

\begin{figure}%[t]
    \centering
    \begin{minipage}{\linewidth}
        \centering
        \hspace{\onefourthfigurewidth\linewidth}
        \begin{minipage}{\onefourthfigurewidth\linewidth}
            \centering
            \includegraphics[width=\linewidth]{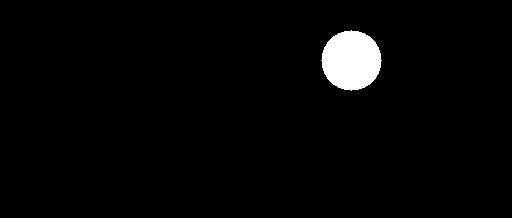}
        \end{minipage}
        \begin{minipage}{\onefourthfigurewidth\linewidth}
            \centering
            \includegraphics[width=\linewidth]{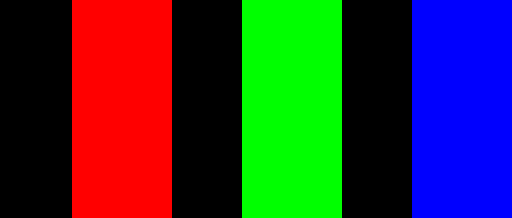}
        \end{minipage}
        \begin{minipage}{\onefourthfigurewidth\linewidth}
            \centering
            \includegraphics[width=\linewidth]{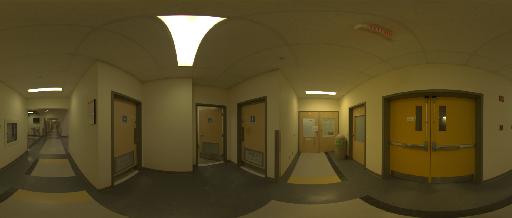}
        \end{minipage}
    \end{minipage}

    \begin{minipage}[t]{\onefourthfigurewidth\linewidth}
        \centering
        \includegraphics[width=\linewidth]{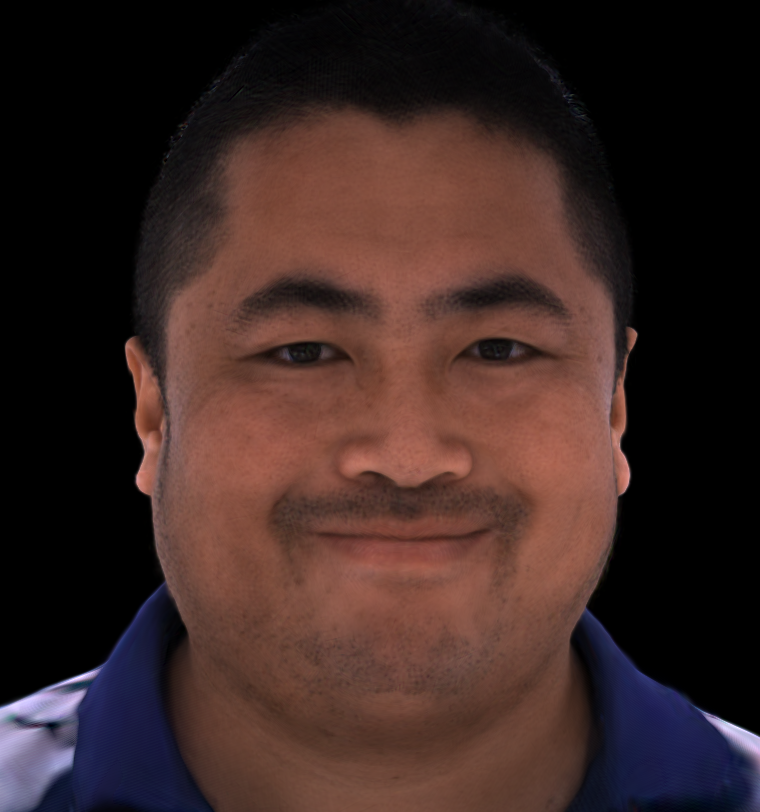}
        \includegraphics[width=\linewidth]{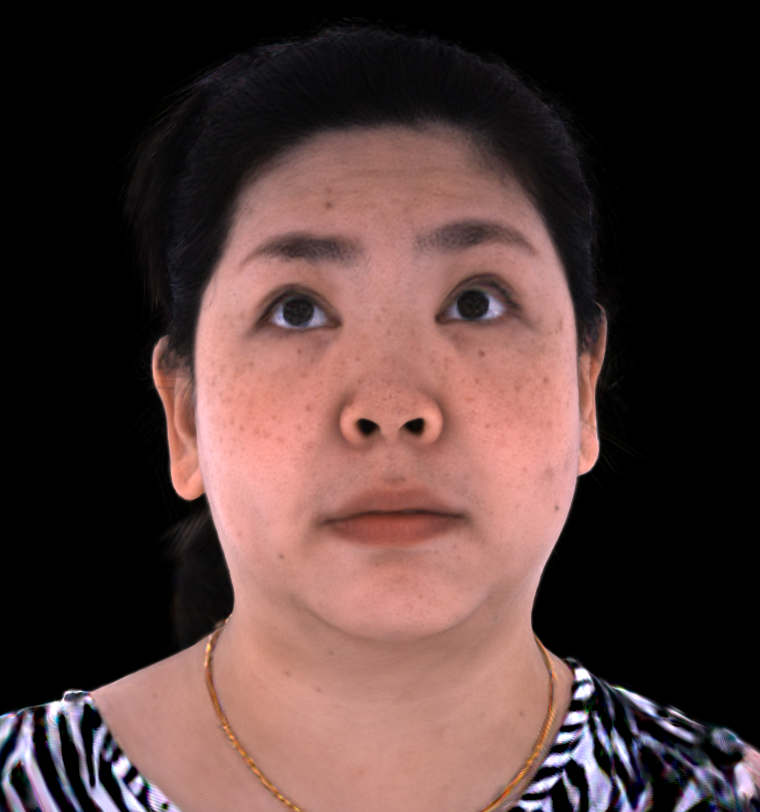}
        \subfloat{Subject}
       % \label{fig:disentanglement_sub1}
    \end{minipage}
    \begin{minipage}[t]{\onefourthfigurewidth\linewidth}
        \centering
        \includegraphics[width=\linewidth]{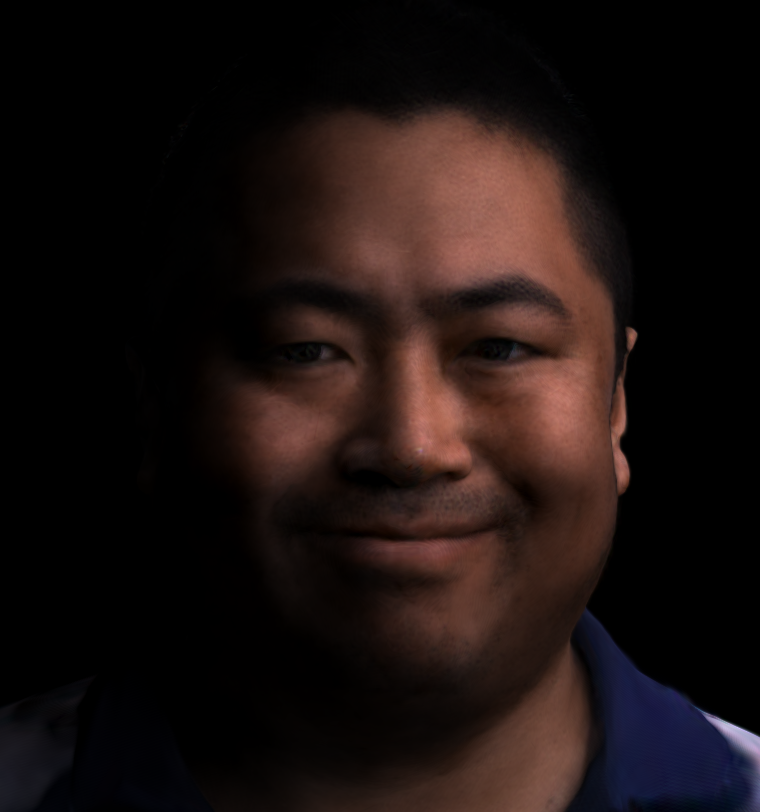}
        \includegraphics[width=\linewidth]{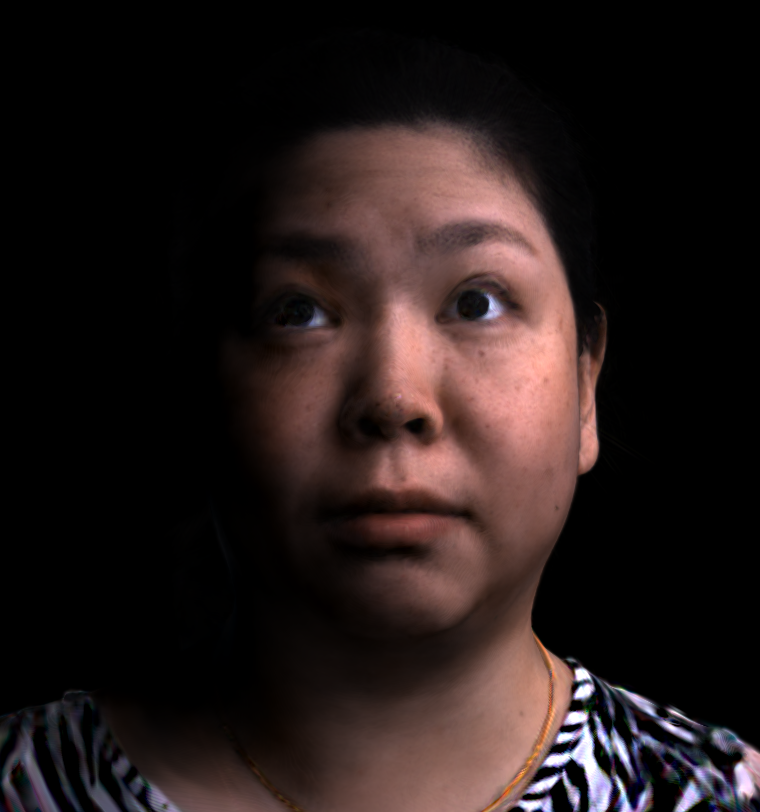}
        \subfloat{Lighting 1}
        %\label{fig:disentanglement_sub1_lightA}
    \end{minipage}
    \begin{minipage}[t]{\onefourthfigurewidth\linewidth}
        \centering
        \includegraphics[width=\linewidth]{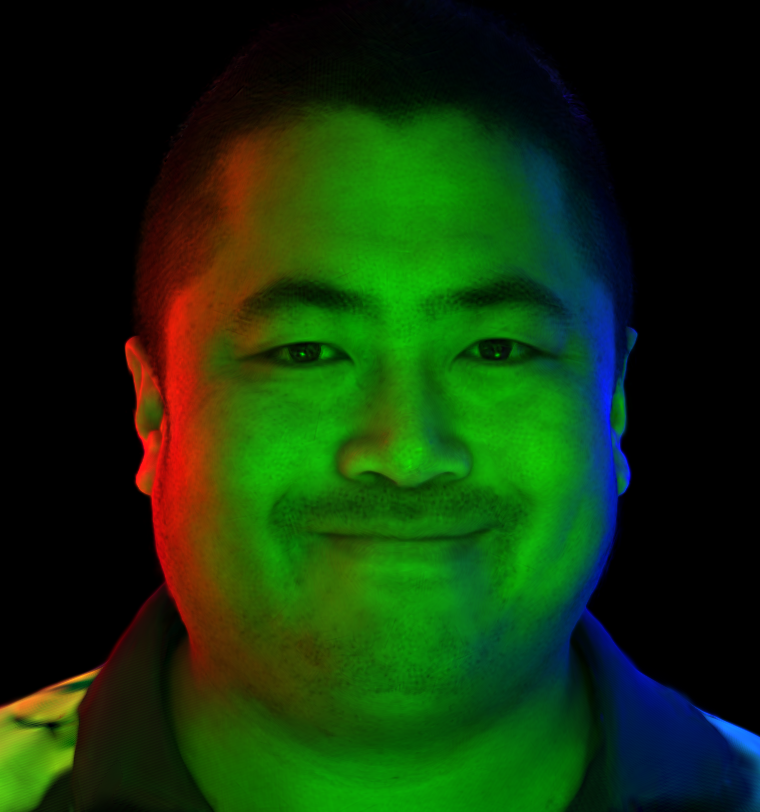}
        \includegraphics[width=\linewidth]{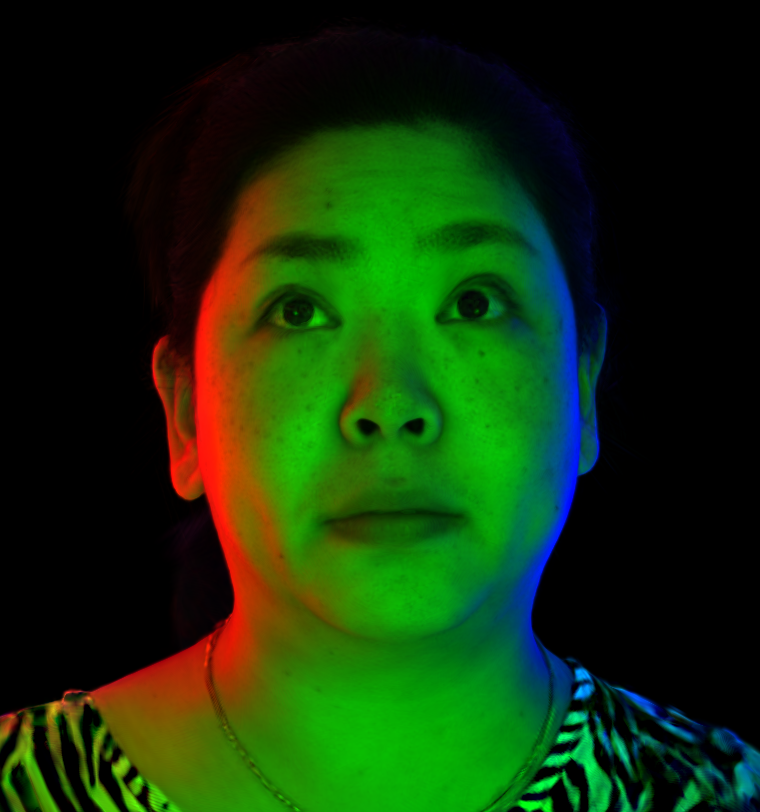}
        \subfloat{Lighting 2}
        %\label{fig:disentanglement_sub1_lightB}
    \end{minipage}
    \begin{minipage}[t]{\onefourthfigurewidth\linewidth}
        \centering
        \includegraphics[width=\linewidth]{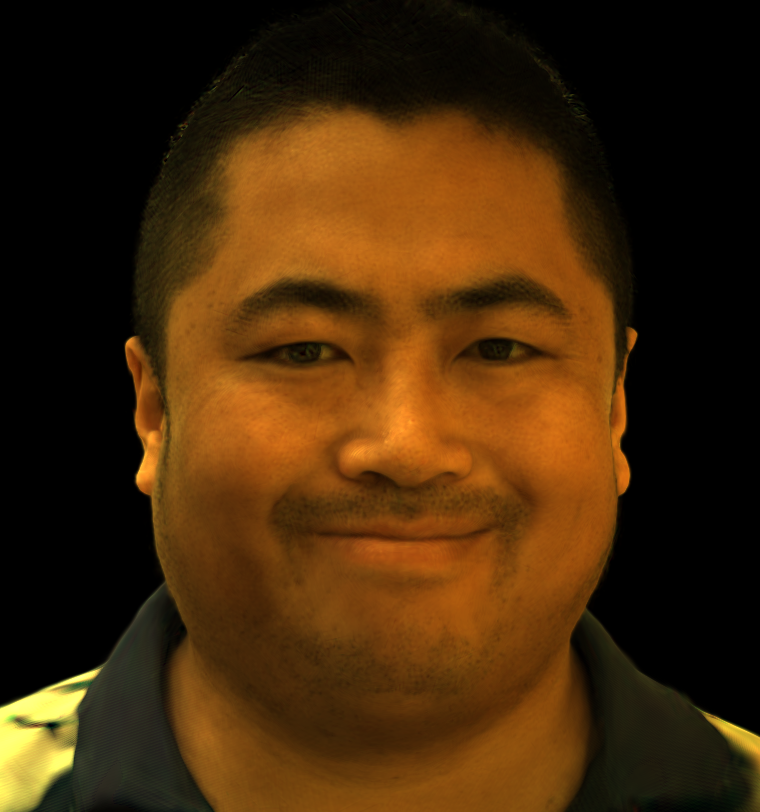}
        \includegraphics[width=\linewidth]{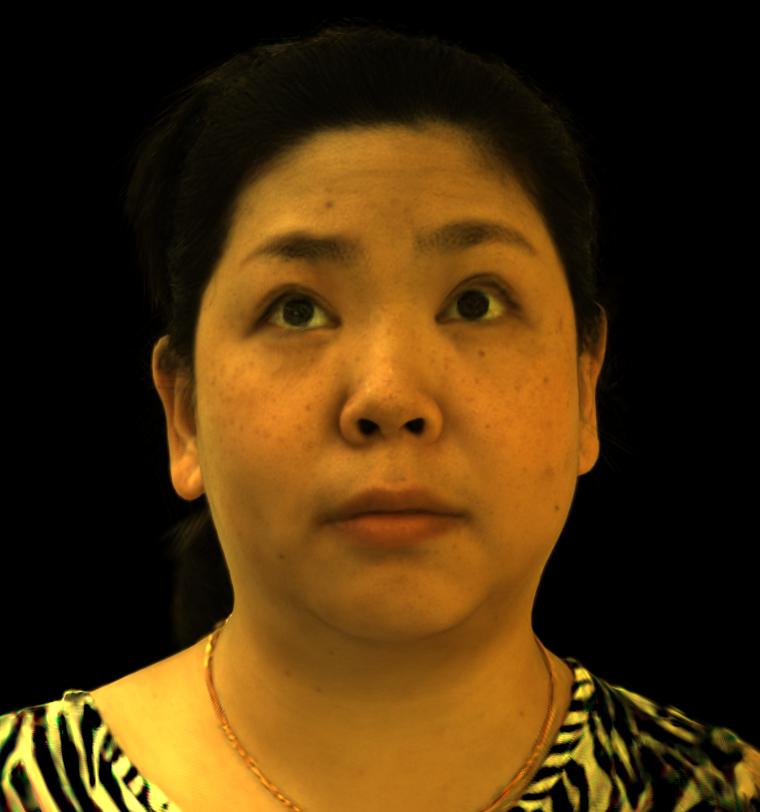}
        \subfloat{Lighting 3}
        %\label{fig:disentanglement_sub2}
    \end{minipage}
%    \caption{The lighting and motion are fully disentangled and consistent across different subjects.}
    \caption{Our model allows real-time global illumination. The lighting condition \yl{is} represented as latitude-longitude environment maps, which is shown on the top.
%    \todo{Update captured input images under different illuminations.}
%    (\subref{fig:disentanglement_sub1})/(\subref{fig:disentanglement_sub2})
    }
\label{fig:relighting}
\end{figure}

Once trained, \sysname can be used for avatar reconstruction with the analysis-by-synthesis scheme as traditional 3DMMs. However, volumetric \yl{avatars include} complex components that cannot be expressed by the constrained parametric space of the prior model, as it was trained with only hundreds of identities. For faithful reconstruction, we further finetune \sysname for personalized avatar generation similar to \cite{buhler2023preface, cao2022authentic, wang2022morf}. We find %release the constraint
%%%YKL Do you mean relaxing? and which constraint(s)?
finetuning can increase the reconstruction quality but make manipulation deteriorate. Similar distortion-editability \yl{trade-off} is also demonstrated in the GAN inversion field~\cite{tov2021designing, zhu2020improved}. Our fitting pipeline is specially designed to preserve the data-driven prior that can create a relightable and animatable avatar from even a single image. We illustrate our pipeline for fitting a single in-the-wild image for clarity, which can be easily extended to image sets.

\paragraph{Inversion}
Given an image, we first detect 2D facial landmarks and optimize the rough rigid transformation and camera projection parameters with respect to the corresponding predefined 3D landmarks on the mean base mesh of \sysname for initialization. Then we jointly optimize the input parameters of \sysname and camera projection by inverse rendering. Particularly, instead of directly \yl{solving for} the identity code $\identitycode$, we find a set of weights $\blendweight$ 
%%%YKL one weight or a set of weights?
% a vector with n identity numbers
that linearly
\yl{blends} the existing identity codes in the training set to constrain the domain. The \yl{objective} function is formulated as:
\begin{equation}
\begin{aligned}
\argmin_{\blendweight, \expressioncode, \light, R, t, \cameraparams} \dataloss+\lambda_{exp}||\expressioncode||_2^2,
\end{aligned}
\end{equation}
where $\dataloss$ is identical to the data term in Equation~\ref{eq:trainloss} and $\lambda_{exp}$ is the weight of regularization term. The lighting $\light$ is restrict to be non-negative during optimization. The view direction $\direction$
can be computed as inversion of the rotation $R$.

\paragraph{Fine-tuning}
After inversion we obtain an avatar that is animatable and relightable but \yl{may} not authentically replicate the image. Inspired by Pivotal Tuning~\cite{roich2022pivotal} and DreamBooth~\cite{ruiz2023dreambooth}, we propose to finetune the parameters $\theta$ of \sysname with the prior preservation technique to better reproduce the individual characteristics:
\begin{equation}
\begin{aligned}
\argmin_{\expressioncode, \light, R, t, \cameraparams, \theta} \dataloss+\lambda_{LR}\loss_{LR}+\lambda_{id}||\identitycode||_2^2+\lambda_{exp}||\expressioncode||_2^2,
\end{aligned}
\end{equation}
where $\lambda_{LR}$, $\lambda_{id}$, and $\lambda_{exp}$ are balancing weights. $\loss_{LR}$ is the locality regularization term that \yl{restricts} the modification to the local region around the \yl{inverted} identity code $\identitycode^{\ast}$. Specifically, the interpolated identity code $\identitycode^{inter}$ is obtained by linearly blending $\identitycode^{\ast}$ with a randomly selected identity code $\identitycode^i$ from the training set with a weight $\alpha$: 
\begin{equation}
\begin{aligned}
\identitycode^{inter}=\alpha\identitycode^{\ast}+(1-\alpha)\identitycode^i. 
\end{aligned}
\end{equation}
Then we use the interpolated identity code $\identitycode^{inter}$, a randomly selected expression code $\expressioncode$, and a lighting condition $\light$ to render \yl{an} image with both the fine-tuned \sysname and the original \sysname. $\loss_{LR}$ measures the difference of these two images similarly to $\dataloss$. 

We have found that fine-tuning for 500 iterations leads to good convergence, taking about five minutes. Our experiments show that the locality regularization significantly improves the prior preservation while \yl{having minor effect} on the detail reconstruction.

\section{Experiments}

\begin{figure}
    \centering
    \begin{minipage}[t]{\onefourthfigurewidth\linewidth}
        \centering
        \includegraphics[width=\linewidth]{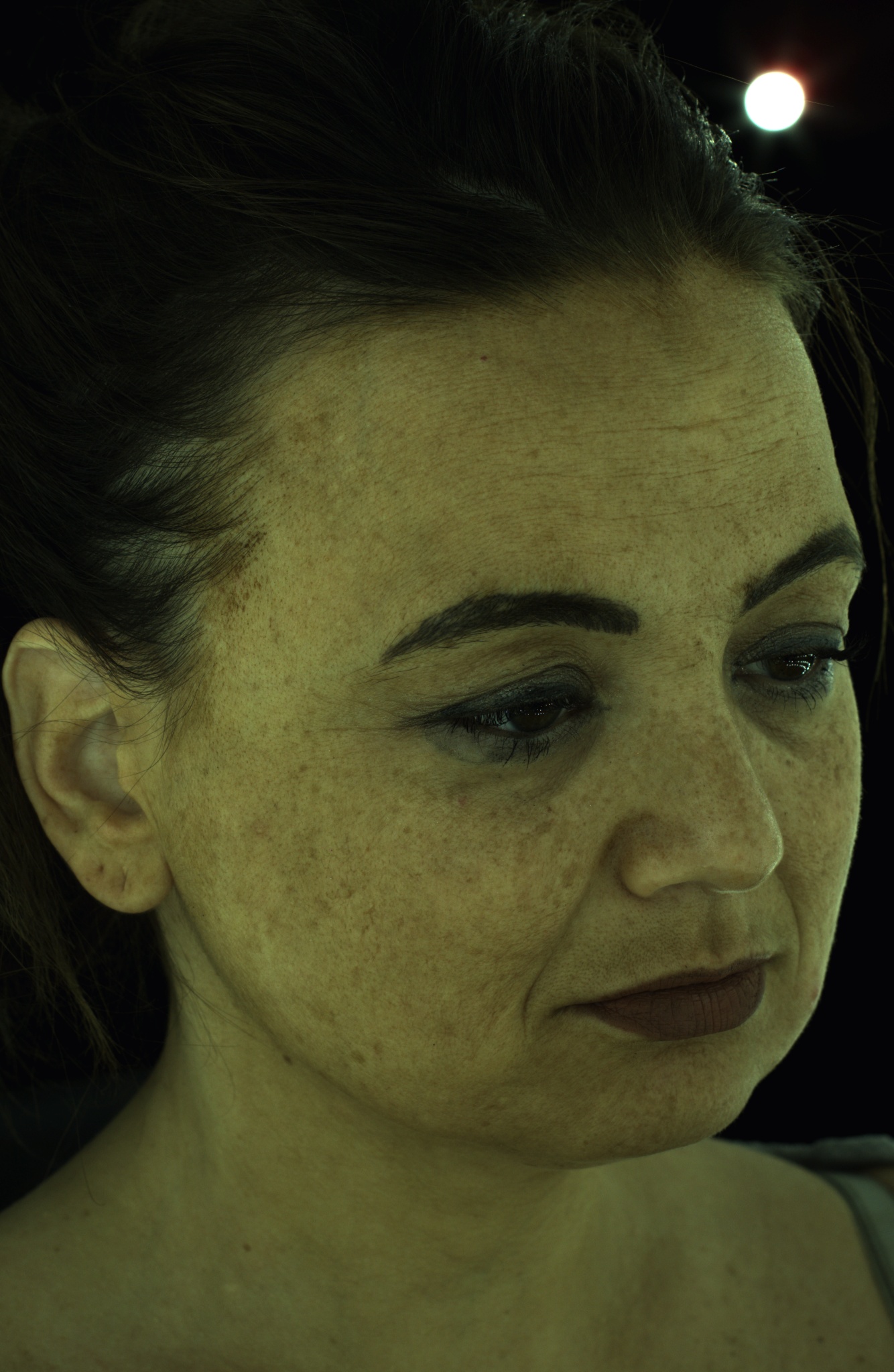}
        \includegraphics[width=\linewidth]{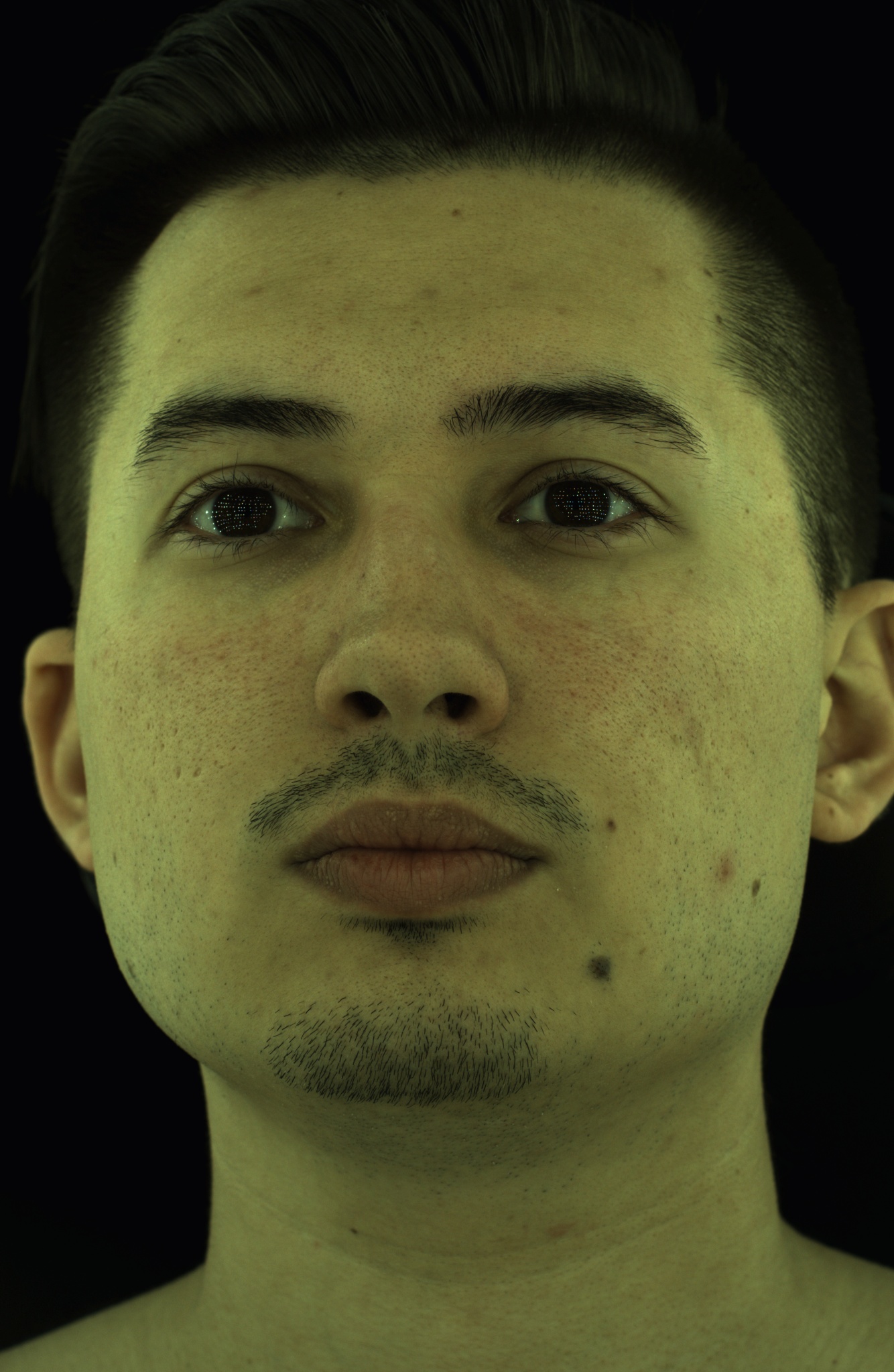}
        \subfloat{Ground truth}
    \end{minipage}
    \begin{minipage}[t]{\onefourthfigurewidth\linewidth}
        \centering
        \includegraphics[width=\linewidth]{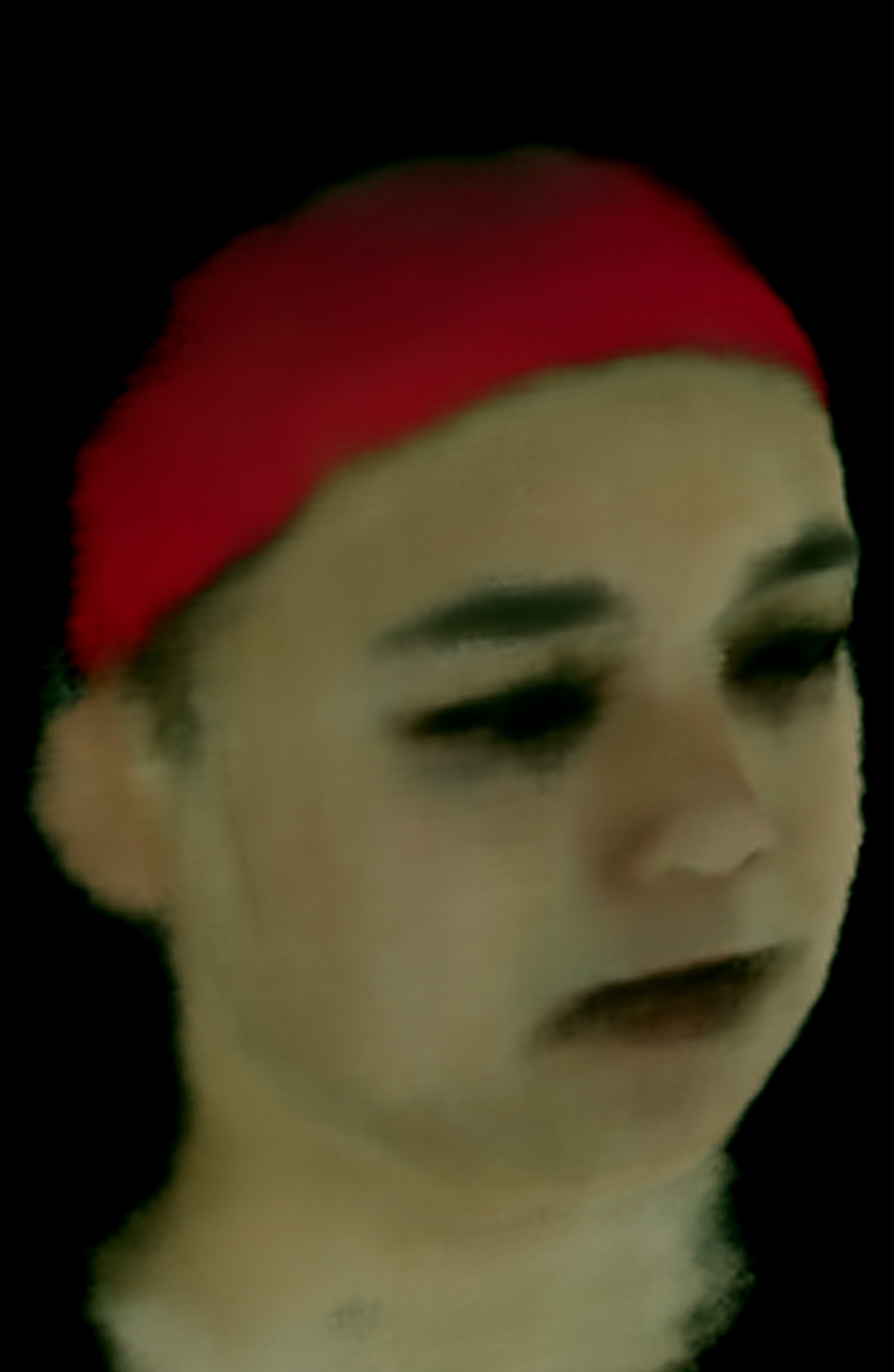}
        \includegraphics[width=\linewidth]{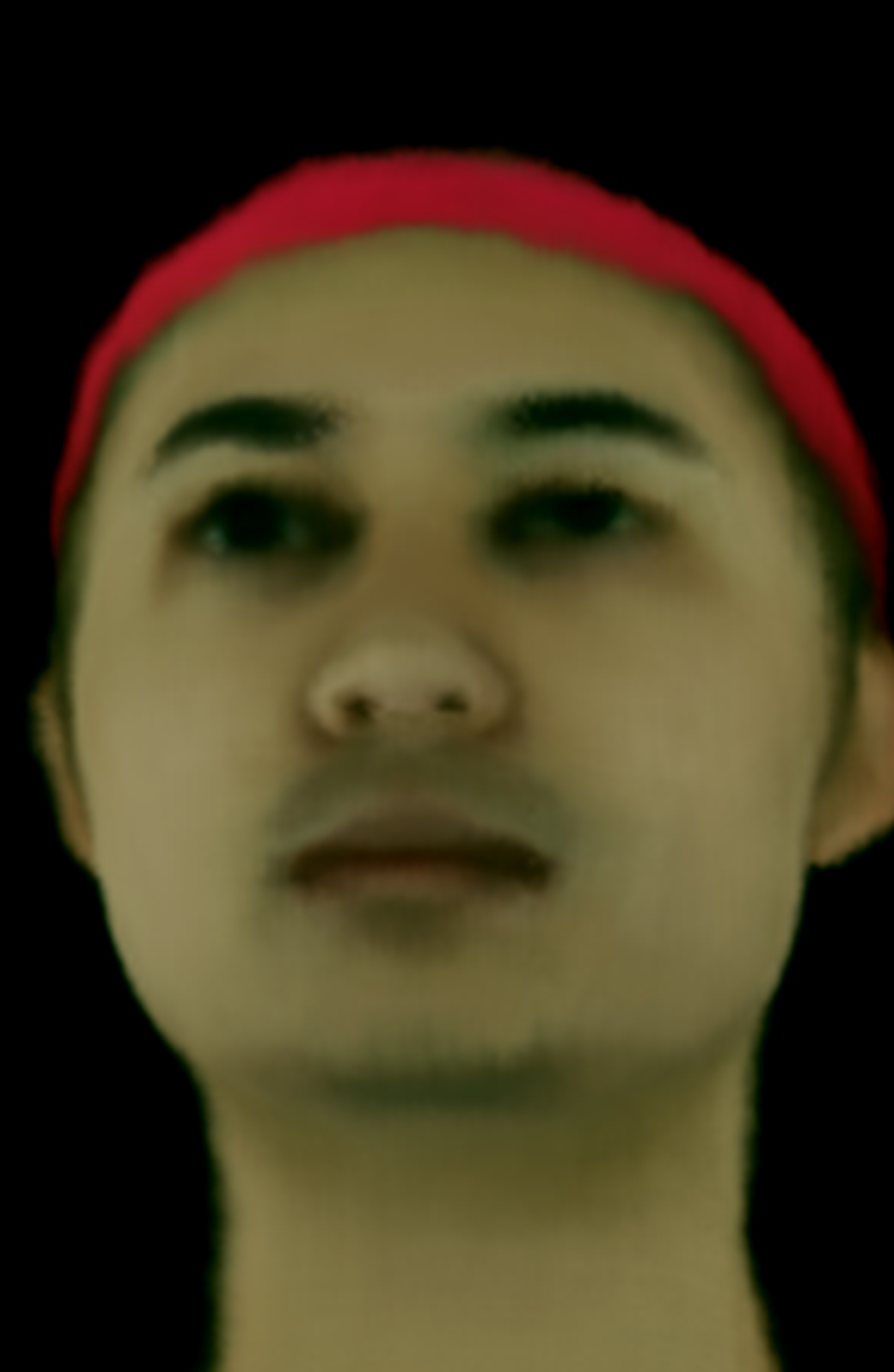}
        \subfloat{MoFaNeRF}
    \end{minipage}
    \begin{minipage}[t]{\onefourthfigurewidth\linewidth}
        \centering
        \includegraphics[width=\linewidth]{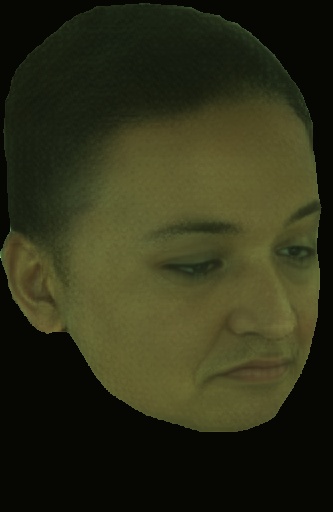}
        \includegraphics[width=\linewidth]{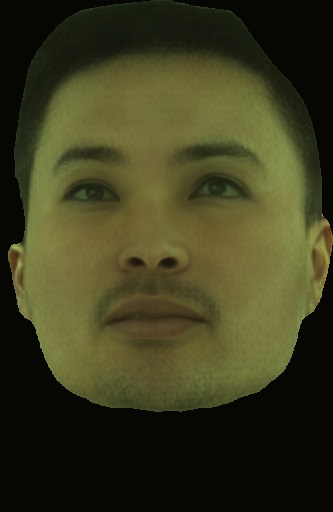}
        \subfloat{HeadNeRF}
    \end{minipage}
    \begin{minipage}[t]{\onefourthfigurewidth\linewidth}
        \centering
        \includegraphics[width=\linewidth]{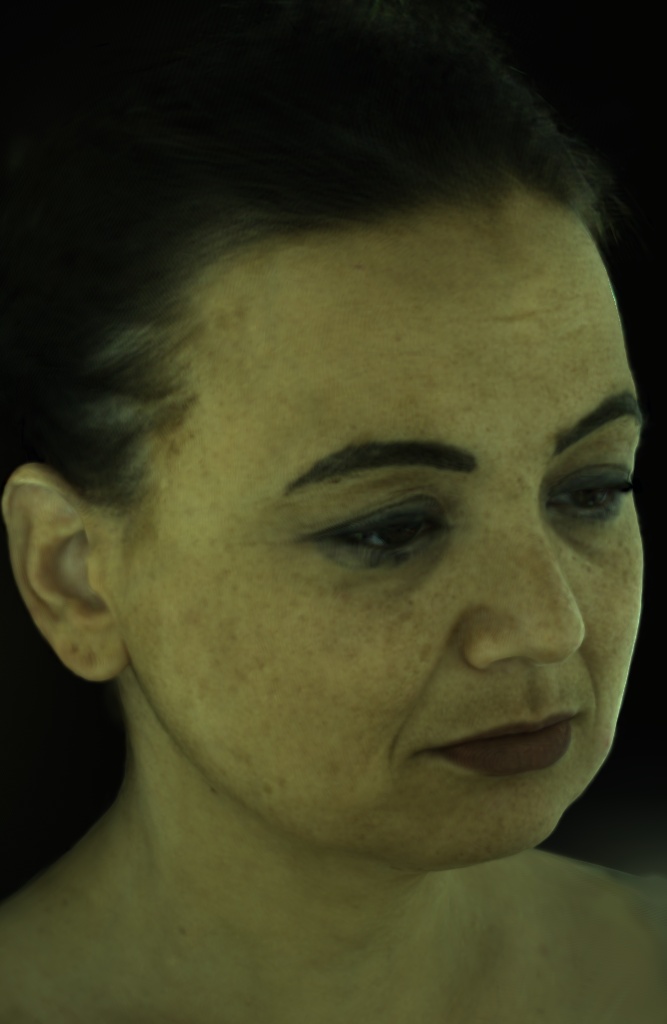}
        \includegraphics[width=\linewidth]{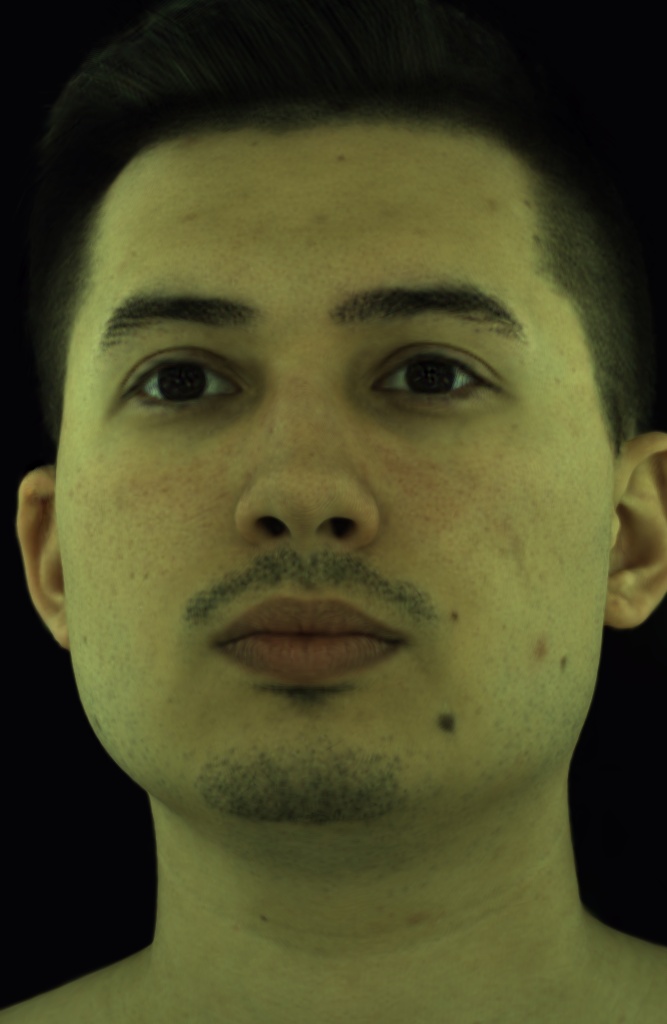}
        \subfloat{Ours}
    \end{minipage}
    %\caption{The trained hybrid mesh-volumetric avatar enjoys nearly consistent topology.}
    \caption{\revision{Qualitative comparison results on novel view synthesis. Our method produces more faithful results compared to existing parametric head models MoFaNeRF~\cite{zhuang2022mofanerf} and HeadNeRF~\cite{hong2022headnerf}.}}
\label{fig:cmp_viewinterp}
\end{figure}

\ifthenelse{\equal{\arxiv}{1}}
{
\begin{figure*}%[t]
    \centering
    \begin{minipage}[t]{\linewidth}
    \begin{minipage}[t]{\linewidth}
        \centering
        \includegraphics[width=\oneninthfigurewidth\linewidth]{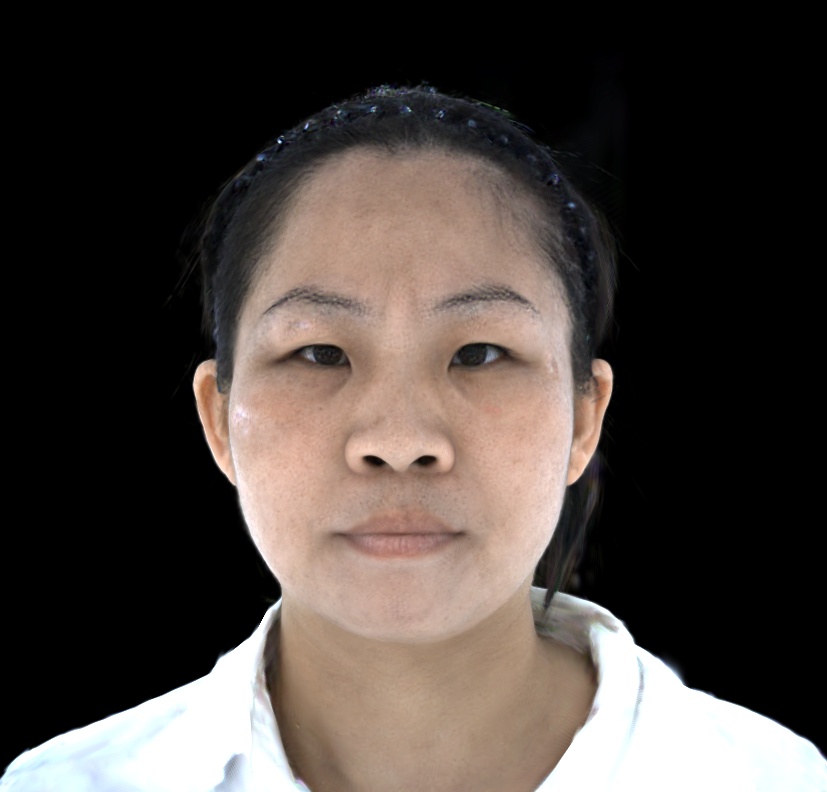}%
        \includegraphics[width=\oneninthfigurewidth\linewidth]{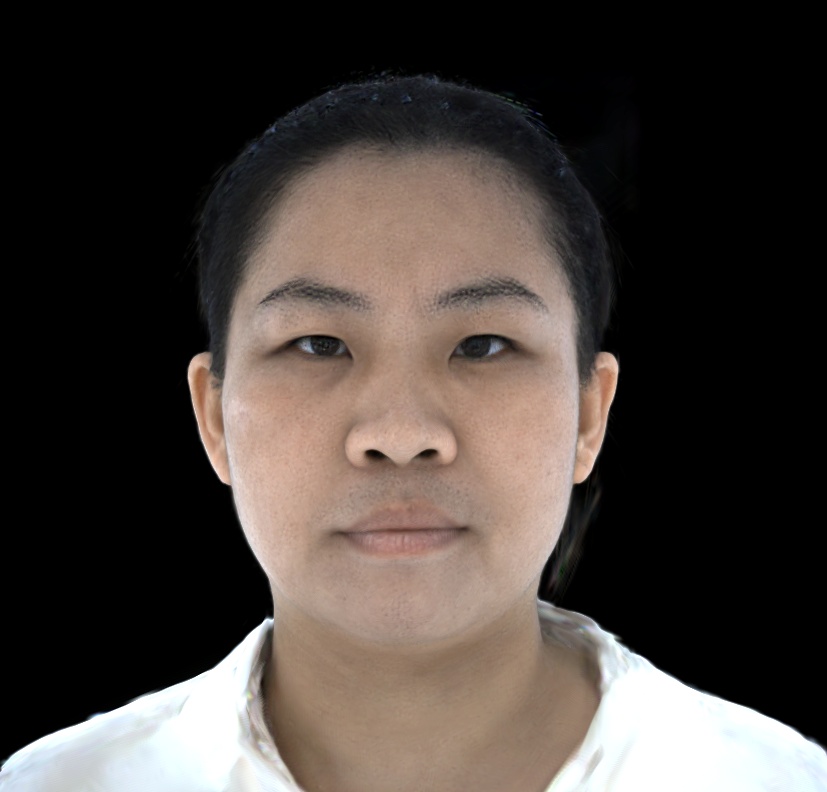}%
        \includegraphics[width=\oneninthfigurewidth\linewidth]{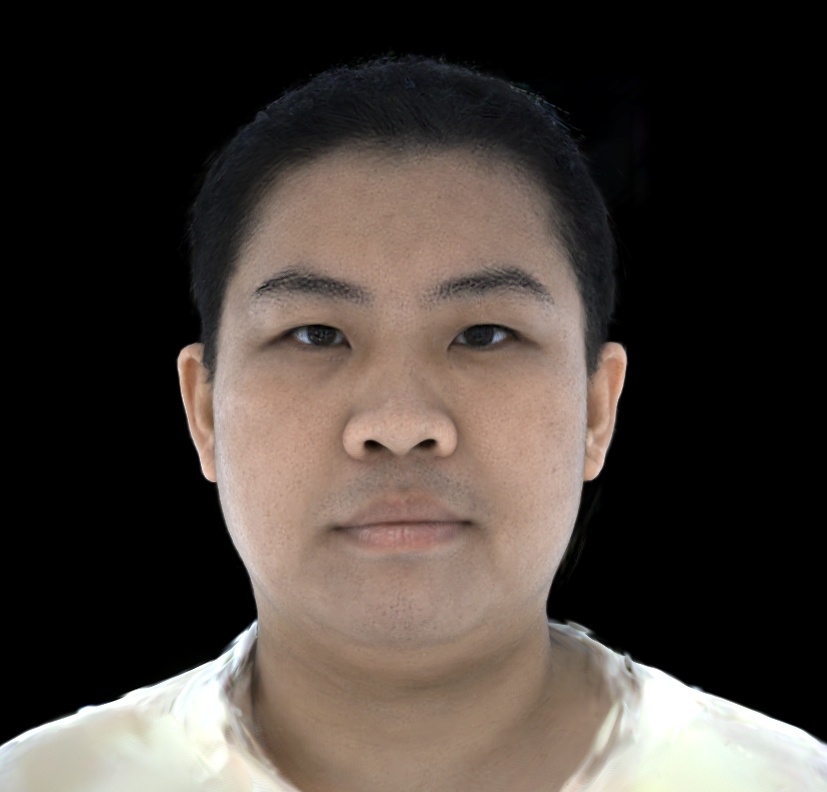}%
        \includegraphics[width=\oneninthfigurewidth\linewidth]{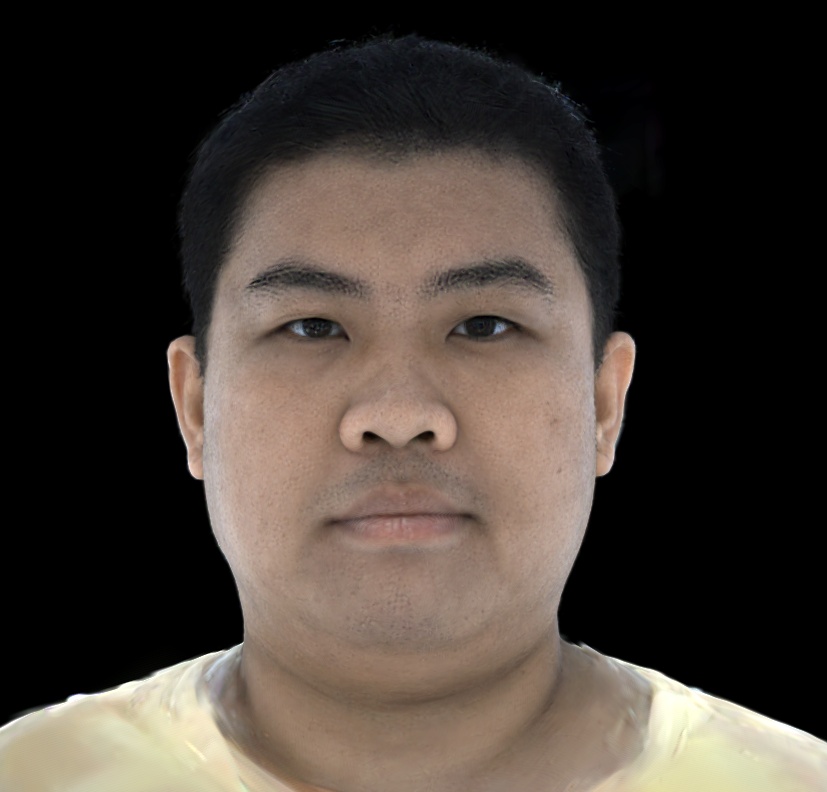}%
        \includegraphics[width=\oneninthfigurewidth\linewidth]{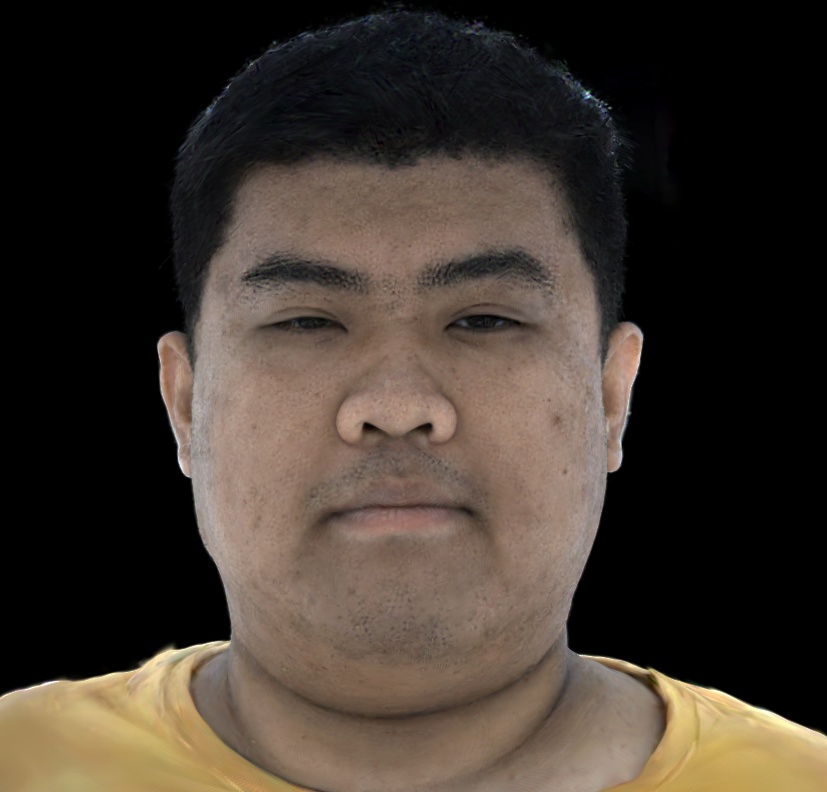}%
        \includegraphics[width=\oneninthfigurewidth\linewidth]{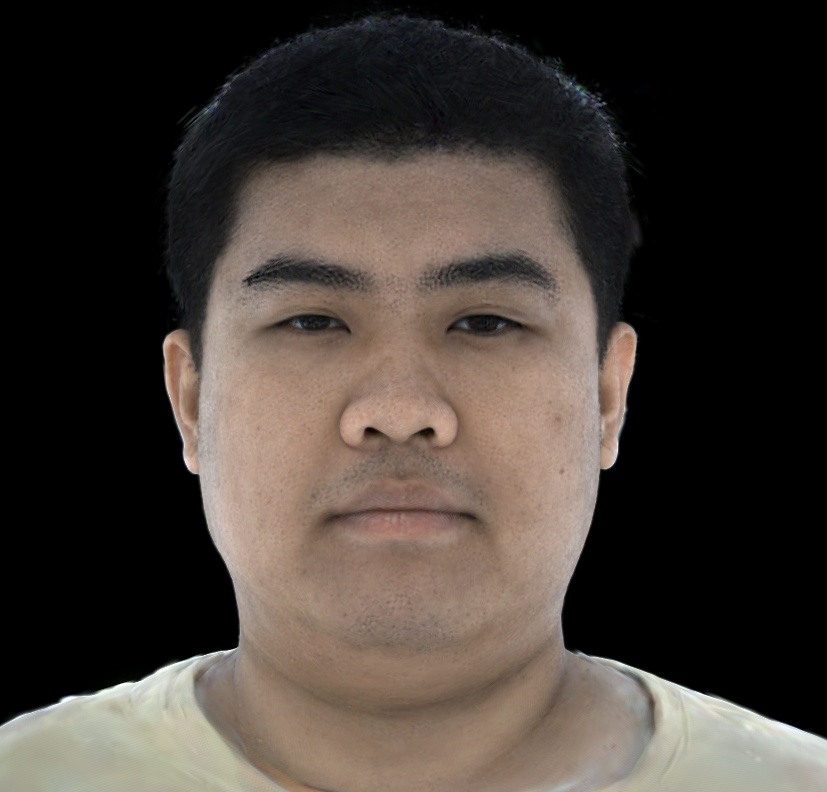}%
        \includegraphics[width=\oneninthfigurewidth\linewidth]{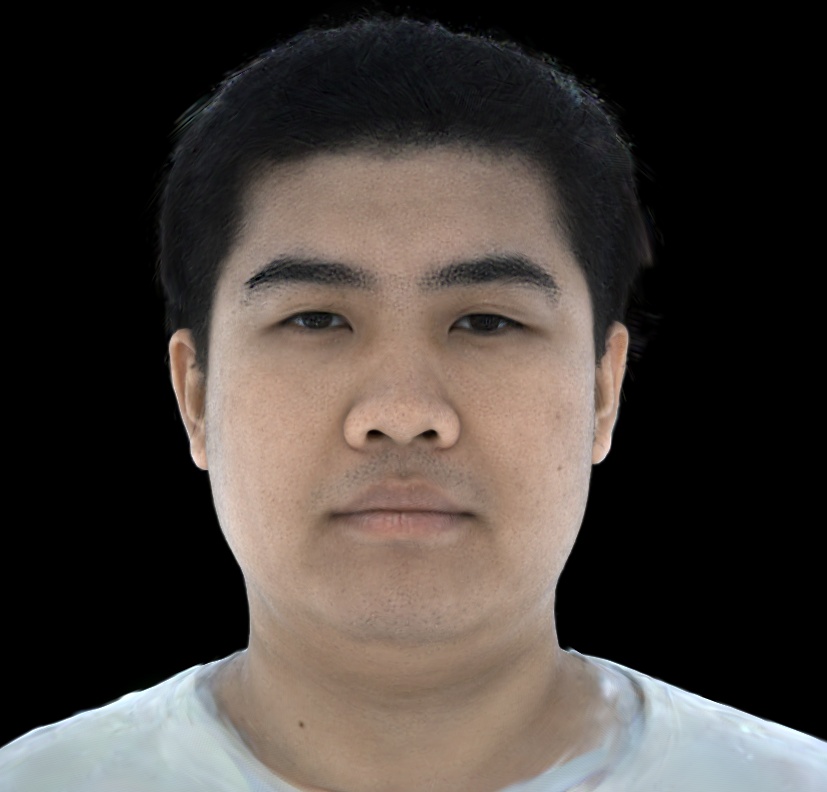}%
        \includegraphics[width=\oneninthfigurewidth\linewidth]{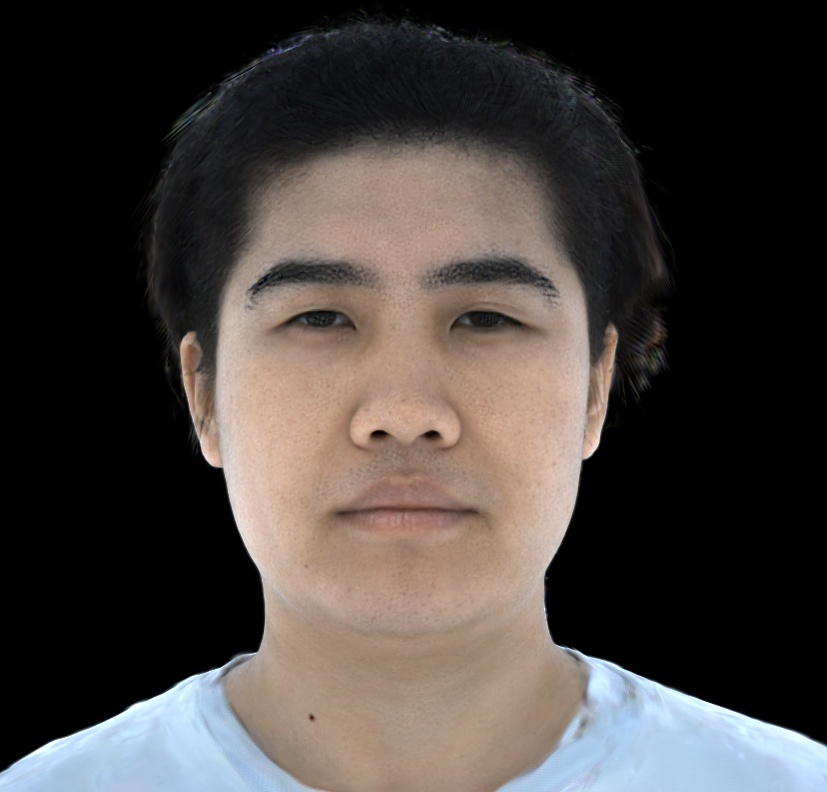}%
        \includegraphics[width=\oneninthfigurewidth\linewidth]{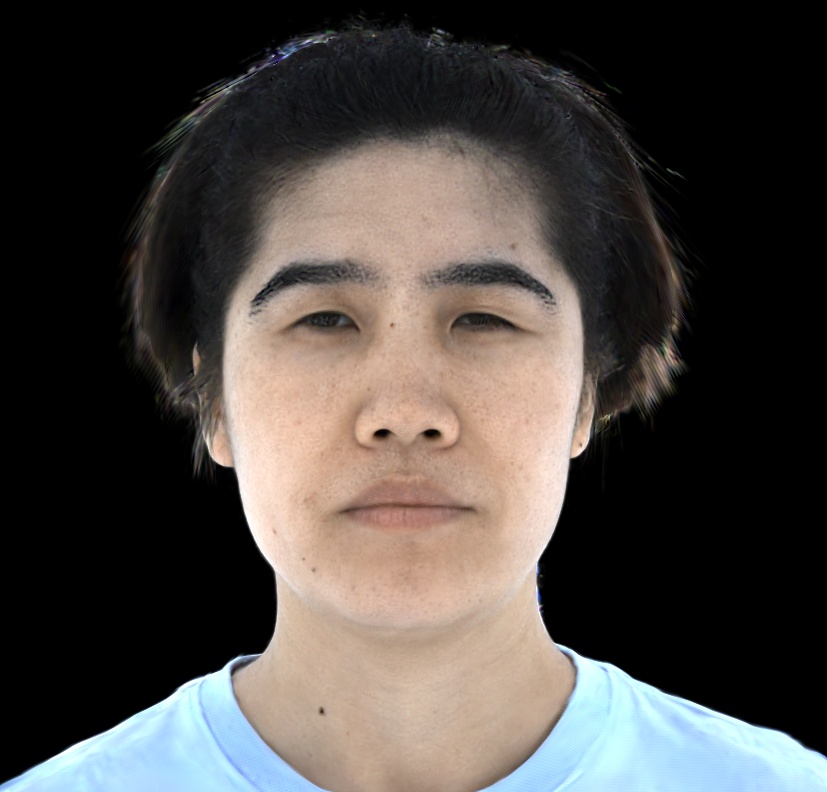}%
    \end{minipage}

    \begin{minipage}[t]{\linewidth}
        \centering
        \includegraphics[width=\oneninthfigurewidth\linewidth]{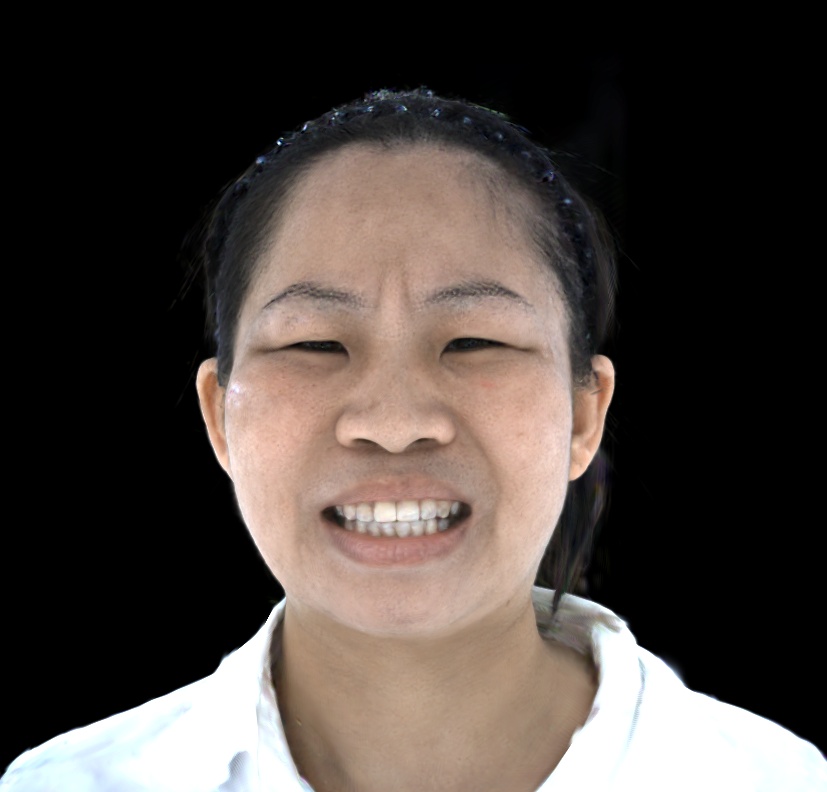}%
        \includegraphics[width=\oneninthfigurewidth\linewidth]{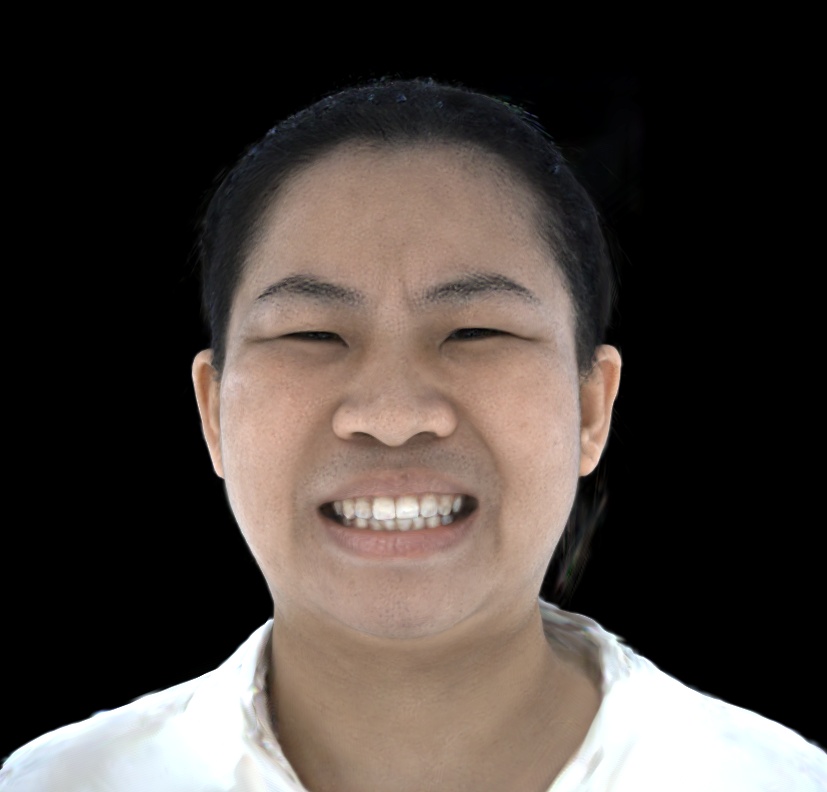}%
        \includegraphics[width=\oneninthfigurewidth\linewidth]{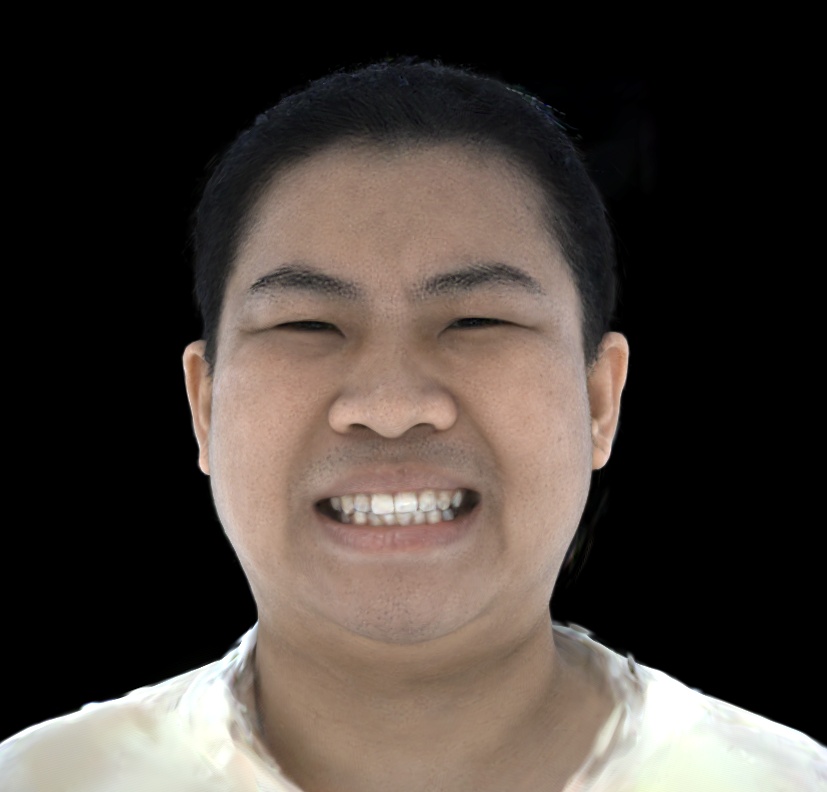}%
        \includegraphics[width=\oneninthfigurewidth\linewidth]{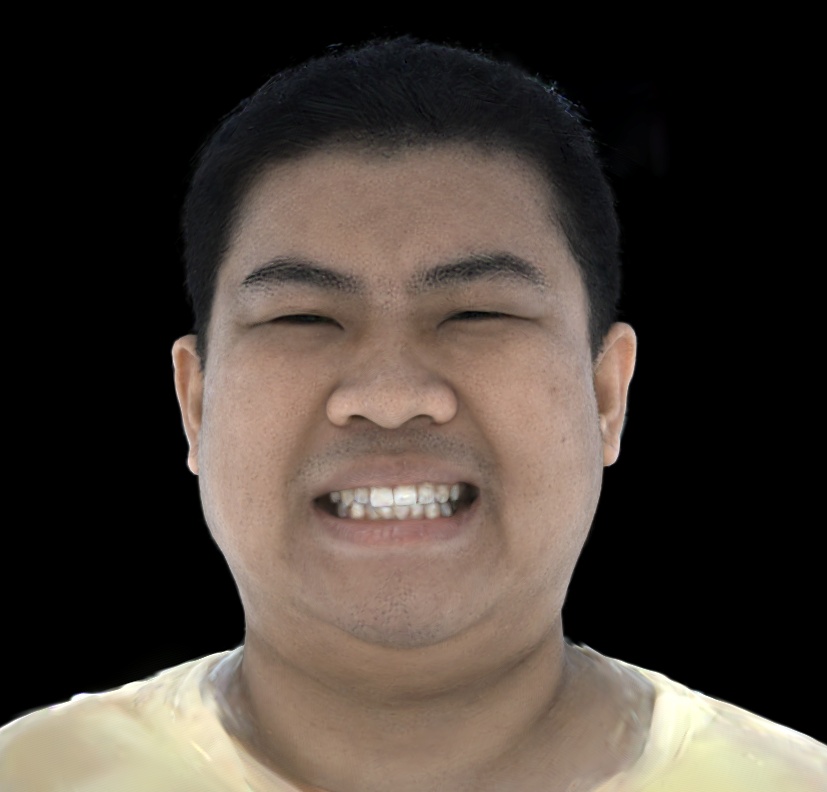}%
        \includegraphics[width=\oneninthfigurewidth\linewidth]{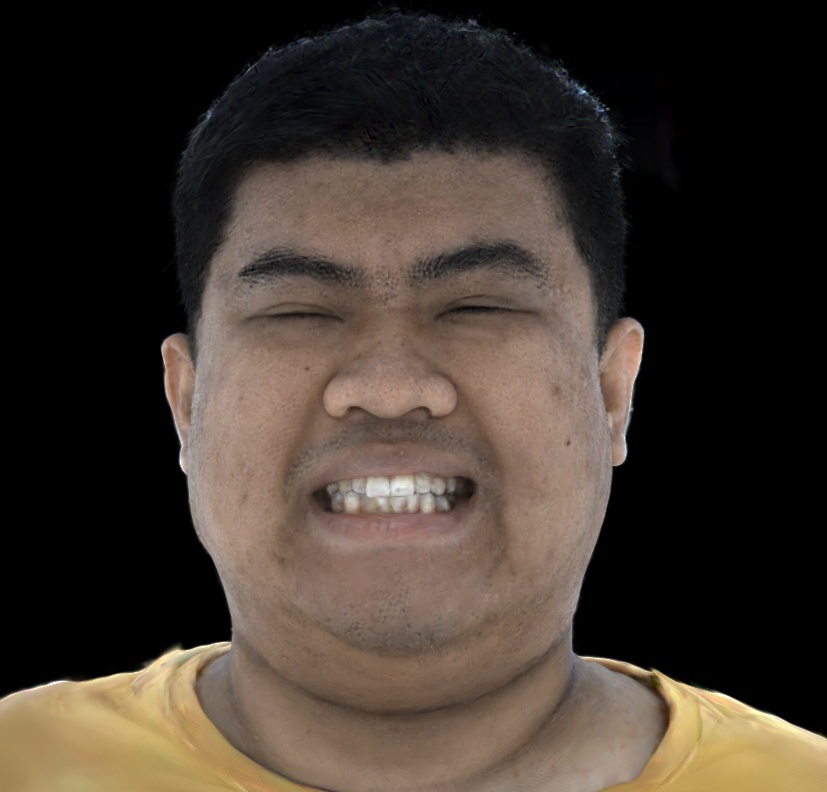}%
        \includegraphics[width=\oneninthfigurewidth\linewidth]{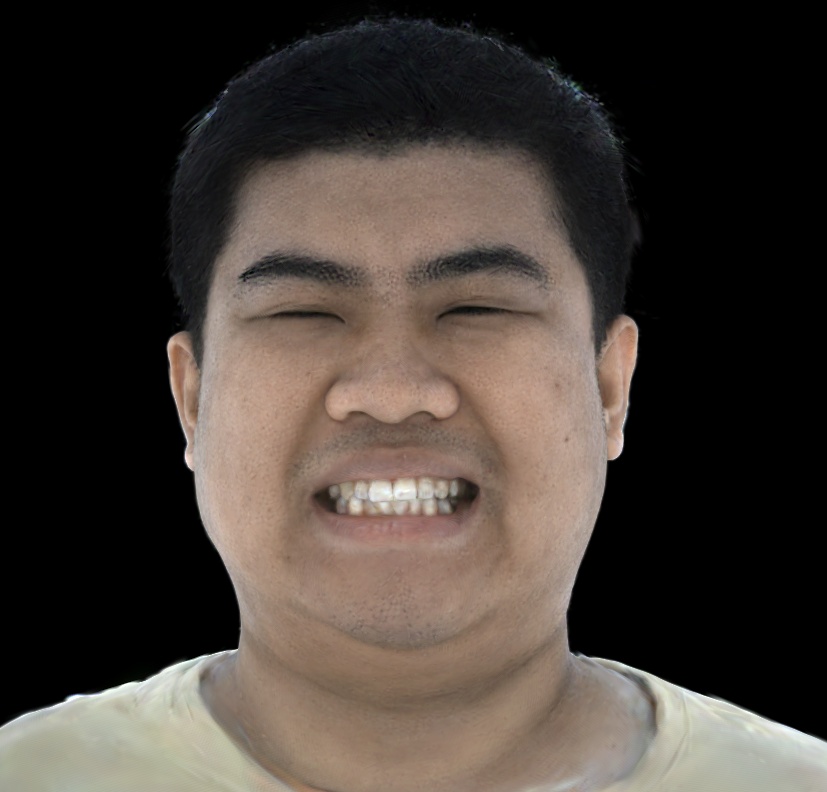}%
        \includegraphics[width=\oneninthfigurewidth\linewidth]{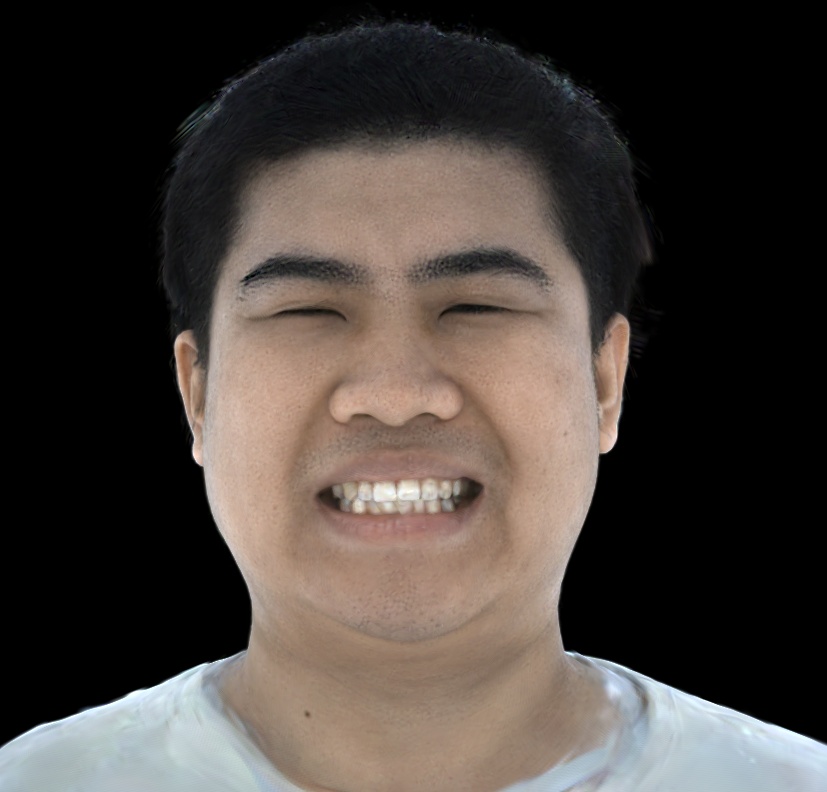}%
        \includegraphics[width=\oneninthfigurewidth\linewidth]{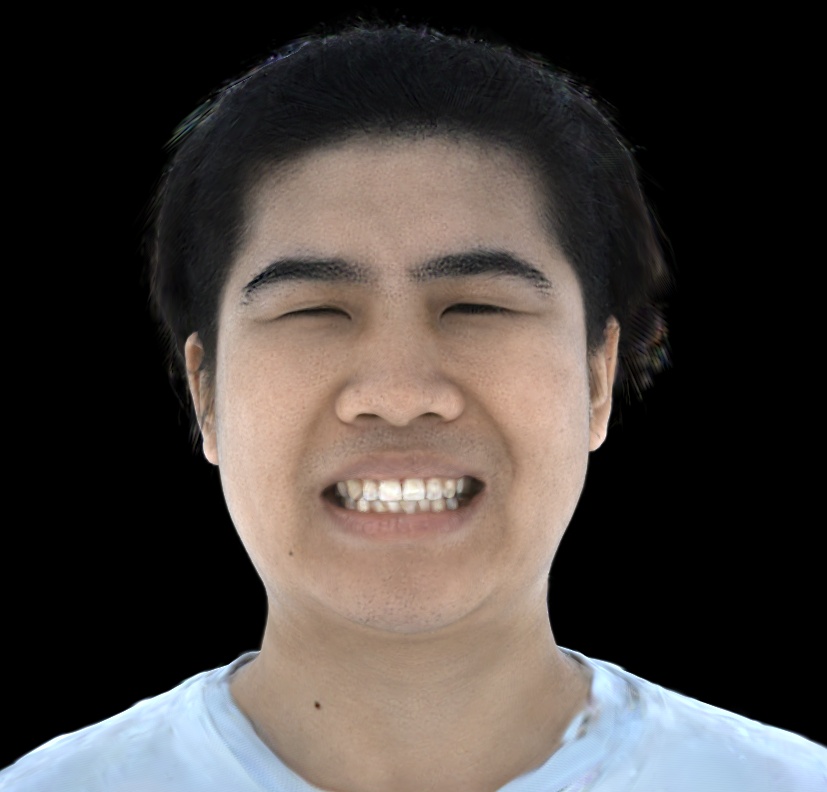}%
        \includegraphics[width=\oneninthfigurewidth\linewidth]{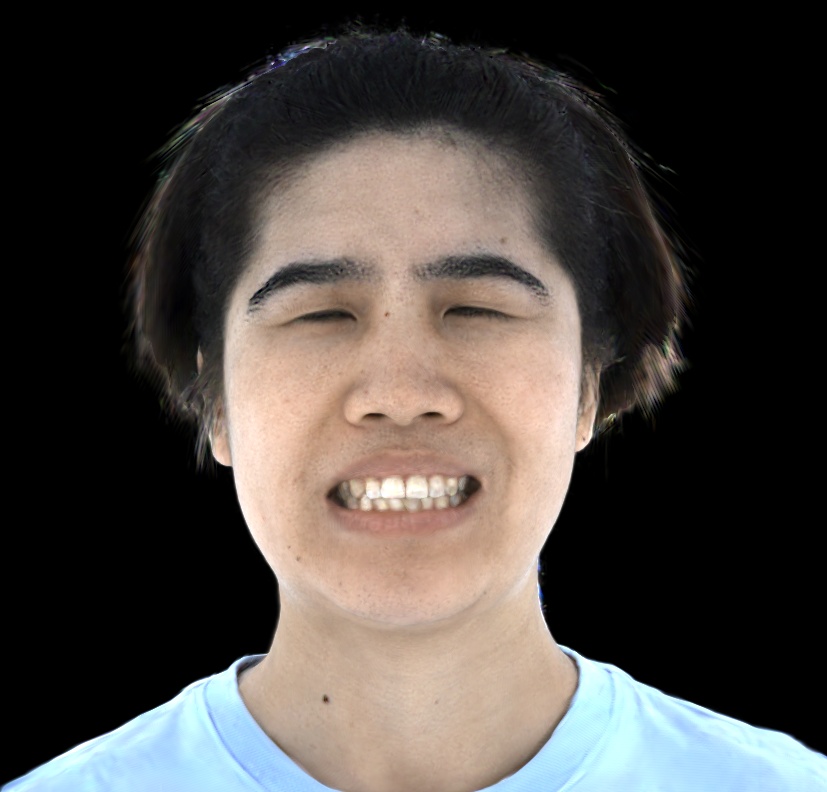}%
    \end{minipage}

    \begin{minipage}[t]{\linewidth}
        \centering
        \includegraphics[width=\oneninthfigurewidth\linewidth]{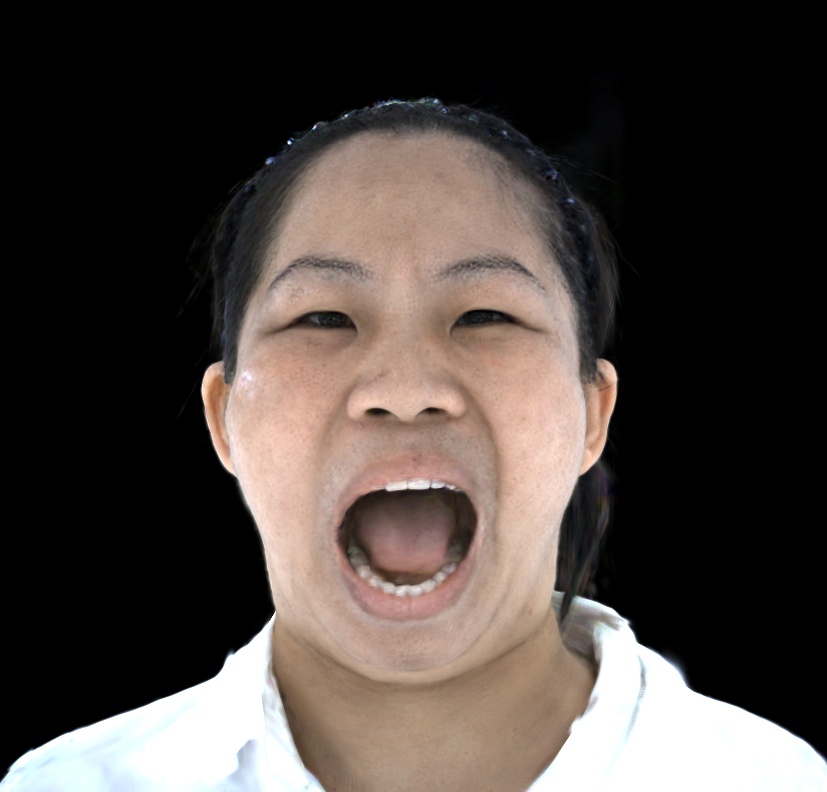}%
        \includegraphics[width=\oneninthfigurewidth\linewidth]{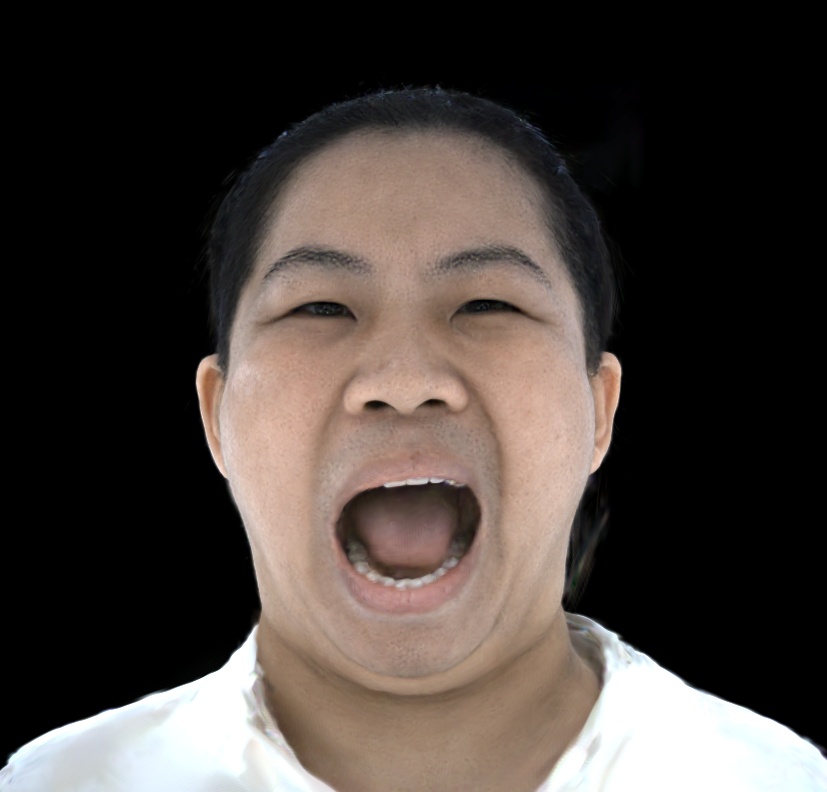}%
        \includegraphics[width=\oneninthfigurewidth\linewidth]{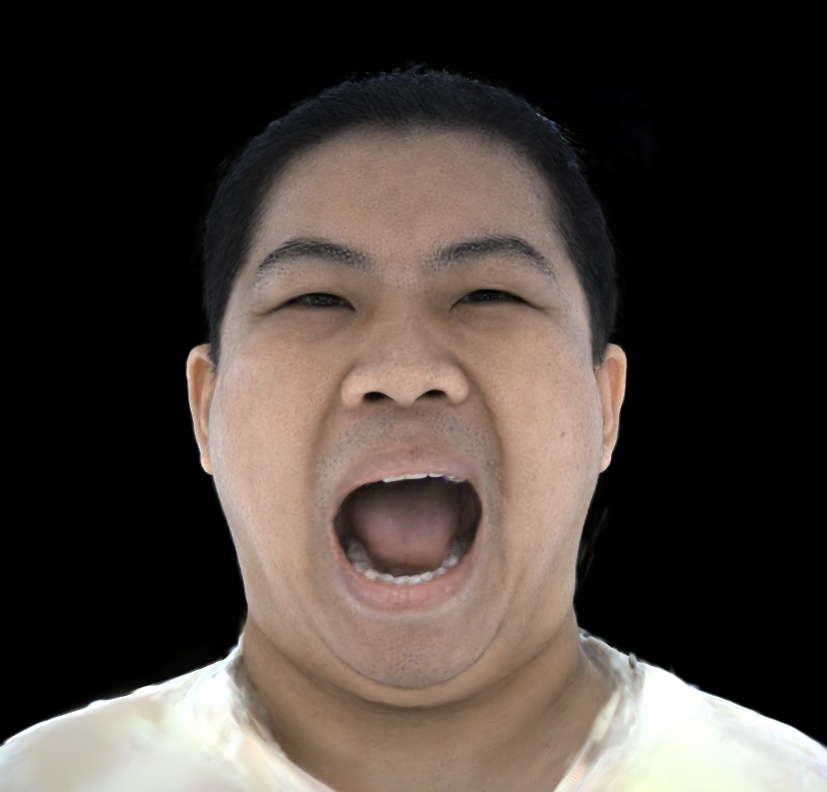}%
        \includegraphics[width=\oneninthfigurewidth\linewidth]{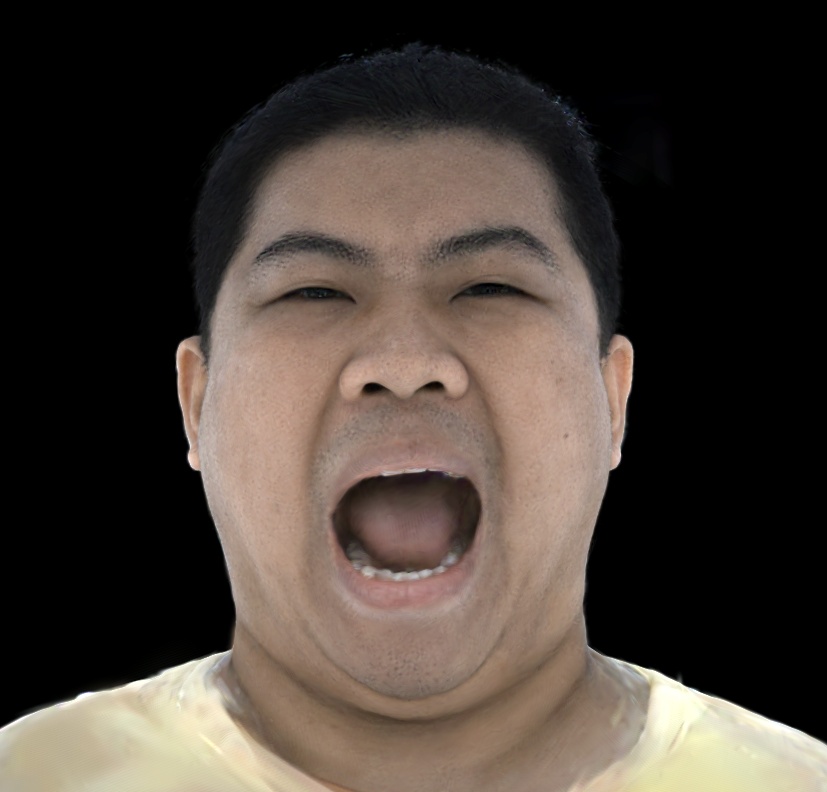}%
        \includegraphics[width=\oneninthfigurewidth\linewidth]{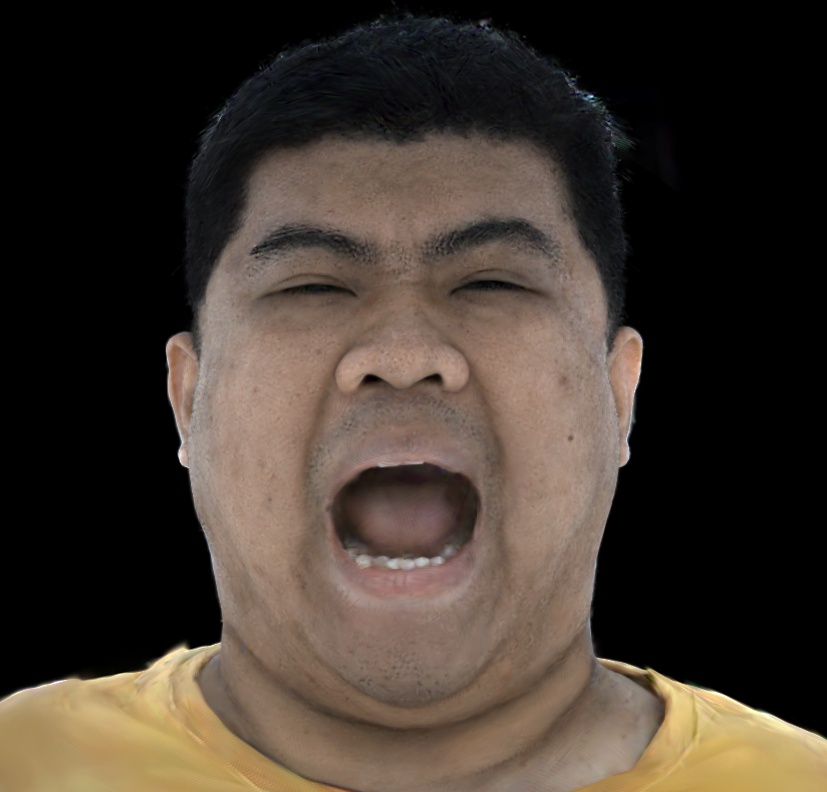}%
        \includegraphics[width=\oneninthfigurewidth\linewidth]{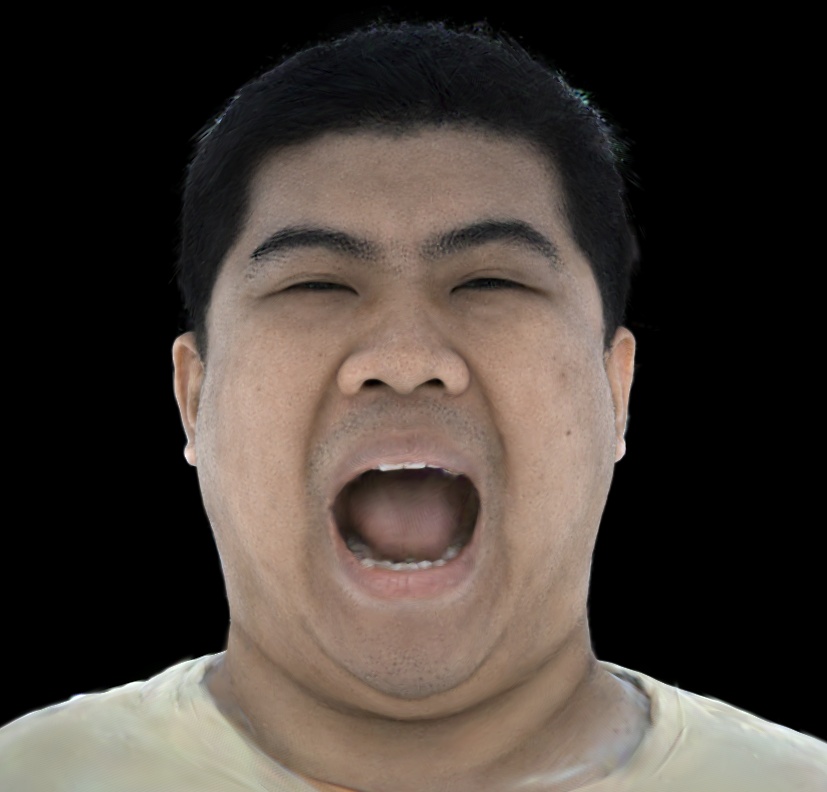}%
        \includegraphics[width=\oneninthfigurewidth\linewidth]{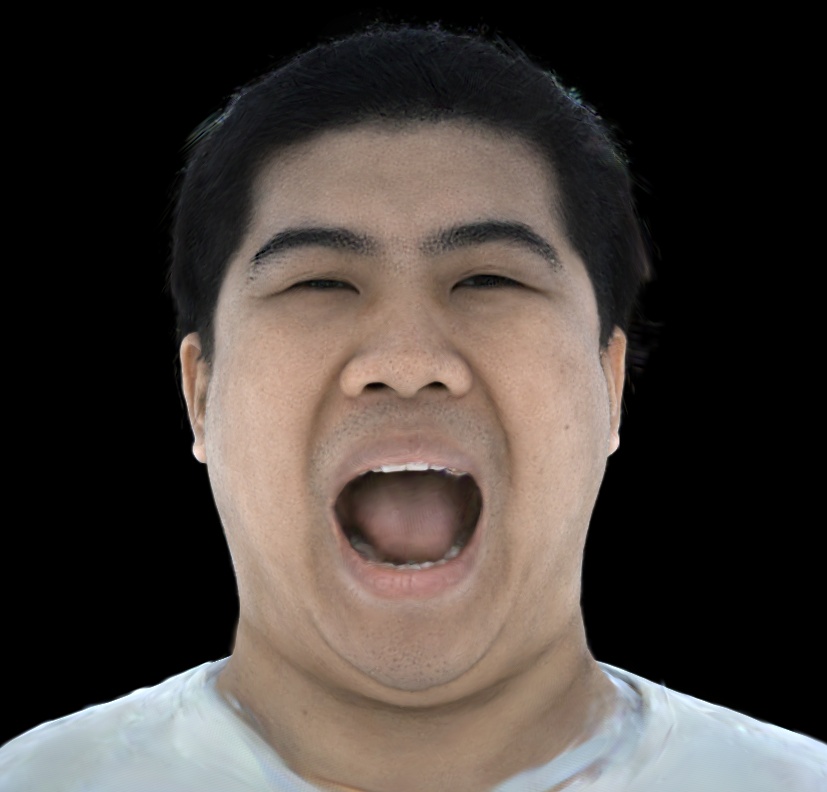}%
        \includegraphics[width=\oneninthfigurewidth\linewidth]{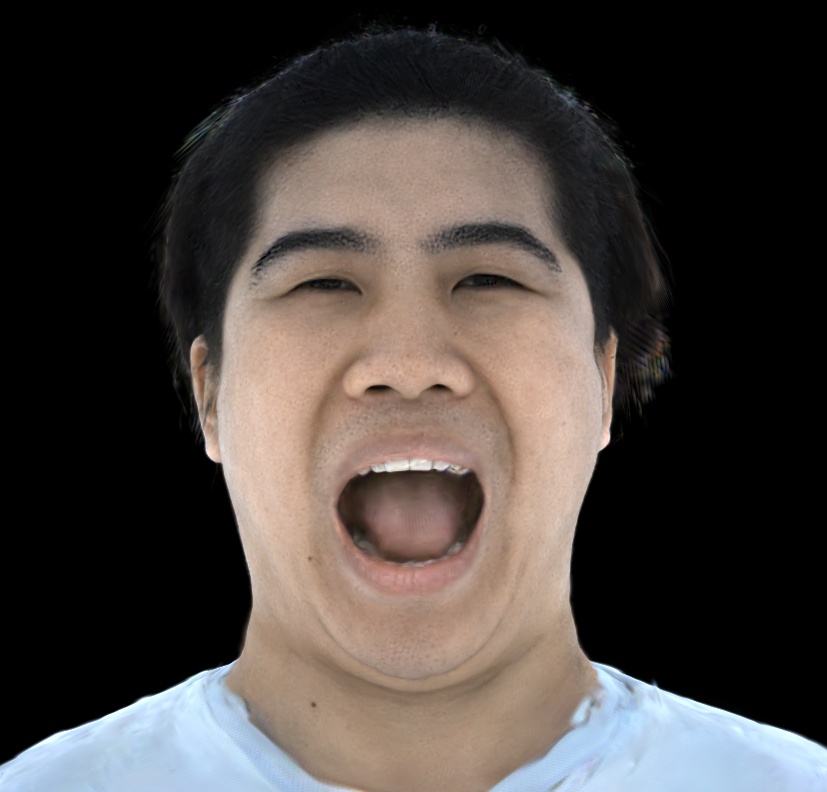}%
        \includegraphics[width=\oneninthfigurewidth\linewidth]{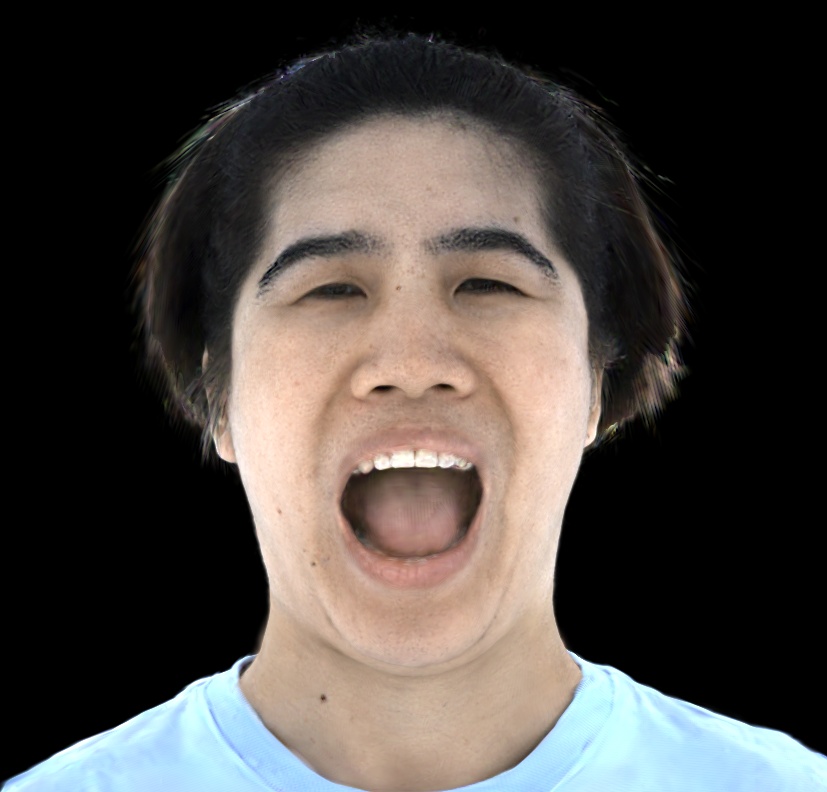}%
    \end{minipage}

%    \vspace{5pt}
    % \begin{minipage}{\linewidth}
    %     \centering
    %     \begin{minipage}[t]{\oneninthfigurewidth\linewidth}
    %         \centering
    %         \subfloat{NL}
    %     \end{minipage}
    %     \begin{minipage}[t]{\oneninthfigurewidth\linewidth}
    %         \centering
    %         \subfloat{NL + ENV}
    %     \end{minipage}
    %     \begin{minipage}[t]{\oneninthfigurewidth\linewidth}
    %         \centering
    %         \subfloat{NL + LCL}
    %     \end{minipage}
    %     \begin{minipage}[t]{\oneninthfigurewidth\linewidth}
    %         \centering
    %         \subfloat{NL}
    %     \end{minipage}
    %     \begin{minipage}[t]{\oneninthfigurewidth\linewidth}
    %         \centering
    %         \subfloat{NL + ENV}
    %     \end{minipage}
    %     \begin{minipage}[t]{\oneninthfigurewidth\linewidth}
    %         \centering
    %         \subfloat{NL + LCL}
    %     \end{minipage}
    %     \begin{minipage}[t]{\oneninthfigurewidth\linewidth}
    %         \centering
    %         \subfloat{NL + TS}
    %     \end{minipage}
    %     \begin{minipage}[t]{\oneninthfigurewidth\linewidth}
    %         \centering
    %         \subfloat{Ours}
    %     \end{minipage}
    %     \begin{minipage}[t]{\oneninthfigurewidth\linewidth}
    %         \centering
    %         \subfloat{Ground truth}
    %     \end{minipage}
    % \end{minipage}
    \end{minipage}
    \caption{Interpolation results between three identities (left, center, right). Our model learns a smooth identity latent space that allows linear interpolation. Besides, the expression keeps unchanged during the interpolation, confirming that the expression and identity spaces have been effectively disentangled. 
    }
\label{fig:interpolation}
\end{figure*}

\begin{figure*}%[t]
    \centering
    \begin{minipage}[t]{\linewidth}
    \begin{minipage}[t]{\linewidth}
        \centering
        \includegraphics[width=\onetenthfigurewidth\linewidth]{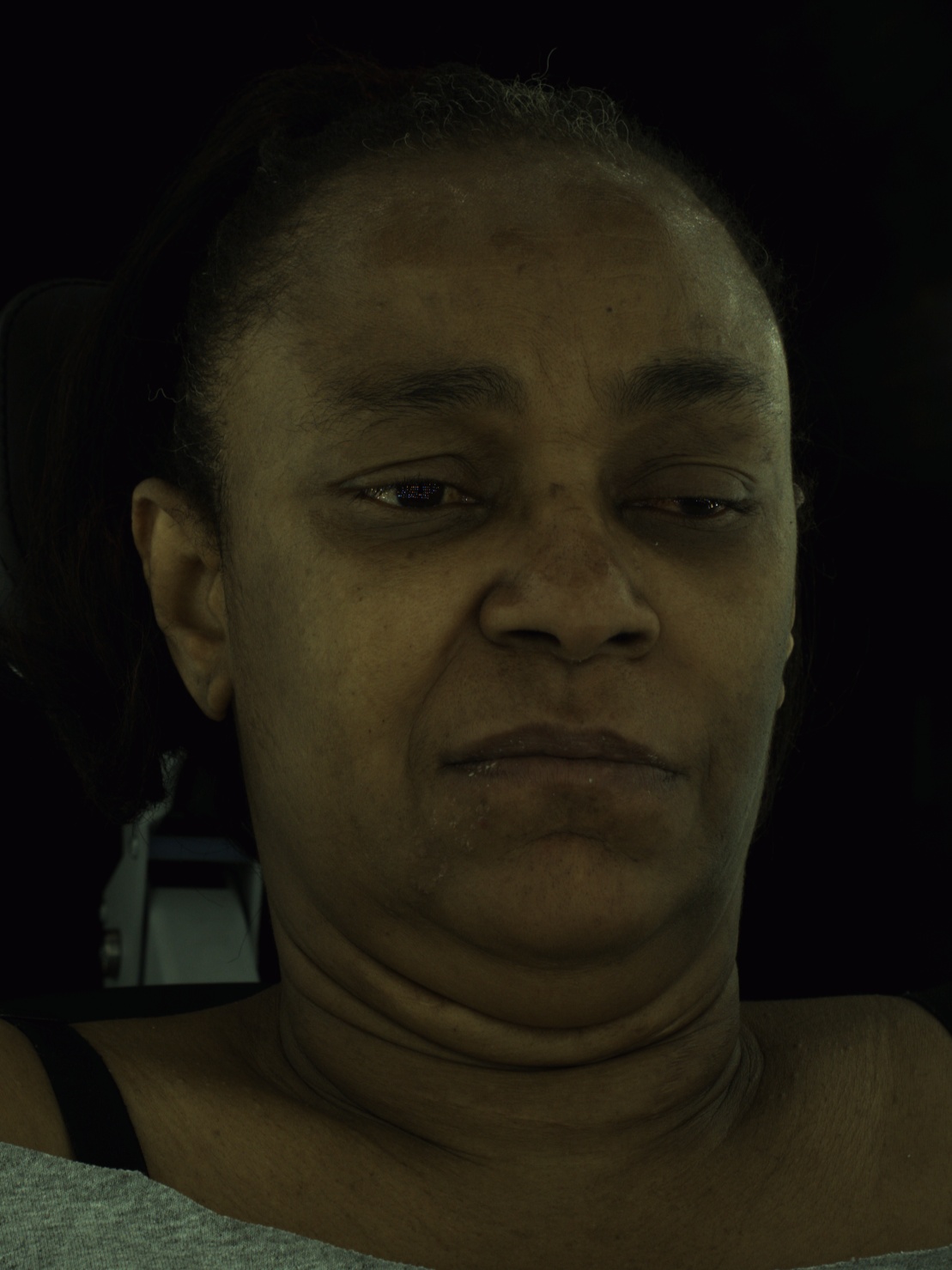}
        \includegraphics[width=\onetenthfigurewidth\linewidth]{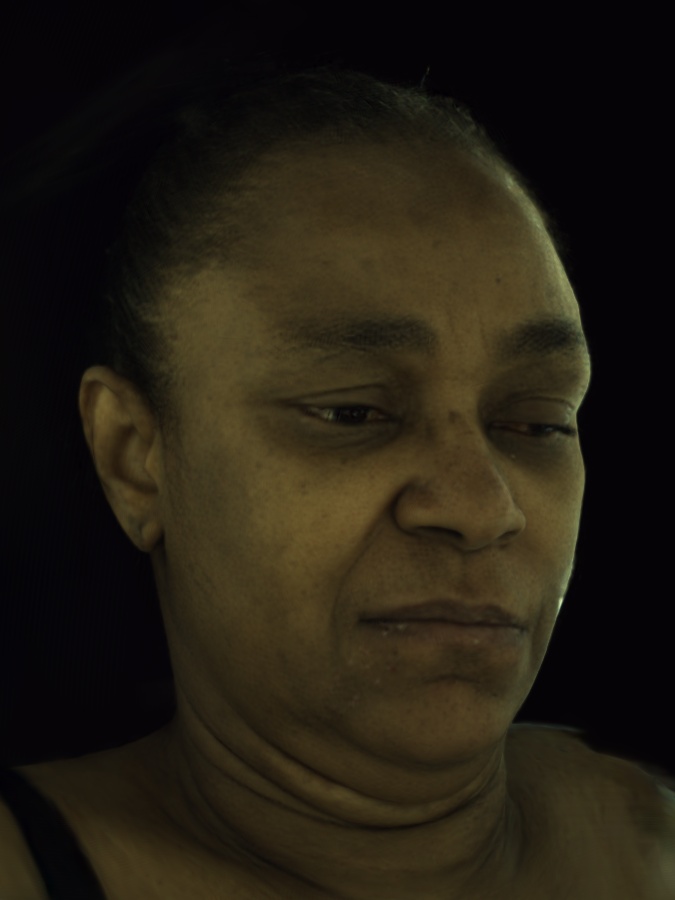}%
        \includegraphics[width=\onetenthfigurewidth\linewidth]{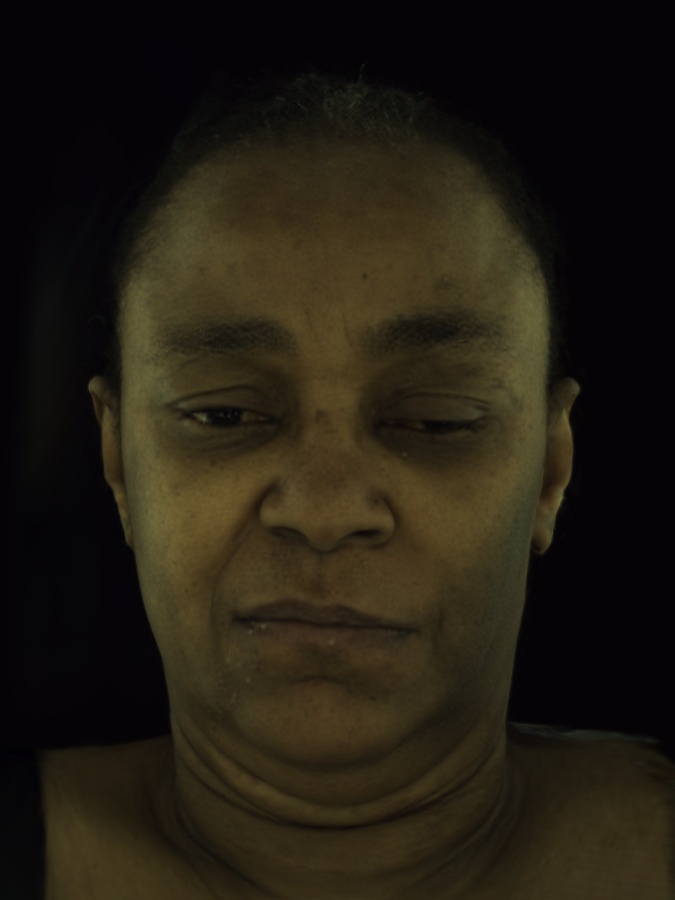}%
        \includegraphics[width=\onetenthfigurewidth\linewidth]{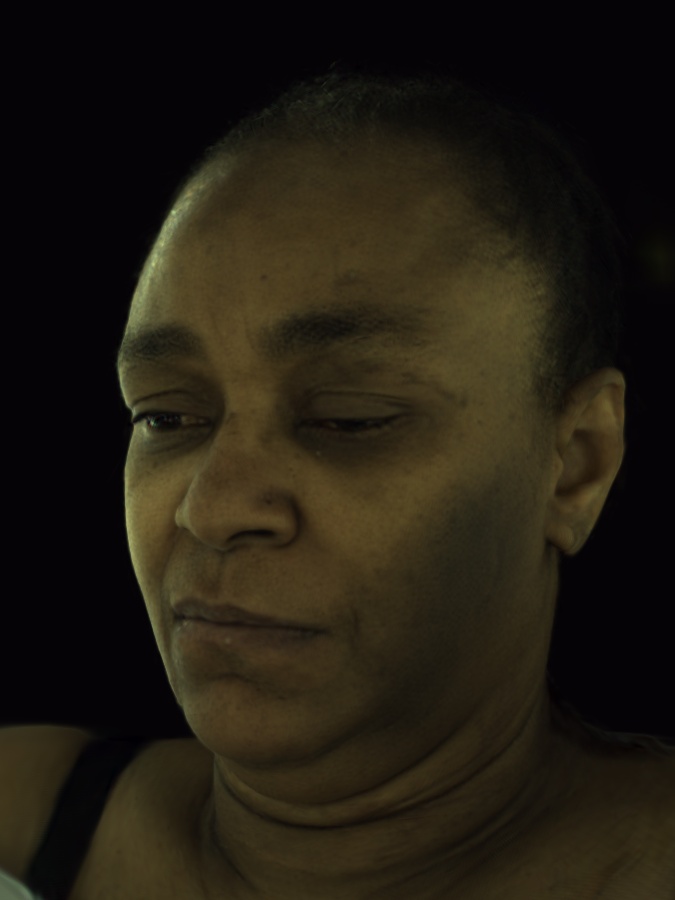}
        \includegraphics[width=\onetenthfigurewidth\linewidth]{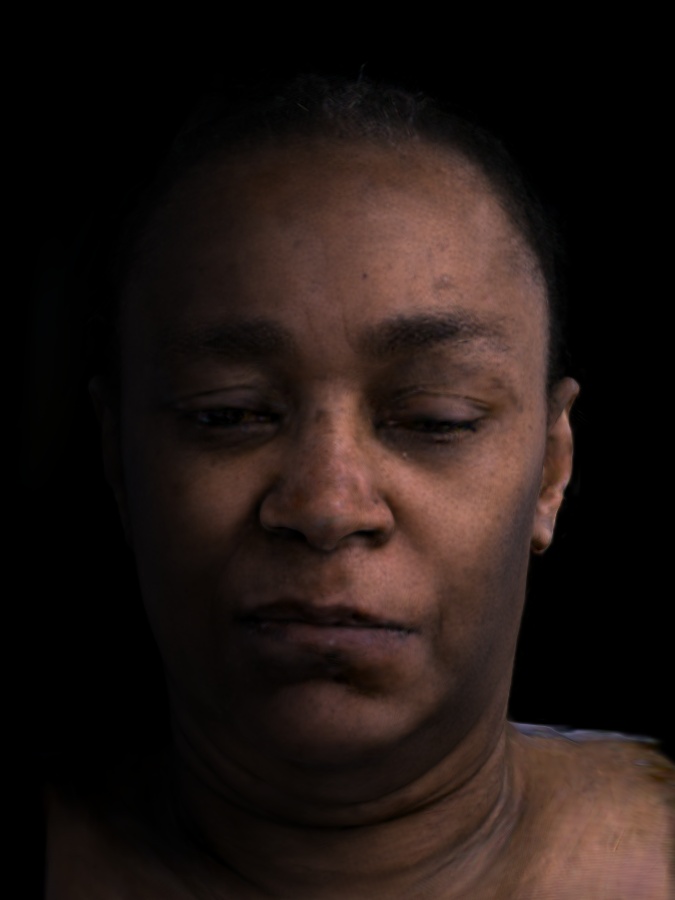}%
        \includegraphics[width=\onetenthfigurewidth\linewidth]{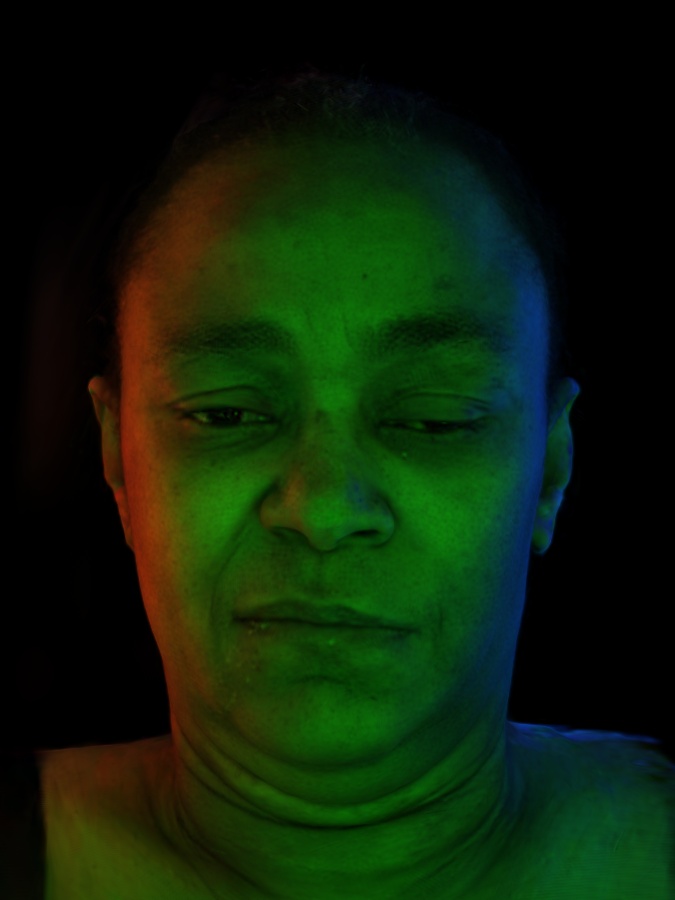}%
        \includegraphics[width=\onetenthfigurewidth\linewidth]{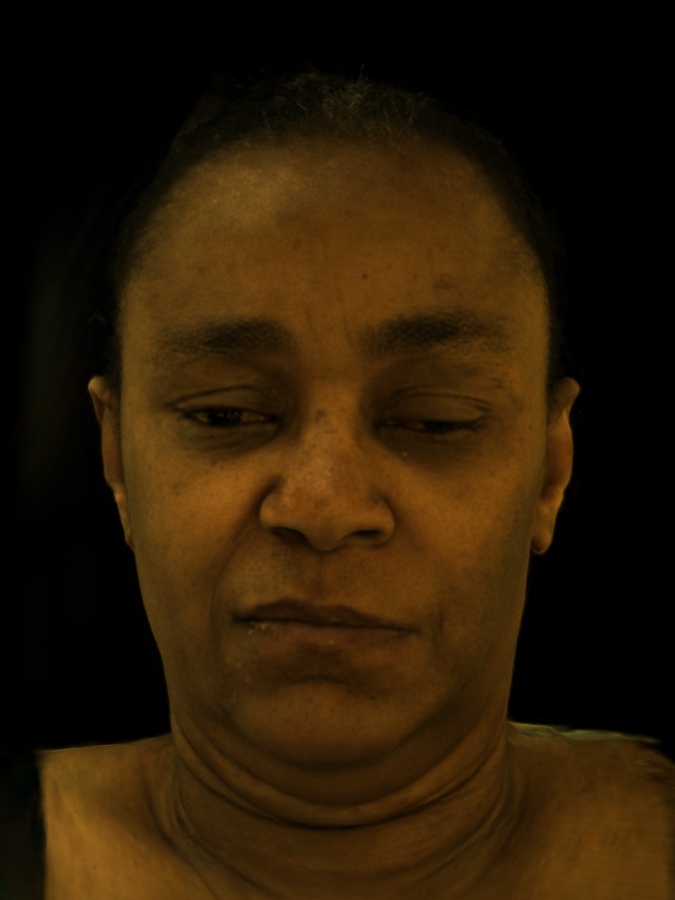}
        \includegraphics[width=\onetenthfigurewidth\linewidth]{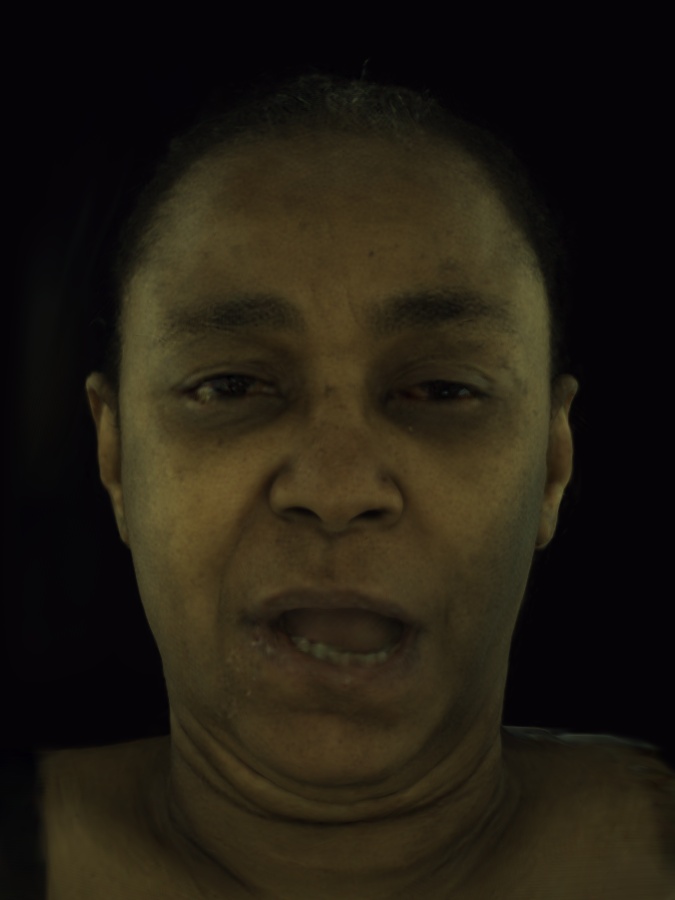}%
        \includegraphics[width=\onetenthfigurewidth\linewidth]{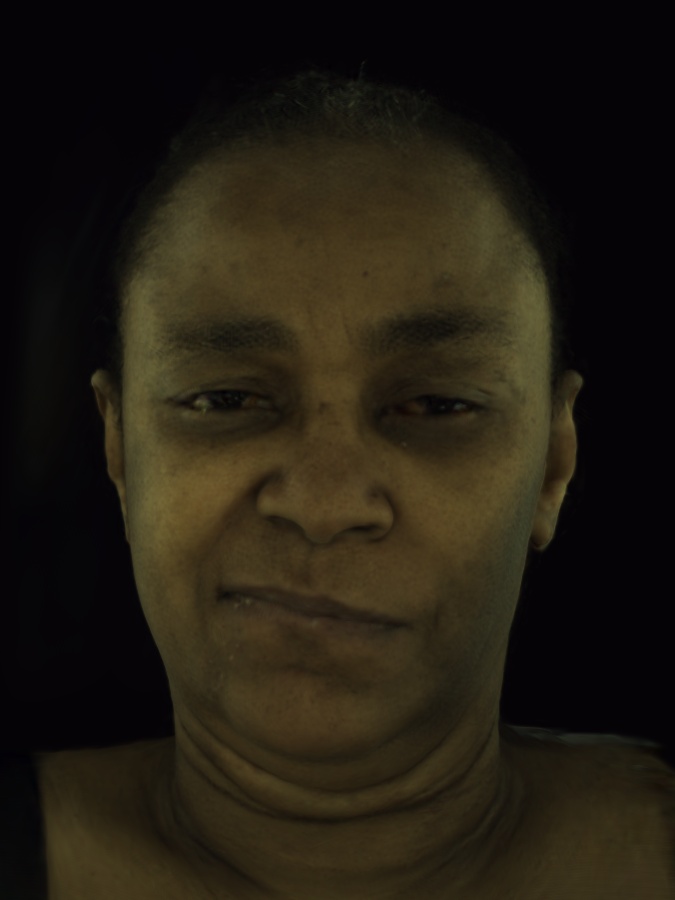}%
        \includegraphics[width=\onetenthfigurewidth\linewidth]{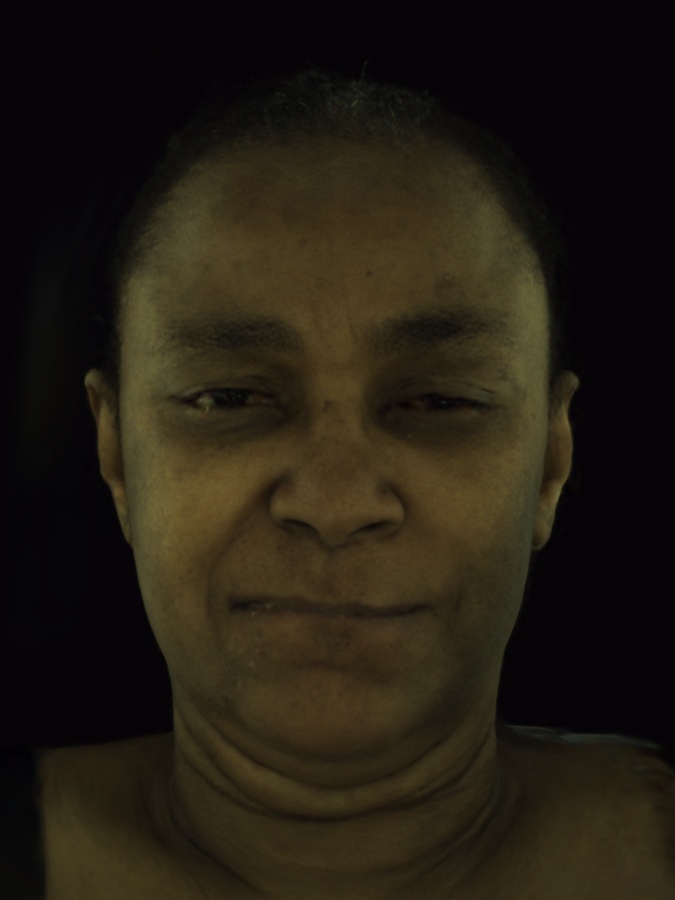}
    \end{minipage}

    \begin{minipage}[t]{\linewidth}
        \centering
        \includegraphics[width=\onetenthfigurewidth\linewidth]{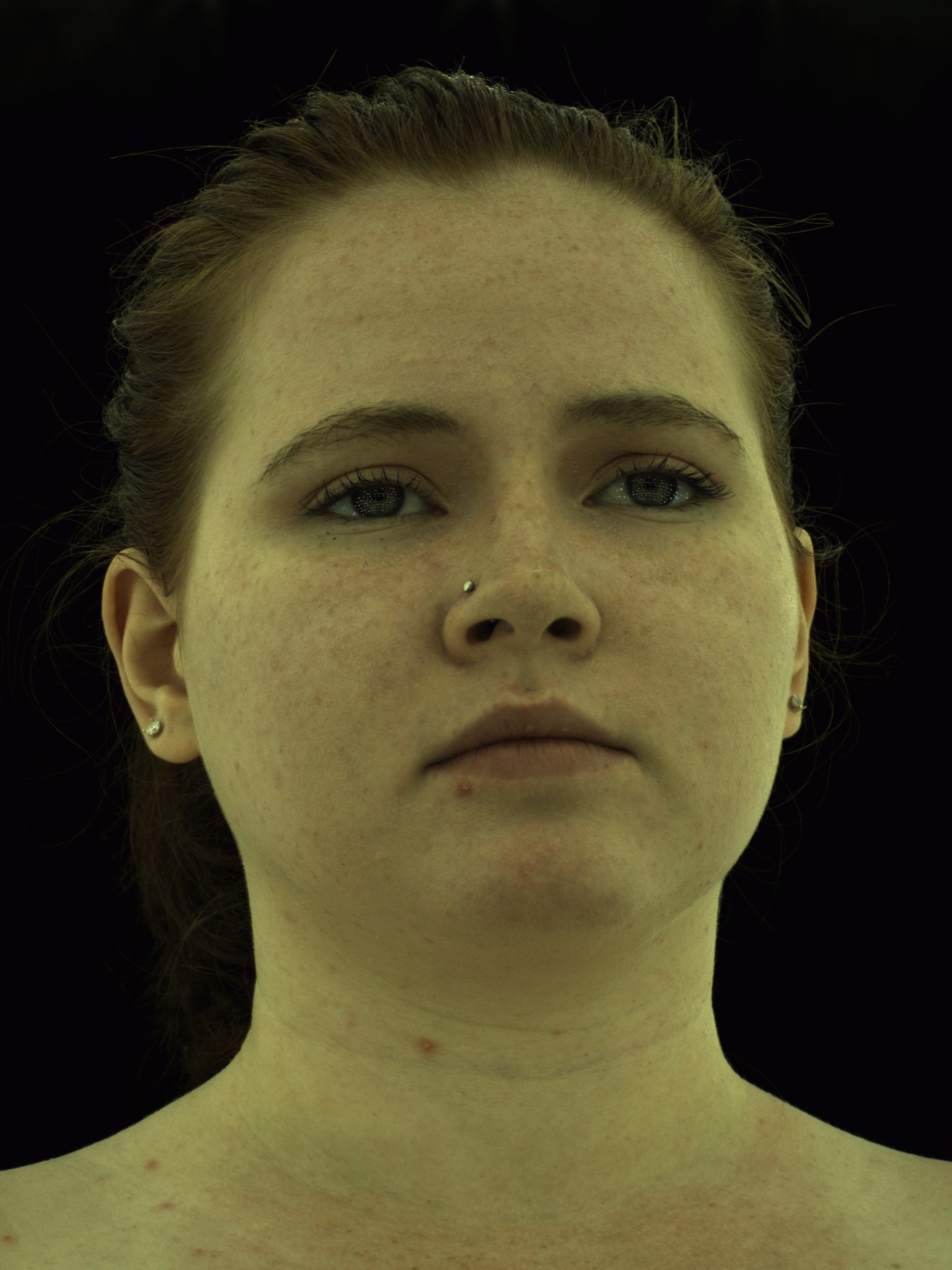}
        \includegraphics[width=\onetenthfigurewidth\linewidth]{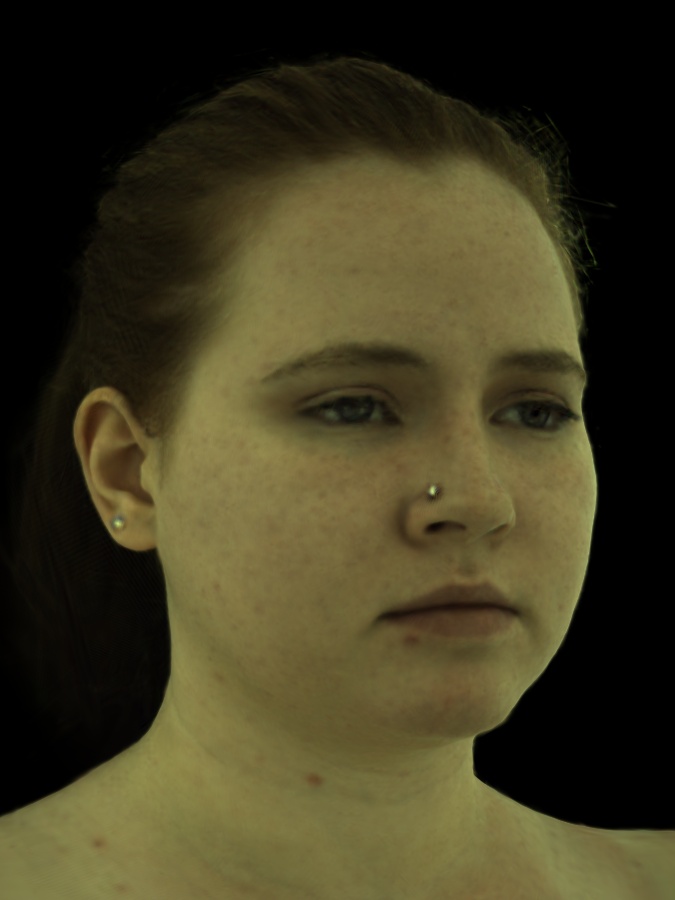}%
        \includegraphics[width=\onetenthfigurewidth\linewidth]{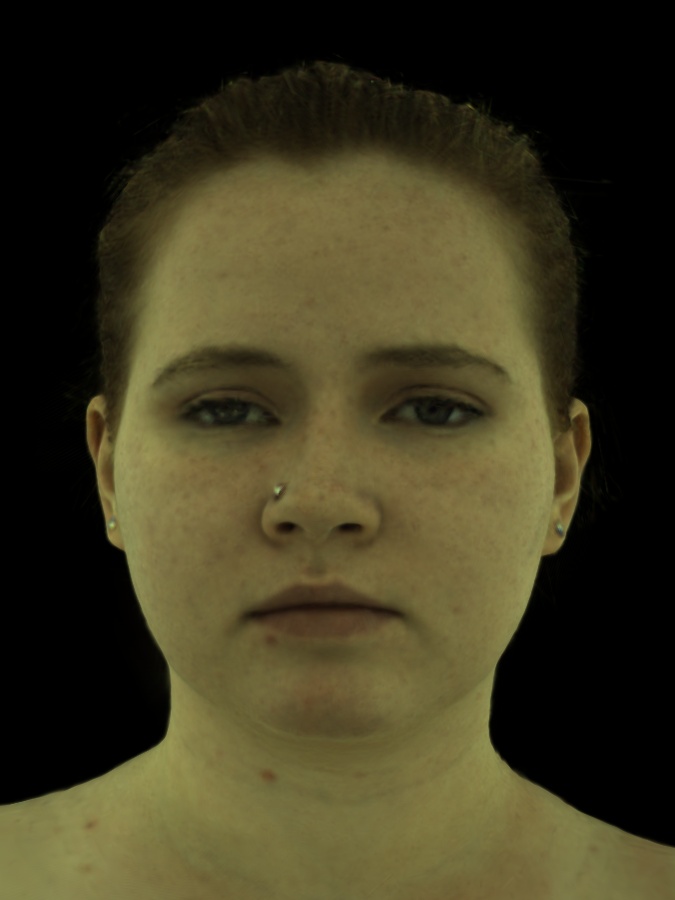}%
        \includegraphics[width=\onetenthfigurewidth\linewidth]{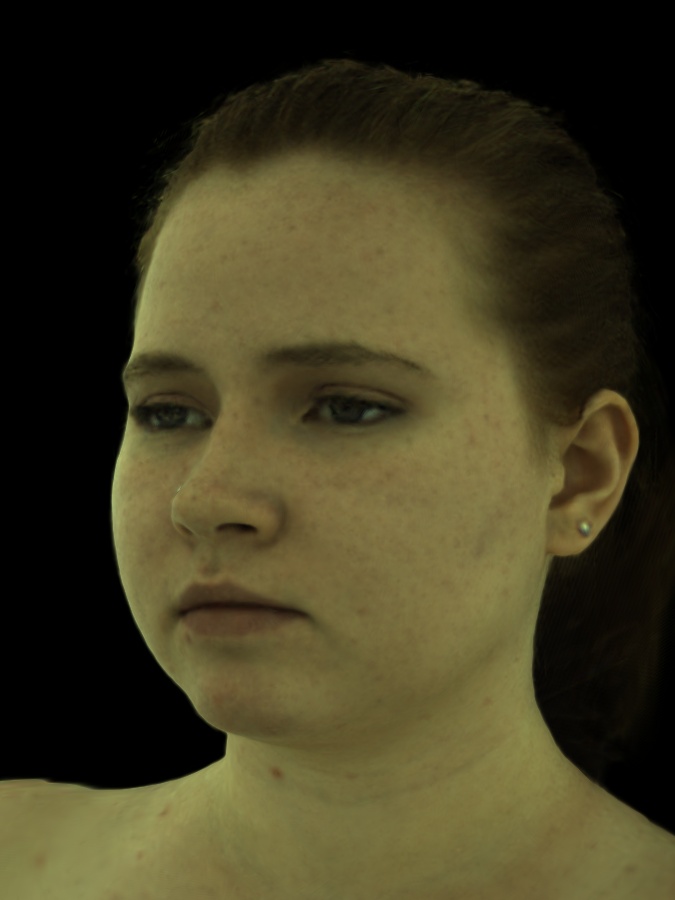}
        \includegraphics[width=\onetenthfigurewidth\linewidth]{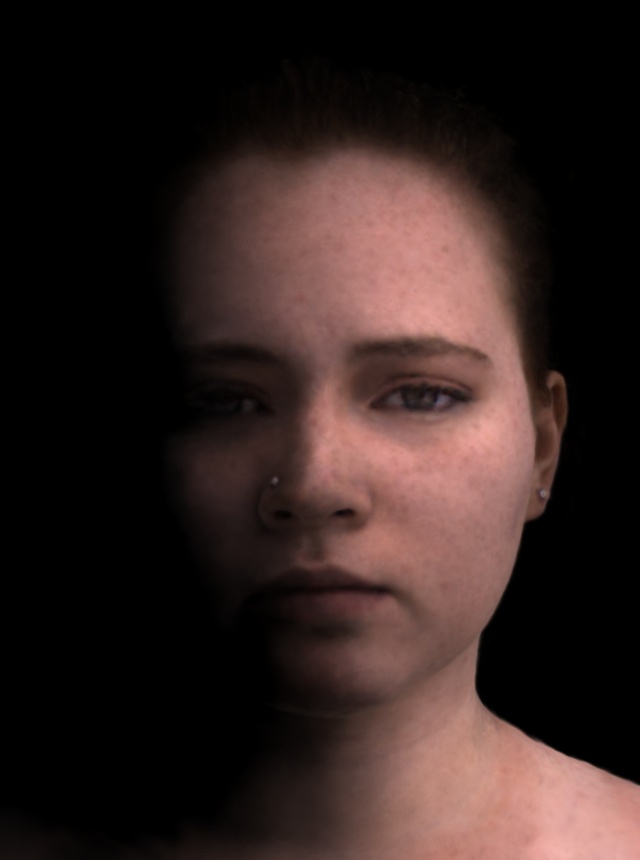}%
        \includegraphics[width=\onetenthfigurewidth\linewidth]{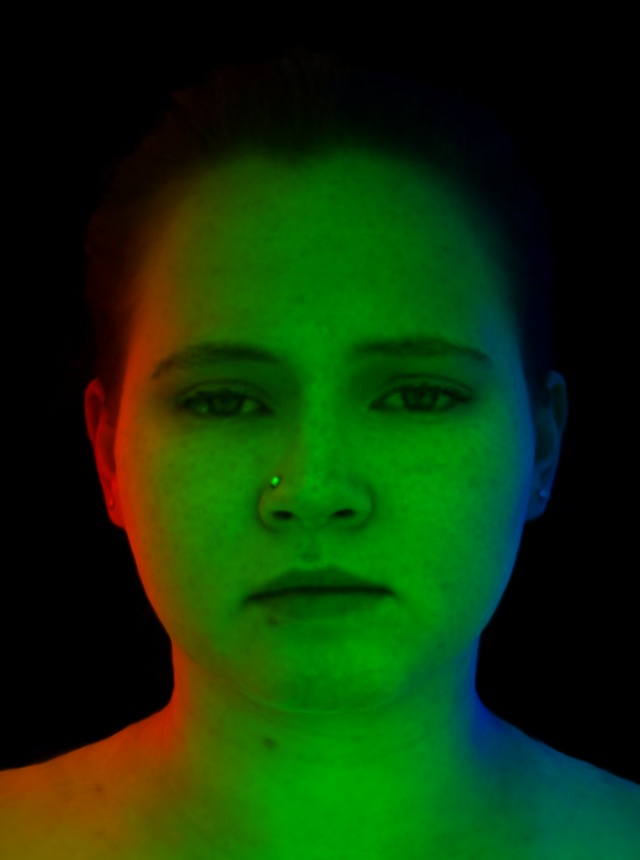}%
        \includegraphics[width=\onetenthfigurewidth\linewidth]{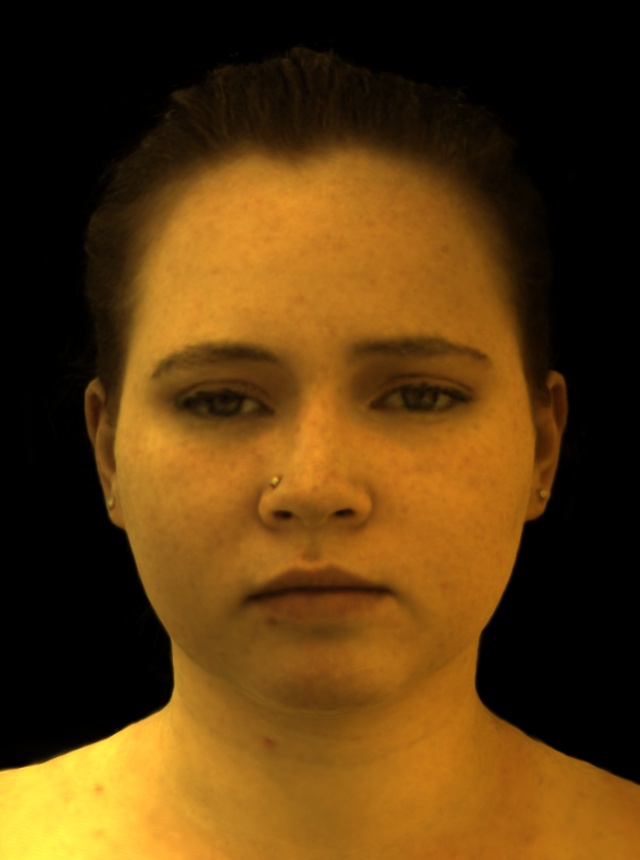}
        \includegraphics[width=\onetenthfigurewidth\linewidth]{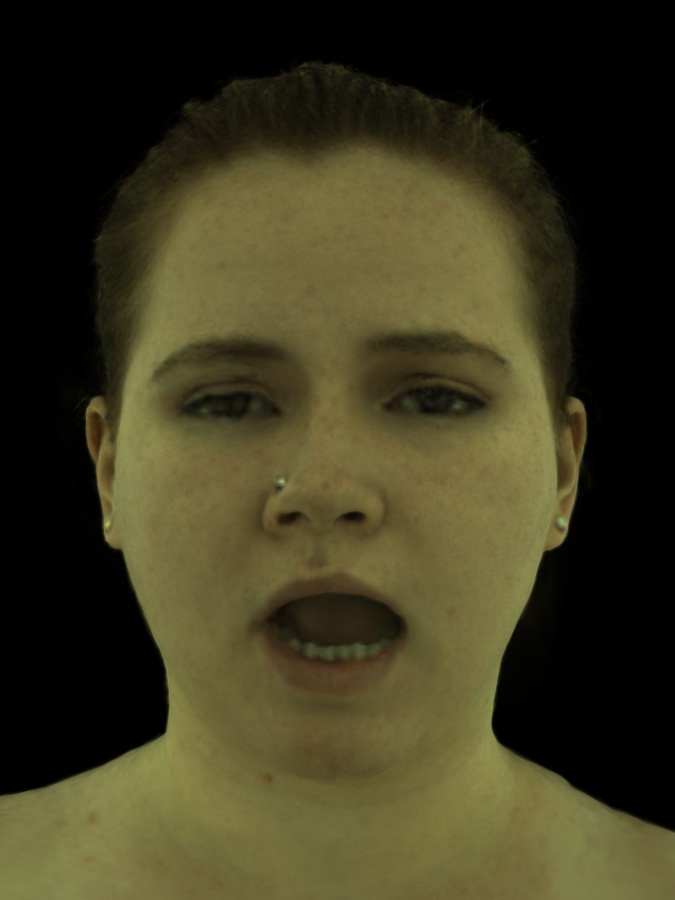}%
        \includegraphics[width=\onetenthfigurewidth\linewidth]{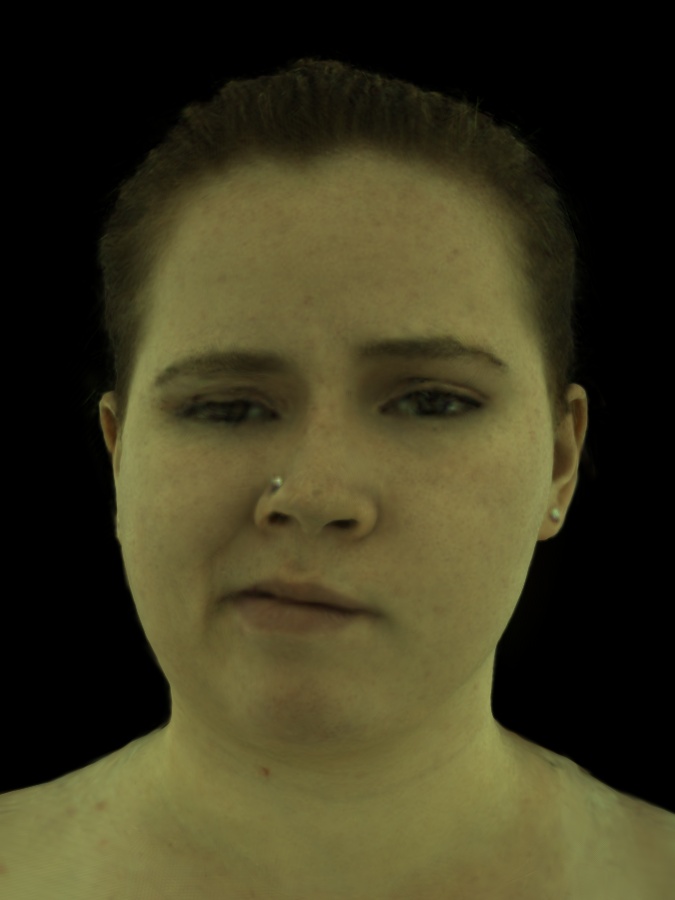}%
        \includegraphics[width=\onetenthfigurewidth\linewidth]{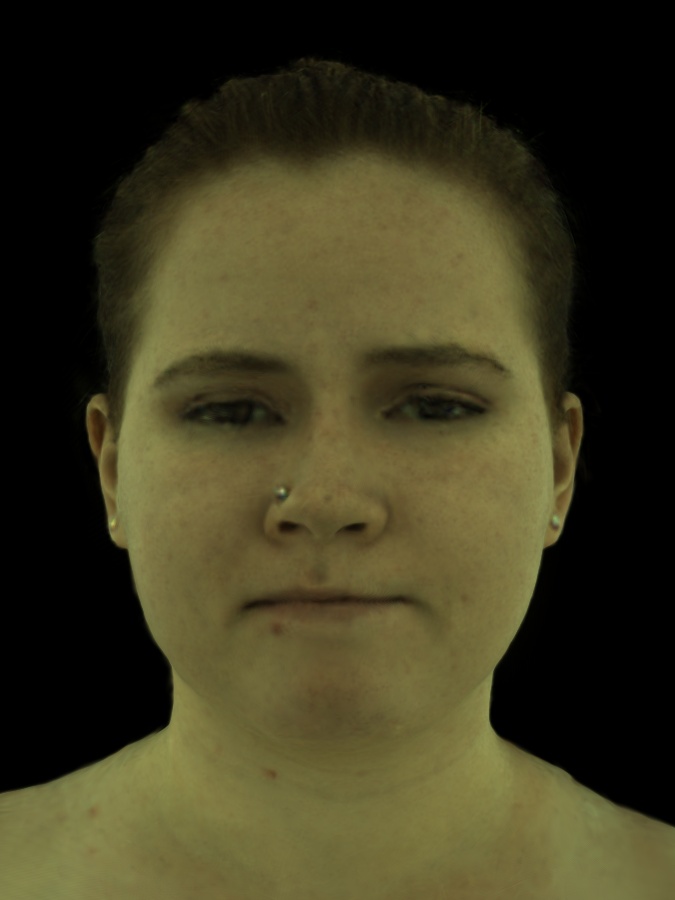}
    \end{minipage}

    \begin{minipage}[t]{\linewidth}
        \centering
        \includegraphics[width=\onetenthfigurewidth\linewidth]{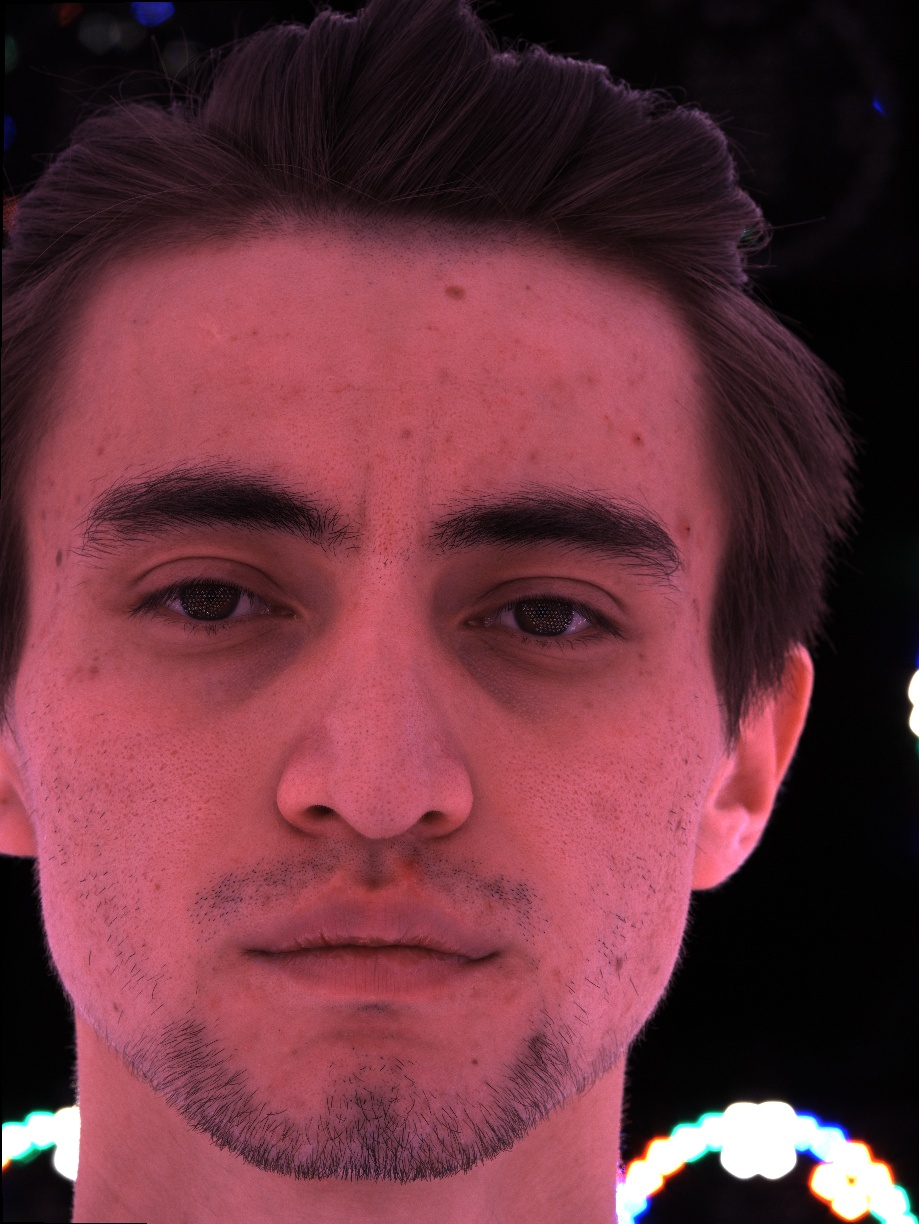}
        \includegraphics[width=\onetenthfigurewidth\linewidth]{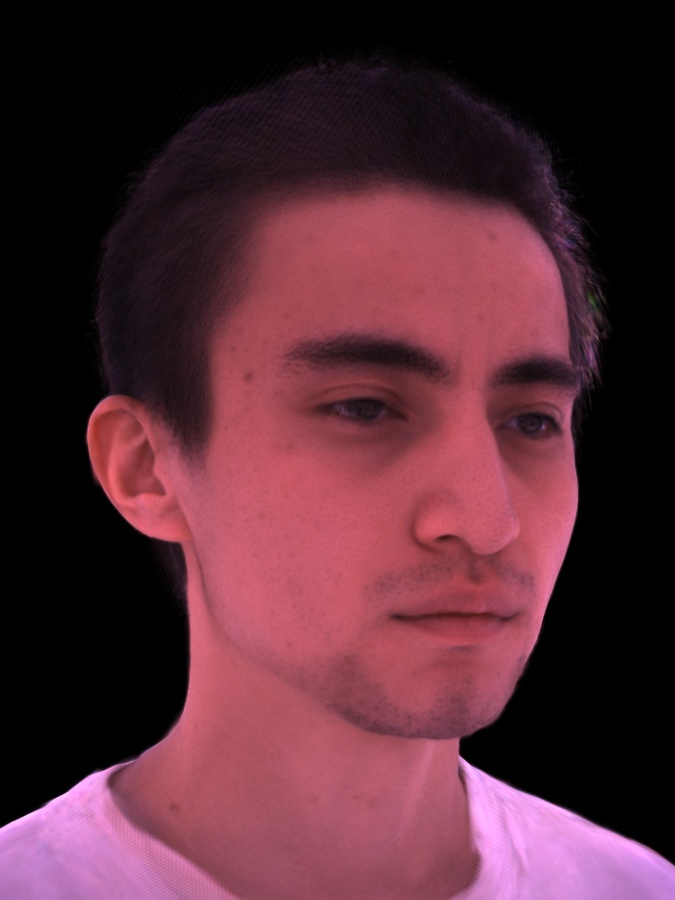}%
        \includegraphics[width=\onetenthfigurewidth\linewidth]{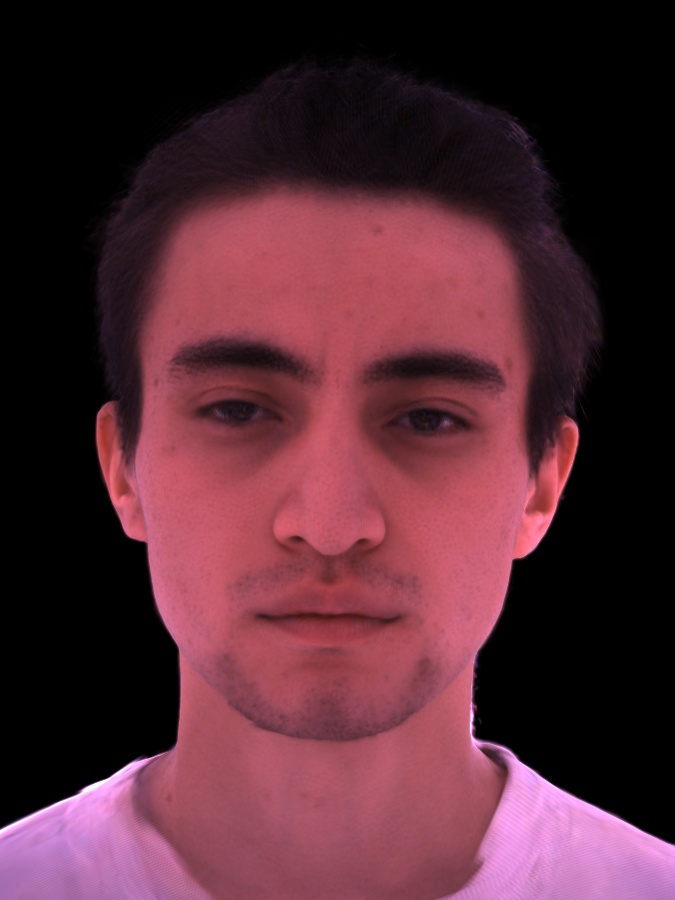}%
        \includegraphics[width=\onetenthfigurewidth\linewidth]{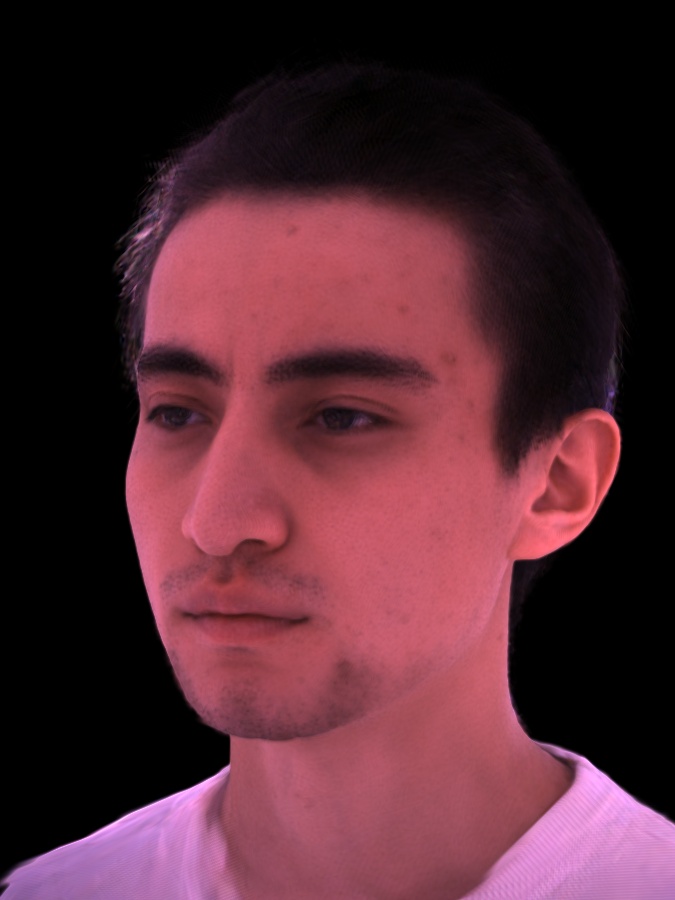}
        \includegraphics[width=\onetenthfigurewidth\linewidth]{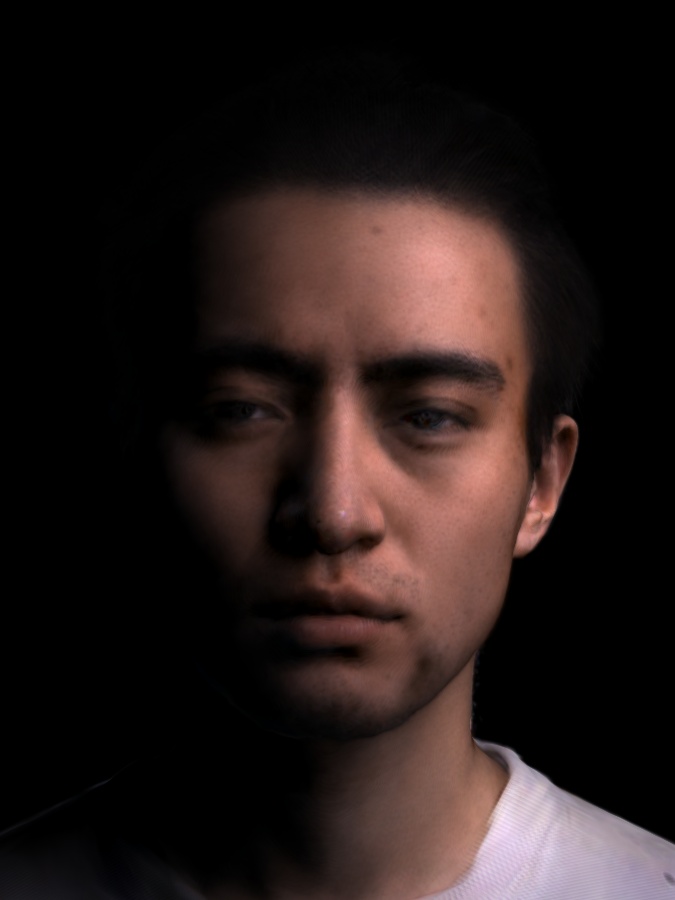}%
        \includegraphics[width=\onetenthfigurewidth\linewidth]{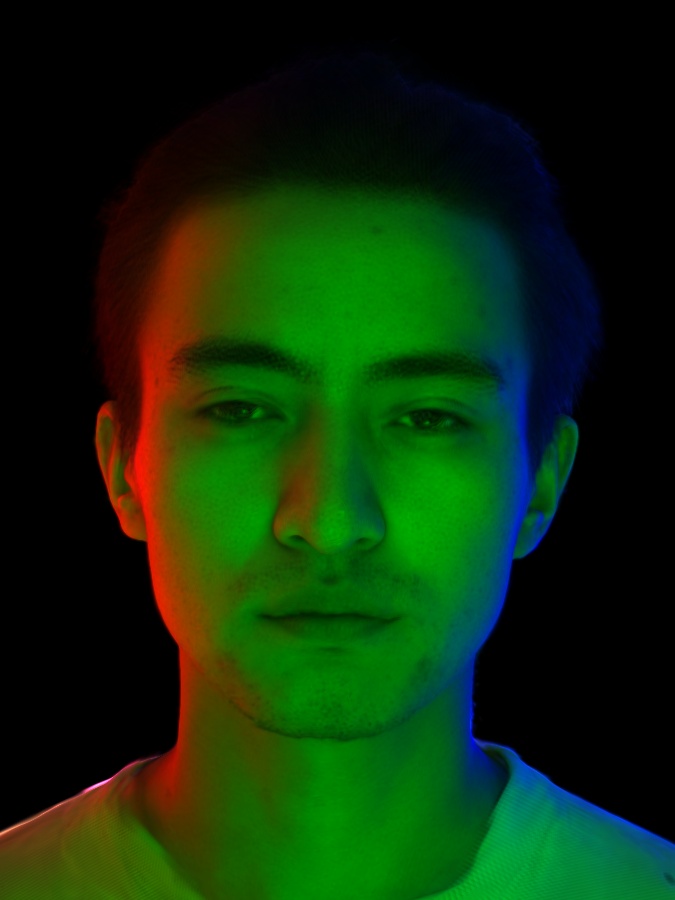}%
        \includegraphics[width=\onetenthfigurewidth\linewidth]{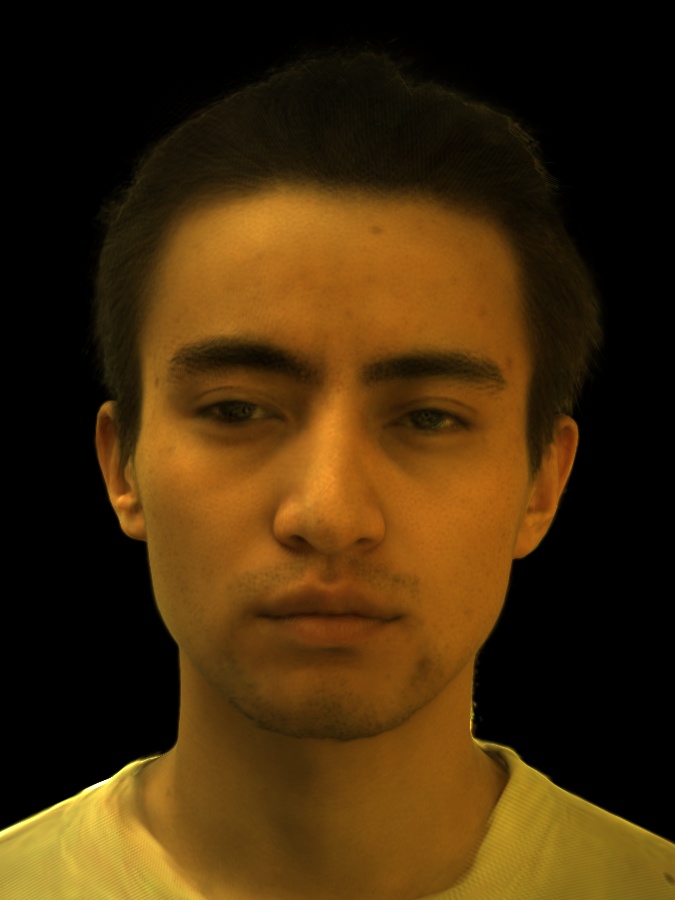}
        \includegraphics[width=\onetenthfigurewidth\linewidth]{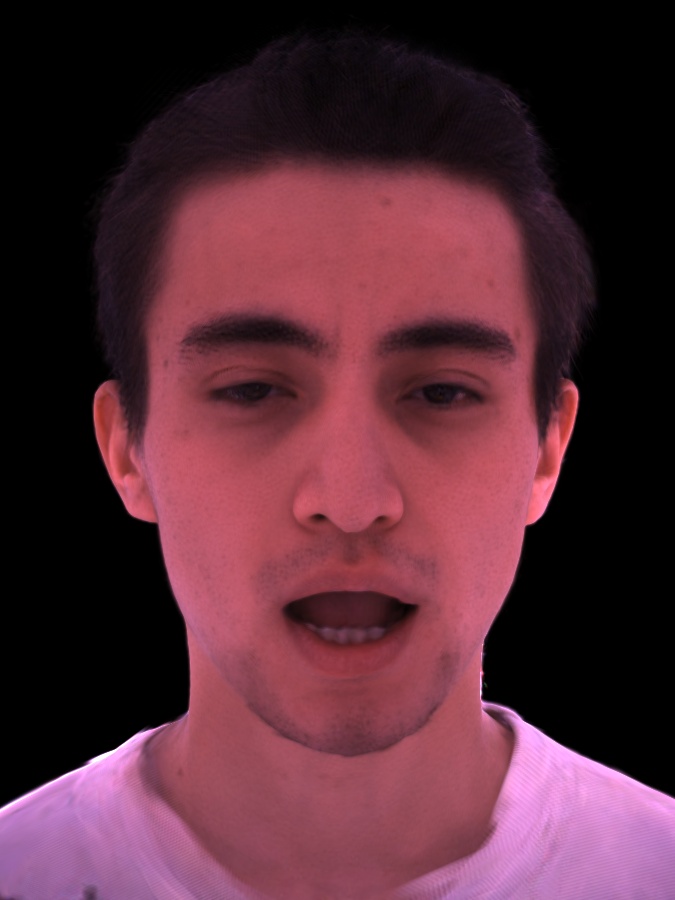}%
        \includegraphics[width=\onetenthfigurewidth\linewidth]{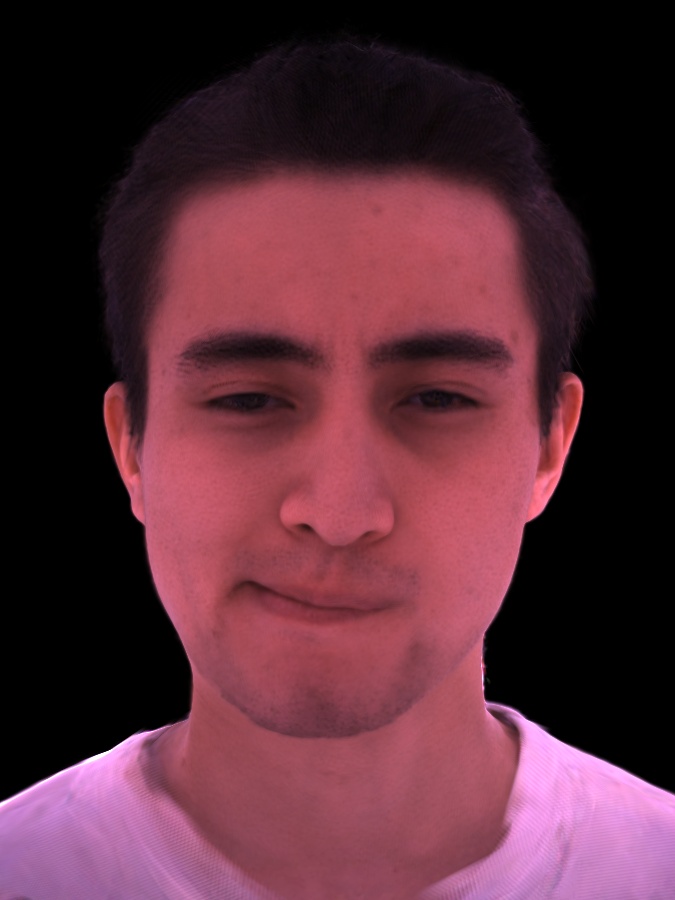}%
        \includegraphics[width=\onetenthfigurewidth\linewidth]{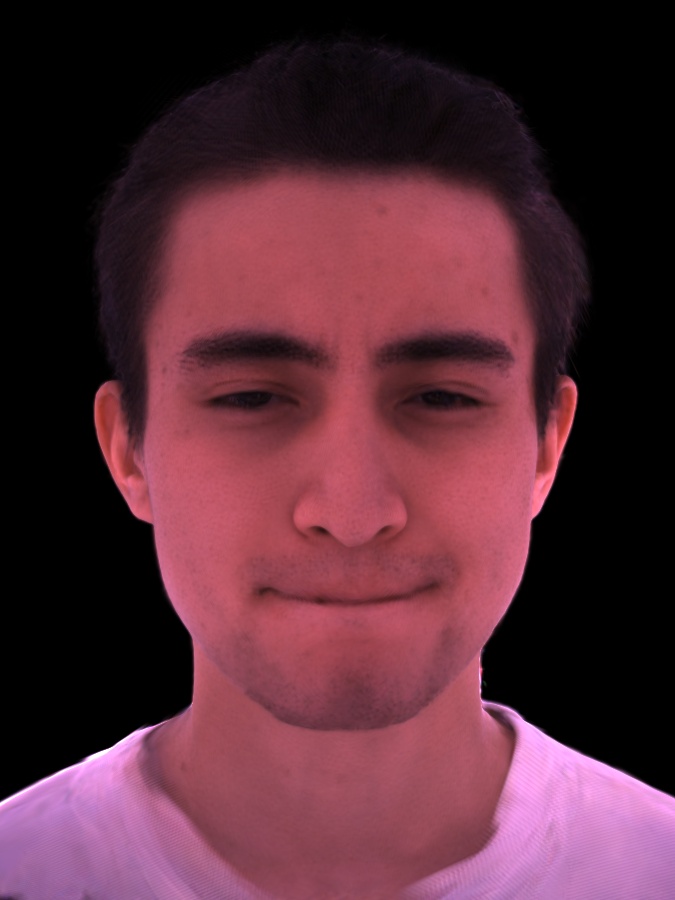}
    \end{minipage}

    \begin{minipage}[t]{\linewidth}
        \centering
        \includegraphics[width=\onetenthfigurewidth\linewidth]{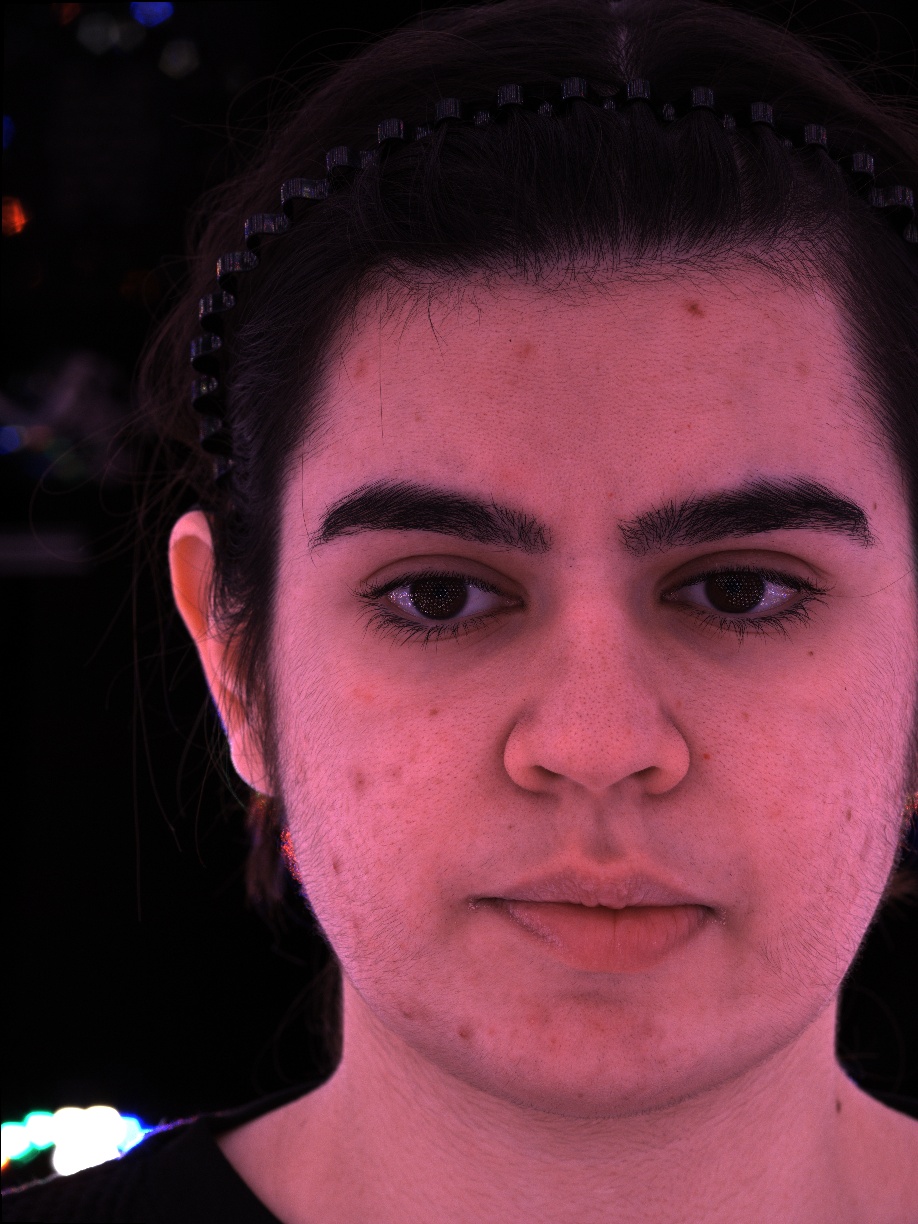}
        \includegraphics[width=\onetenthfigurewidth\linewidth]{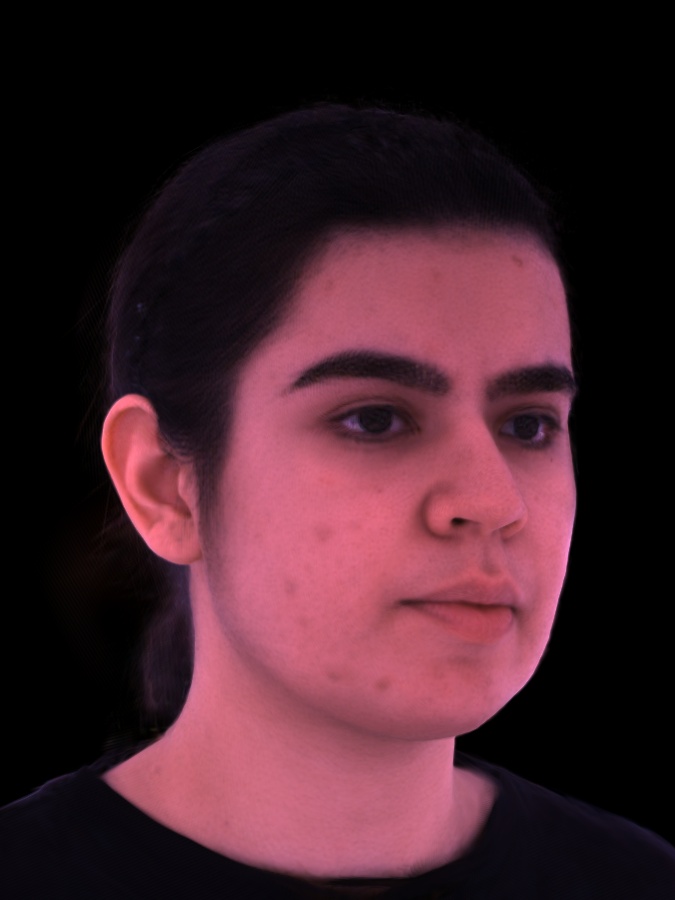}%
        \includegraphics[width=\onetenthfigurewidth\linewidth]{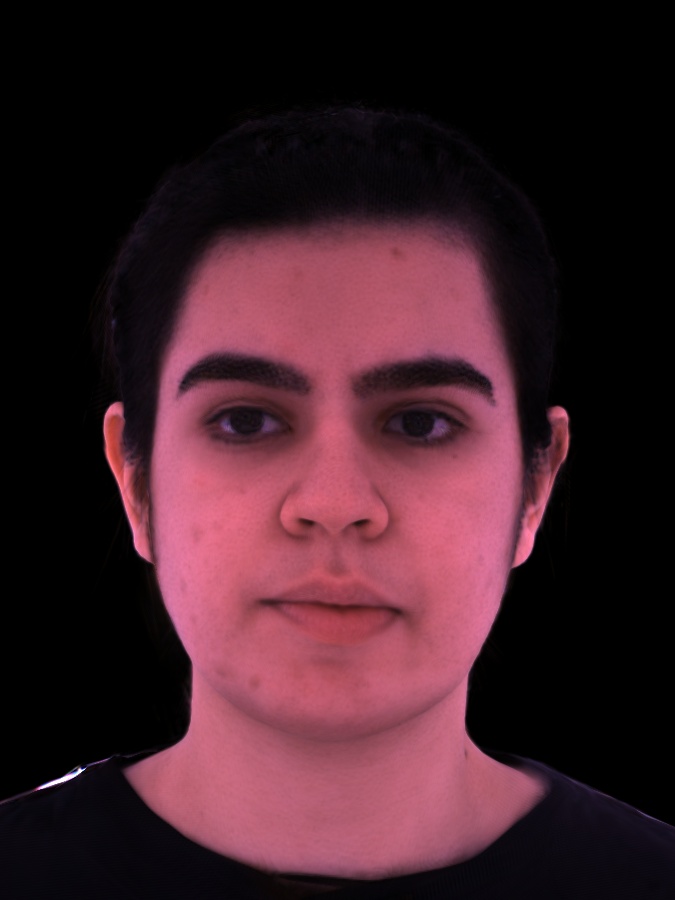}%
        \includegraphics[width=\onetenthfigurewidth\linewidth]{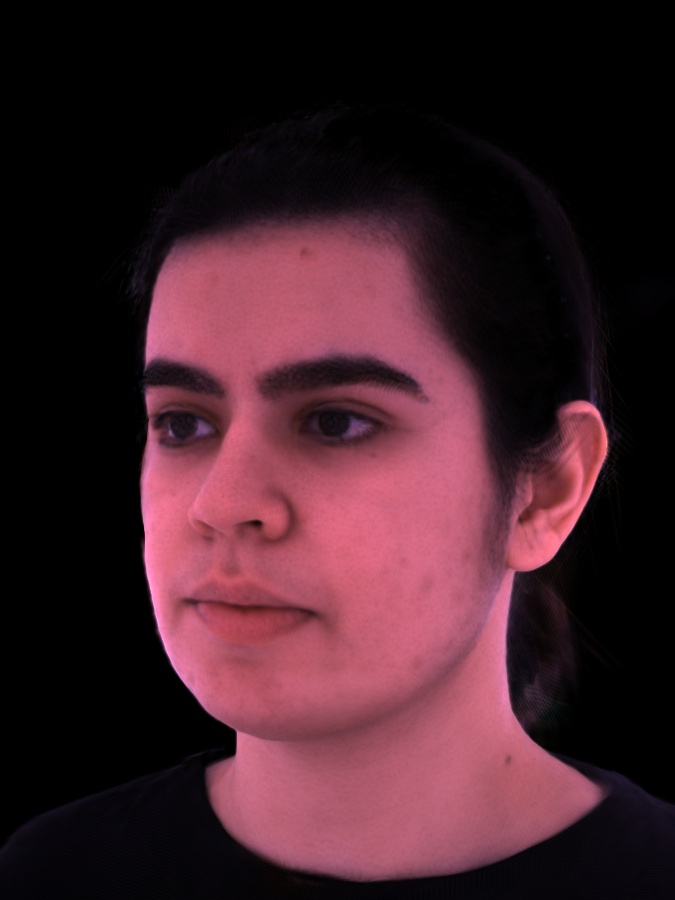}
        \includegraphics[width=\onetenthfigurewidth\linewidth]{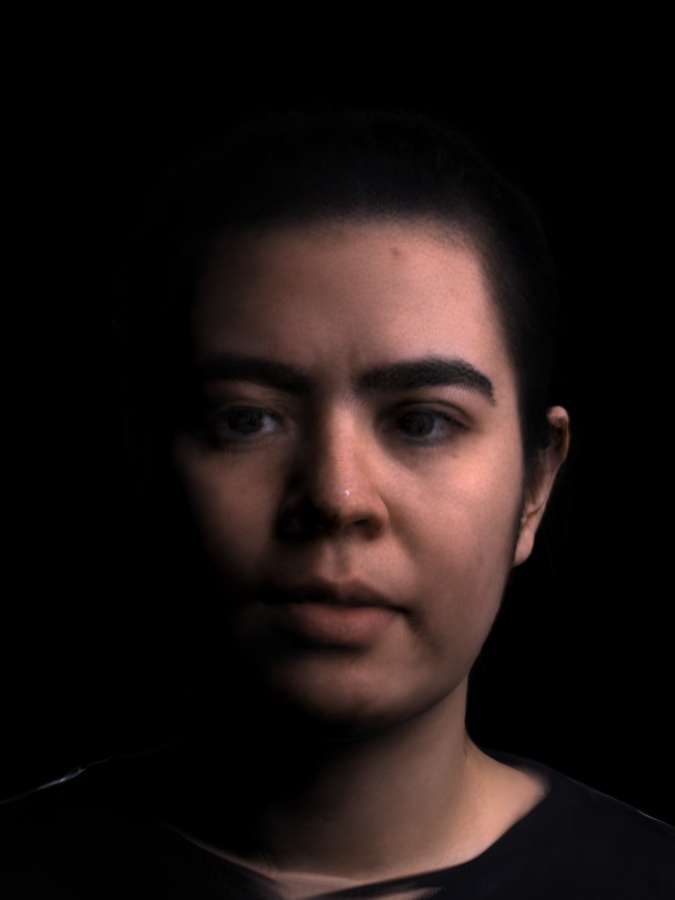}%
        \includegraphics[width=\onetenthfigurewidth\linewidth]{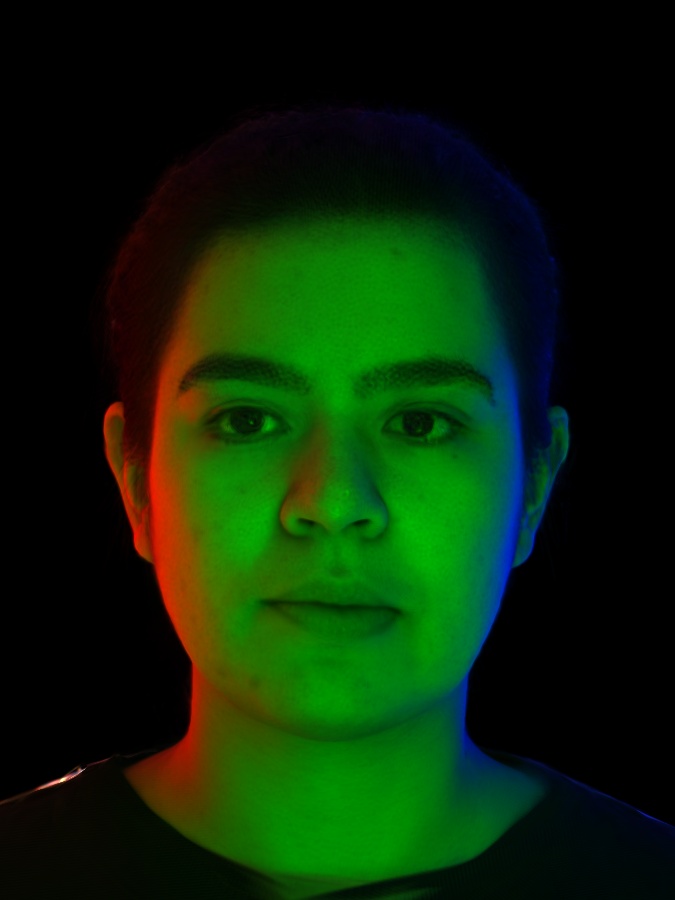}%
        \includegraphics[width=\onetenthfigurewidth\linewidth]{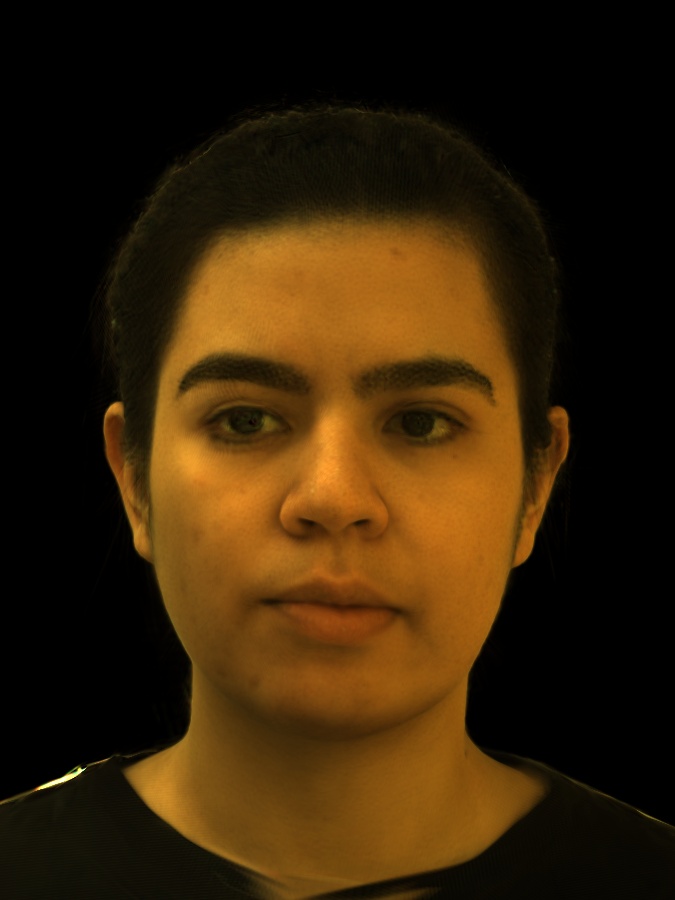}
        \includegraphics[width=\onetenthfigurewidth\linewidth]{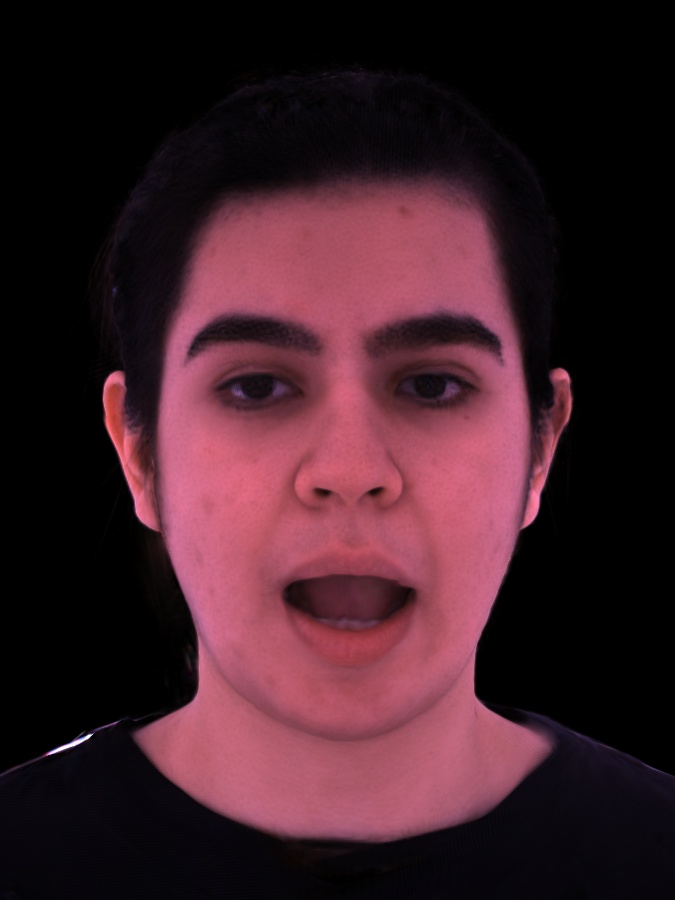}%
        \includegraphics[width=\onetenthfigurewidth\linewidth]{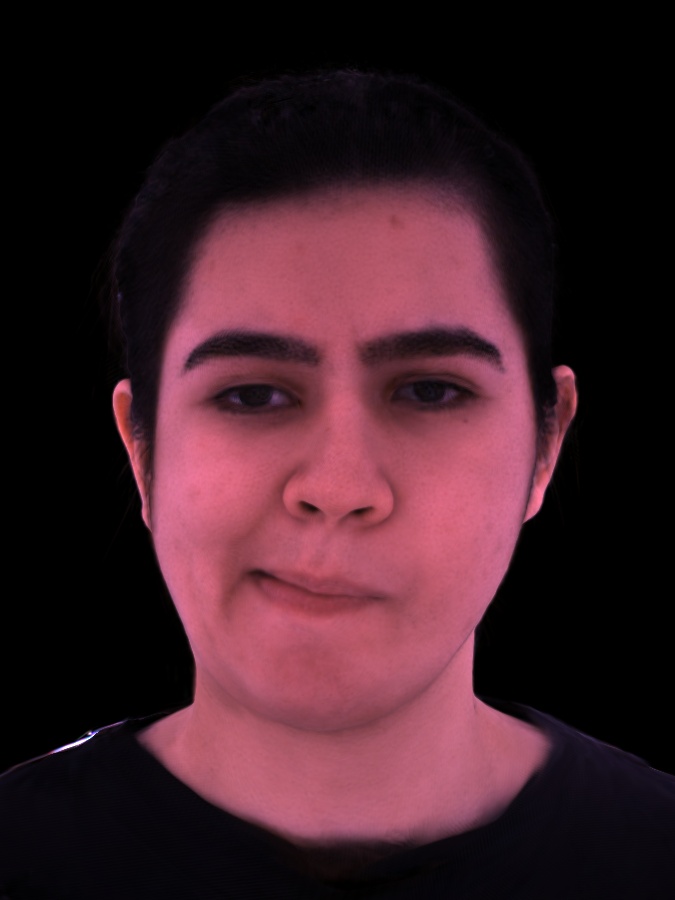}%
        \includegraphics[width=\onetenthfigurewidth\linewidth]{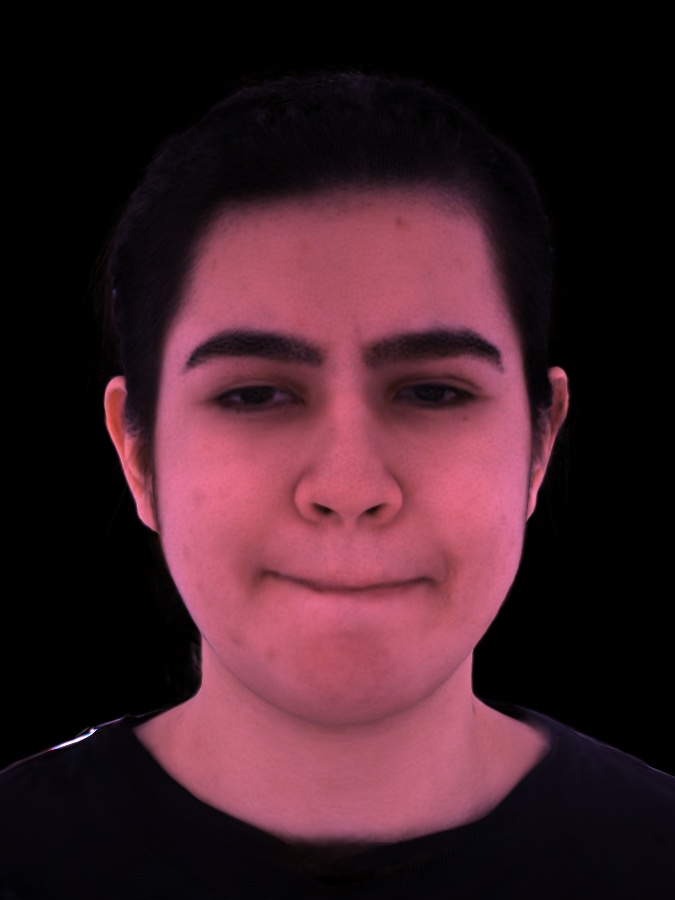}
    \end{minipage}

    \begin{minipage}[t]{\linewidth}
        \centering
        \includegraphics[width=\onetenthfigurewidth\linewidth]{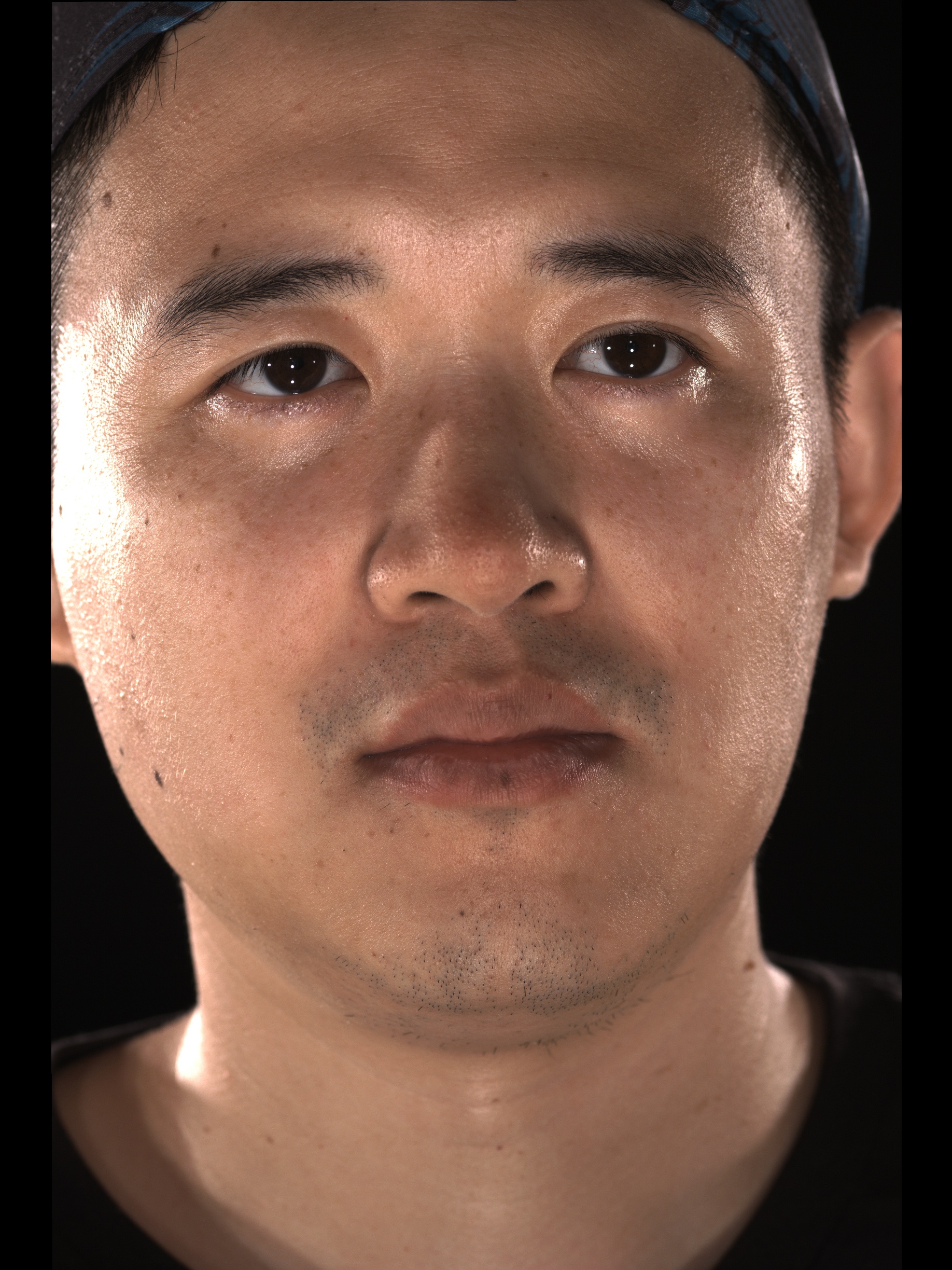}
        \includegraphics[width=\onetenthfigurewidth\linewidth]{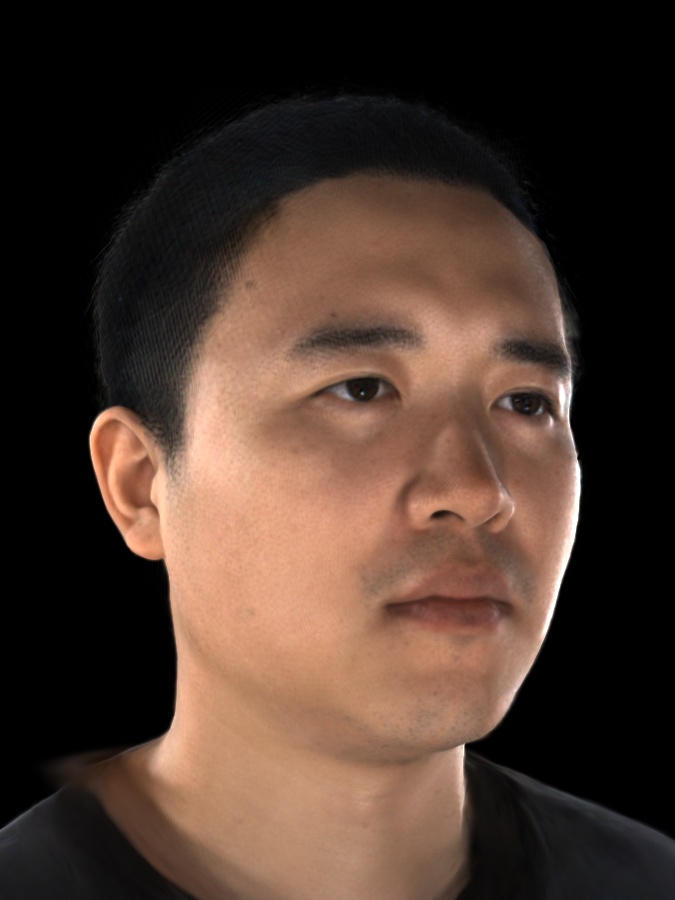}%
        \includegraphics[width=\onetenthfigurewidth\linewidth]{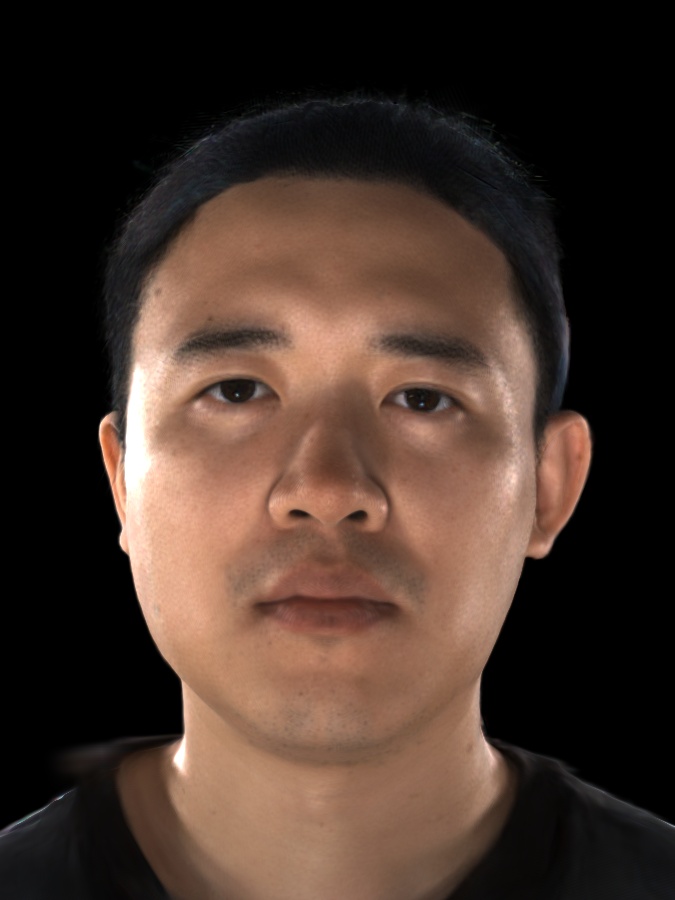}%
        \includegraphics[width=\onetenthfigurewidth\linewidth]{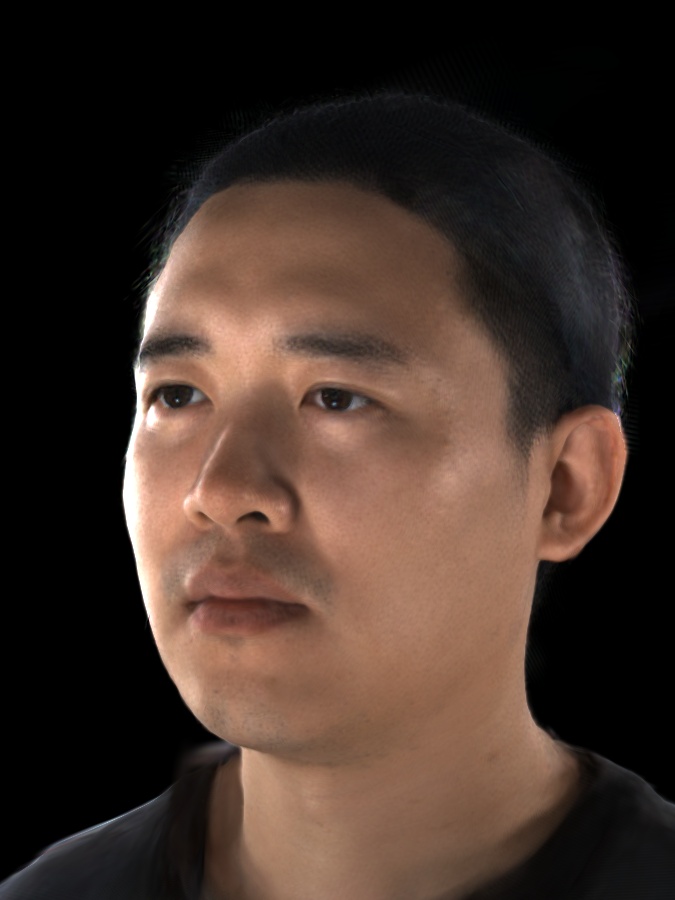}
        \includegraphics[width=\onetenthfigurewidth\linewidth]{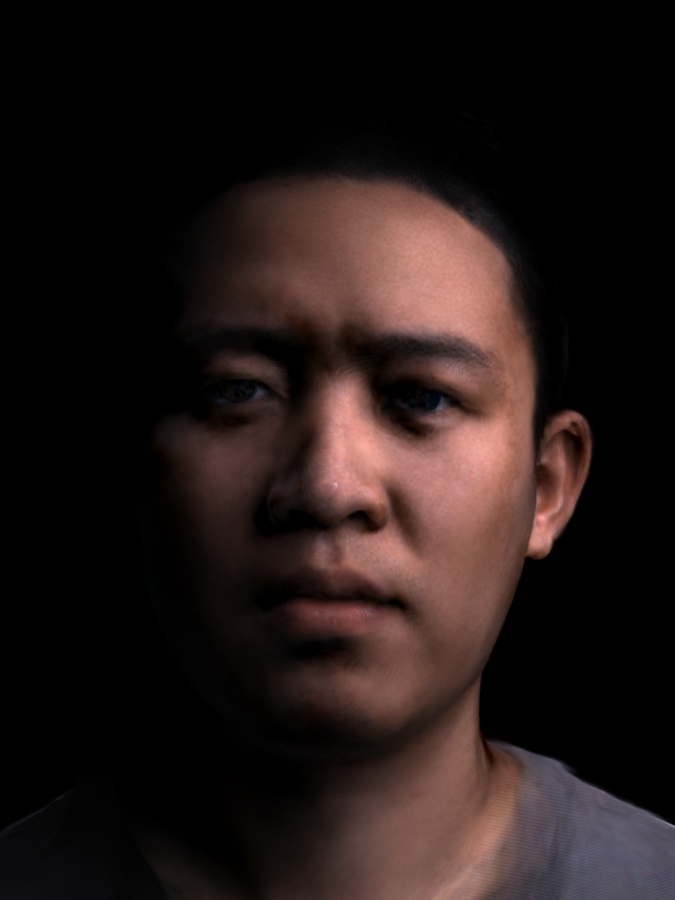}%
        \includegraphics[width=\onetenthfigurewidth\linewidth]{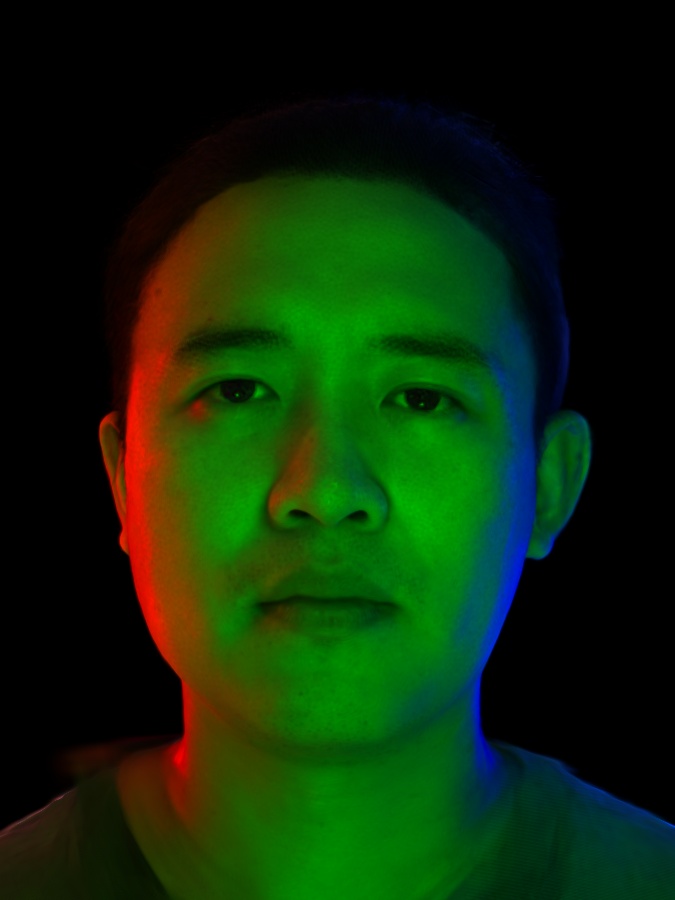}%
        \includegraphics[width=\onetenthfigurewidth\linewidth]{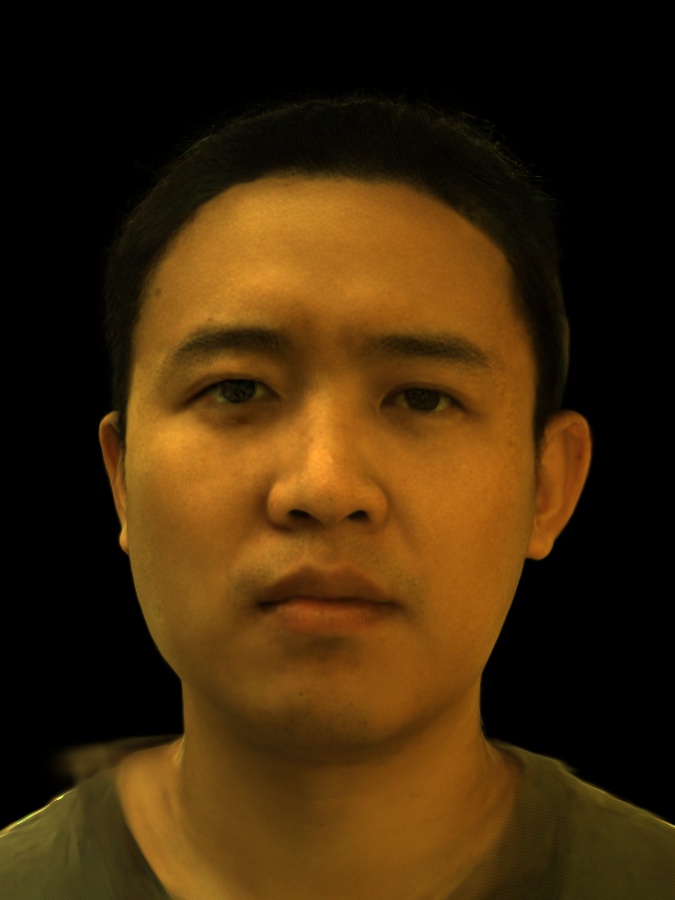}
        \includegraphics[width=\onetenthfigurewidth\linewidth]{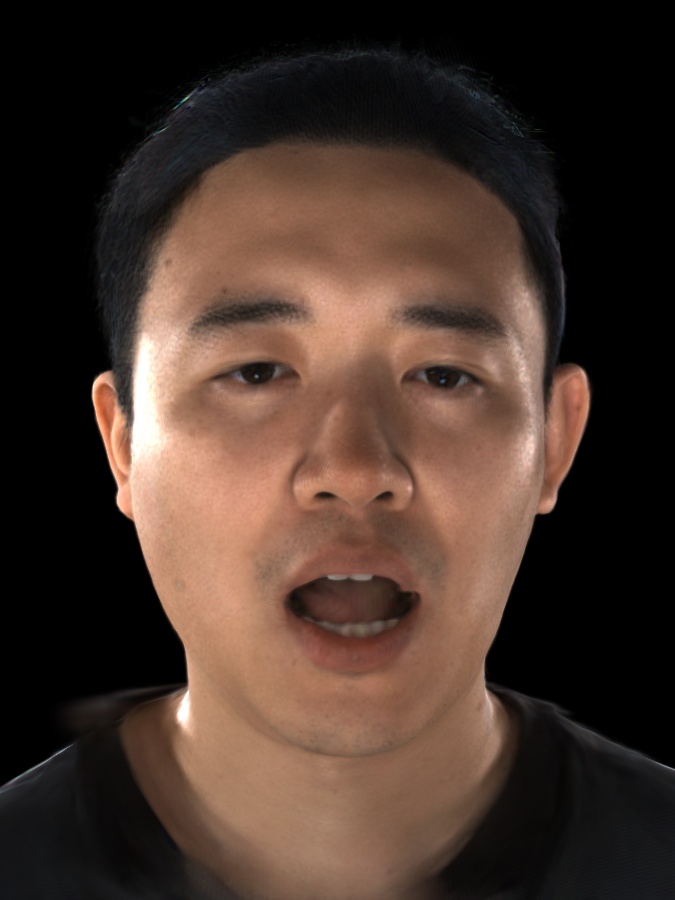}%
        \includegraphics[width=\onetenthfigurewidth\linewidth]{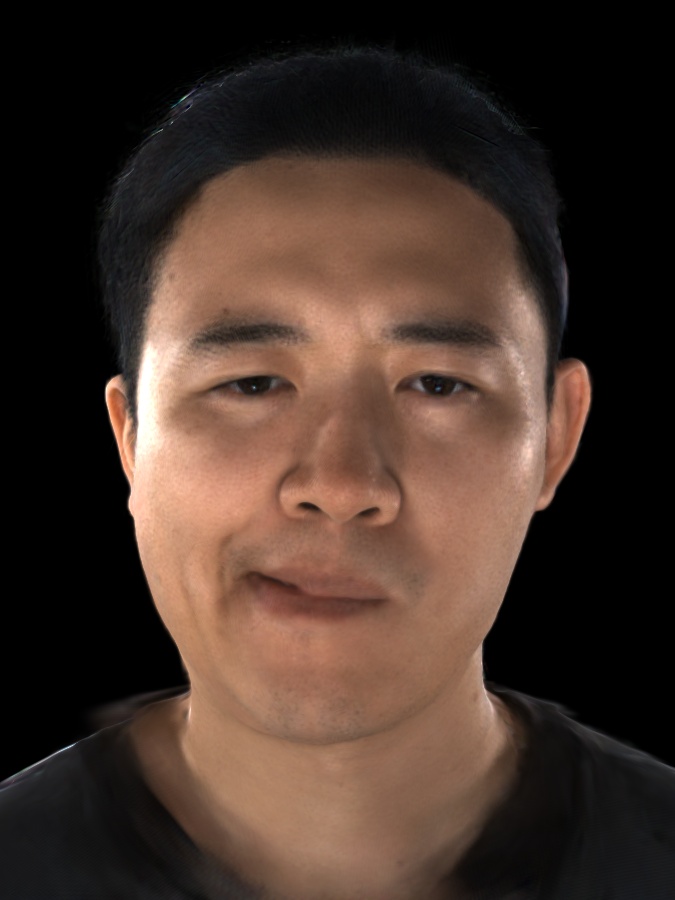}%
        \includegraphics[width=\onetenthfigurewidth\linewidth]{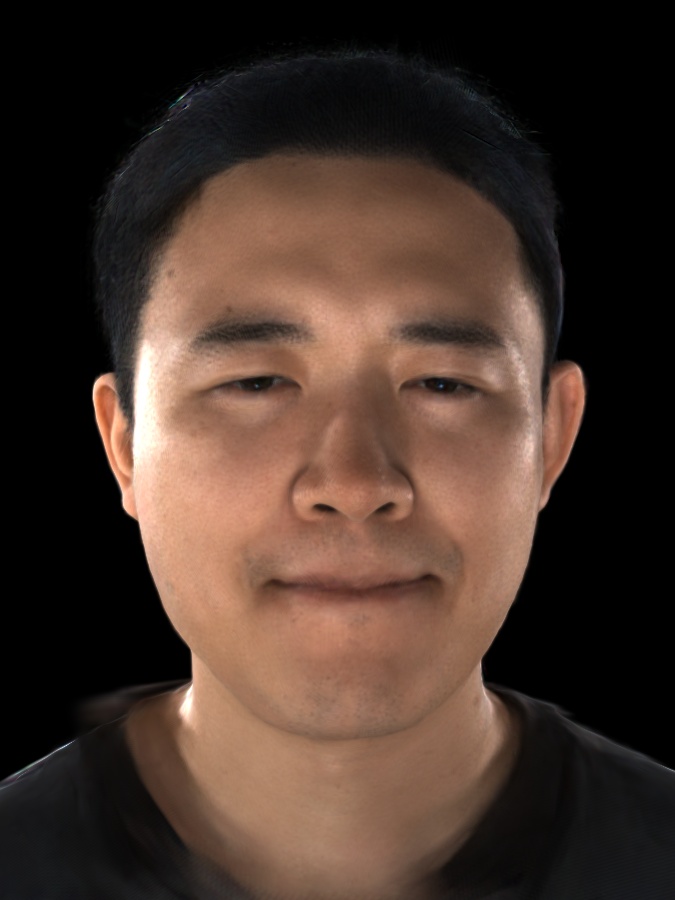}
    \end{minipage}

    \begin{minipage}[t]{\linewidth}
        \centering
        \includegraphics[width=\onetenthfigurewidth\linewidth]{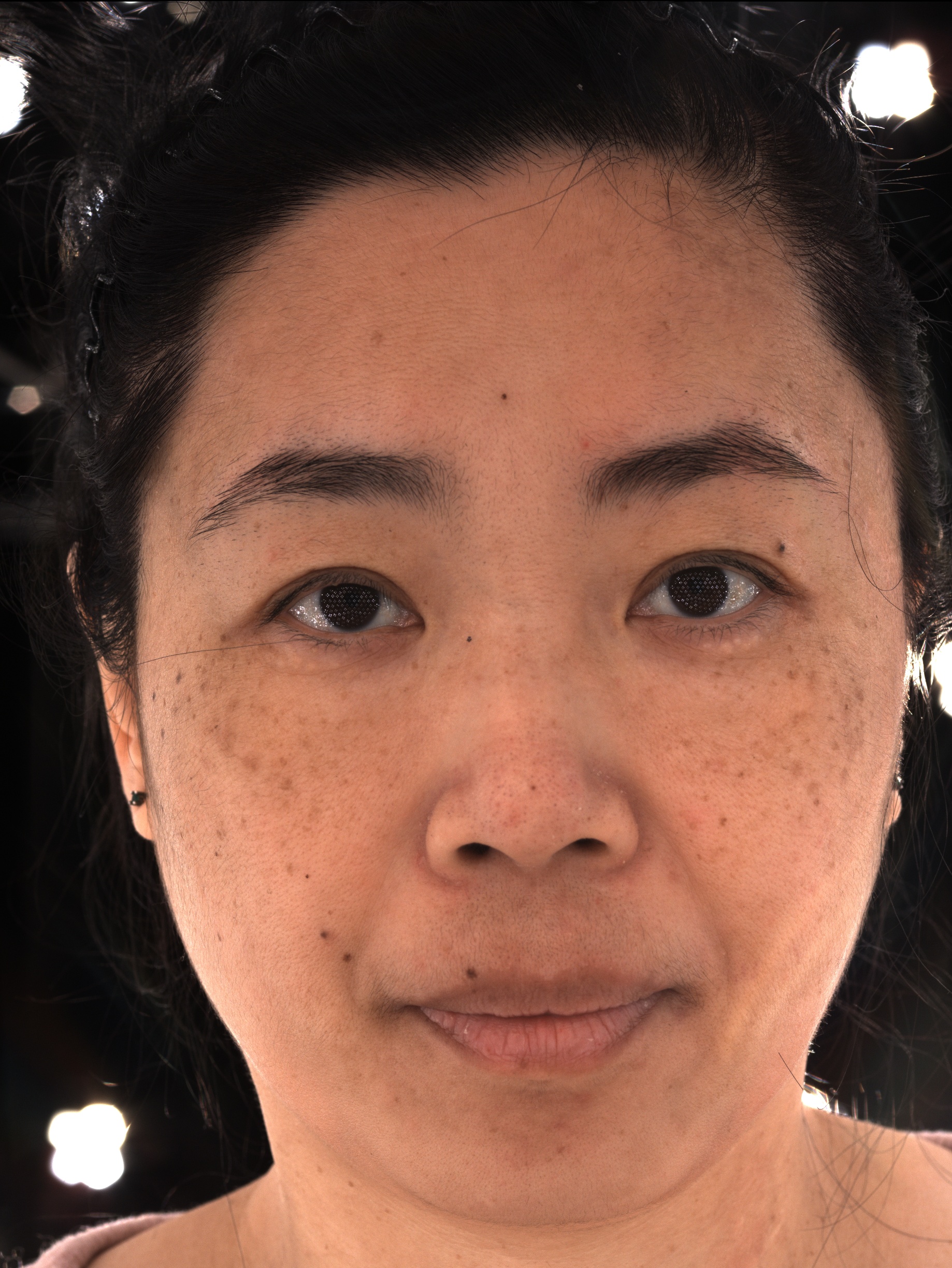}
        \includegraphics[width=\onetenthfigurewidth\linewidth]{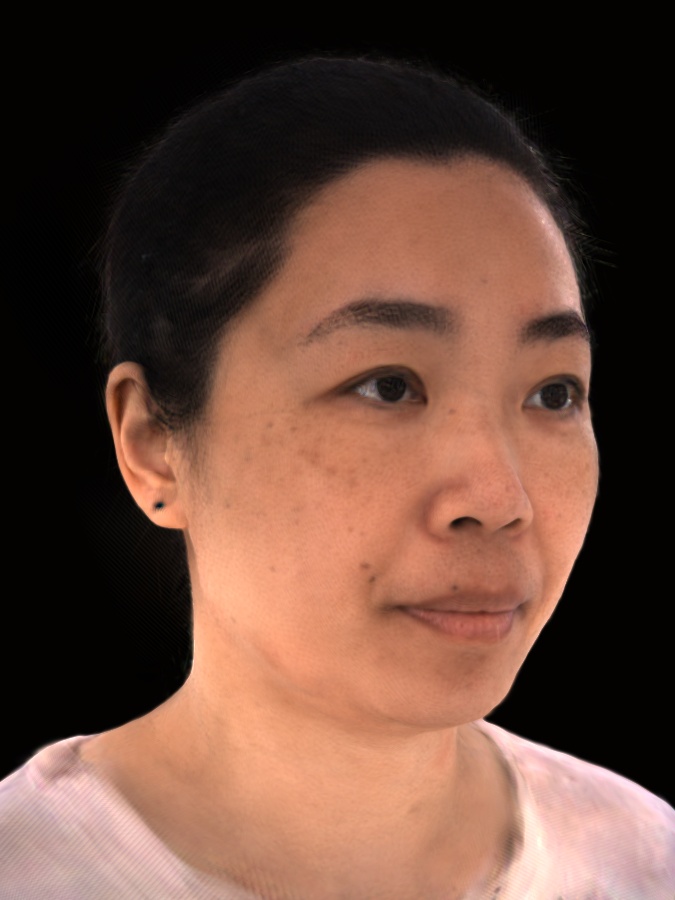}%
        \includegraphics[width=\onetenthfigurewidth\linewidth]{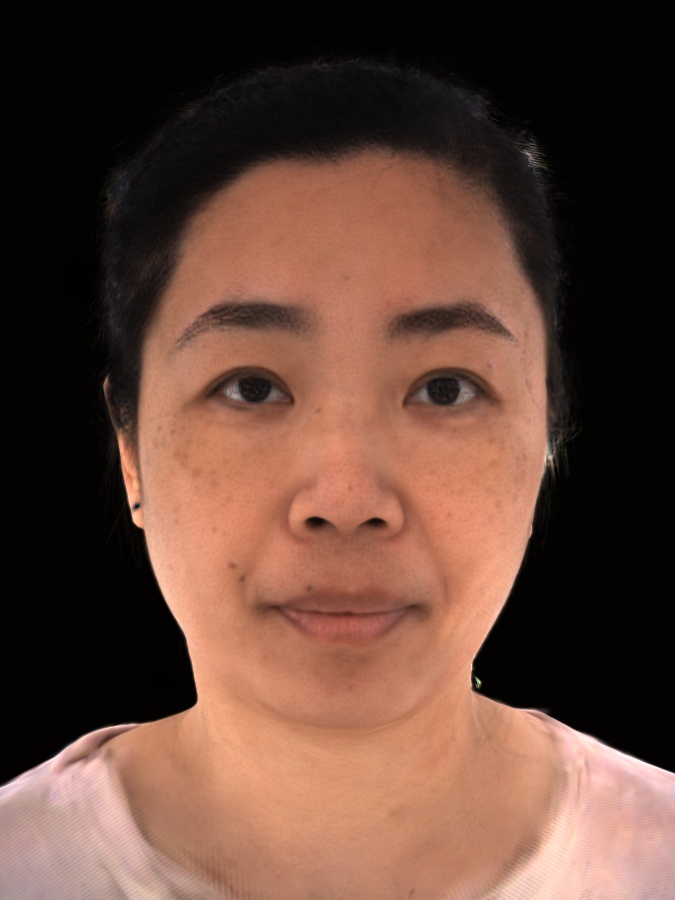}%
        \includegraphics[width=\onetenthfigurewidth\linewidth]{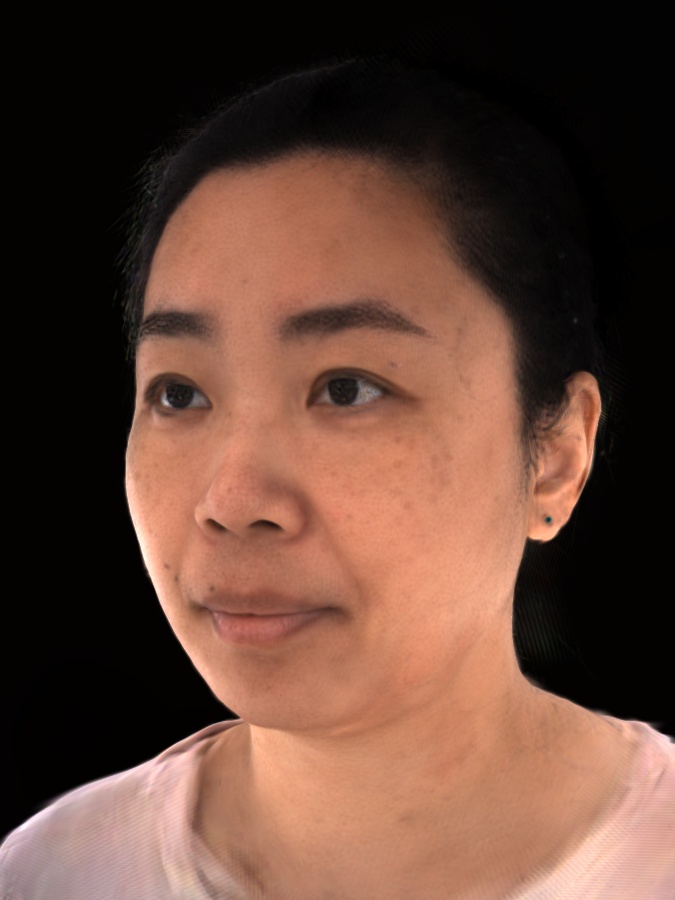}
        \includegraphics[width=\onetenthfigurewidth\linewidth]{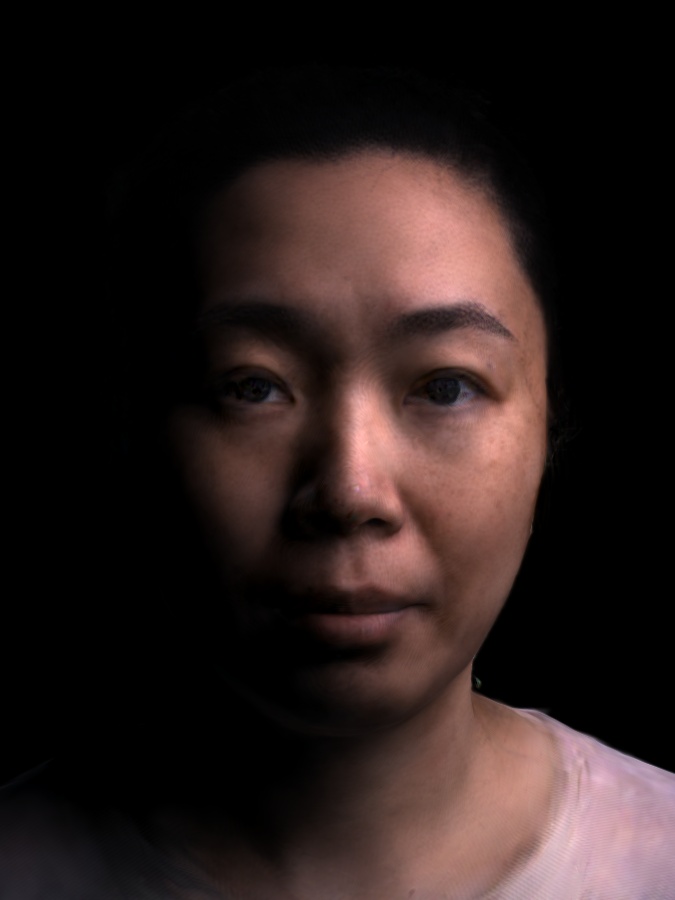}%
        \includegraphics[width=\onetenthfigurewidth\linewidth]{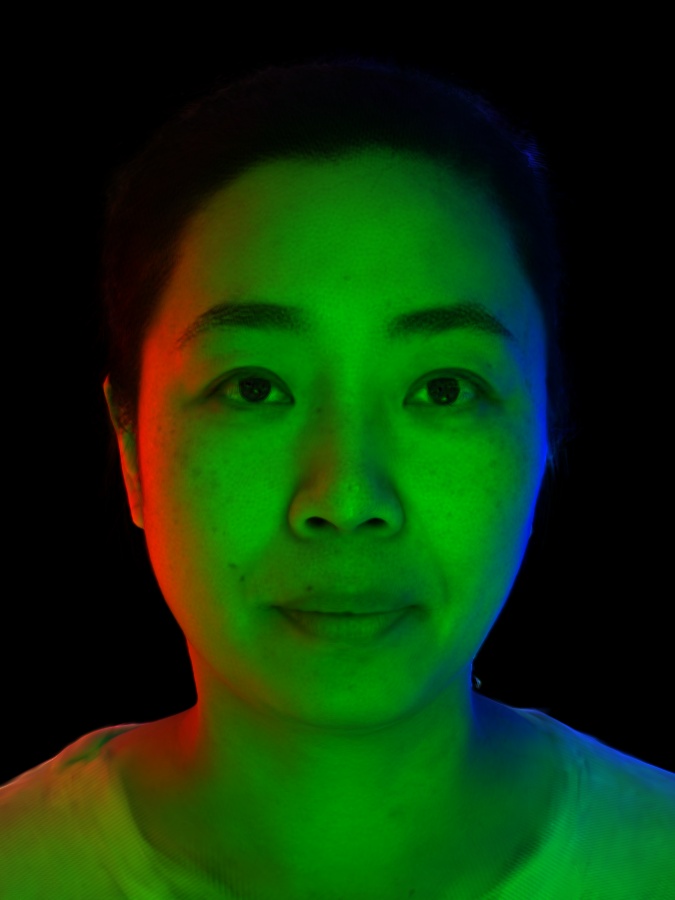}%
        \includegraphics[width=\onetenthfigurewidth\linewidth]{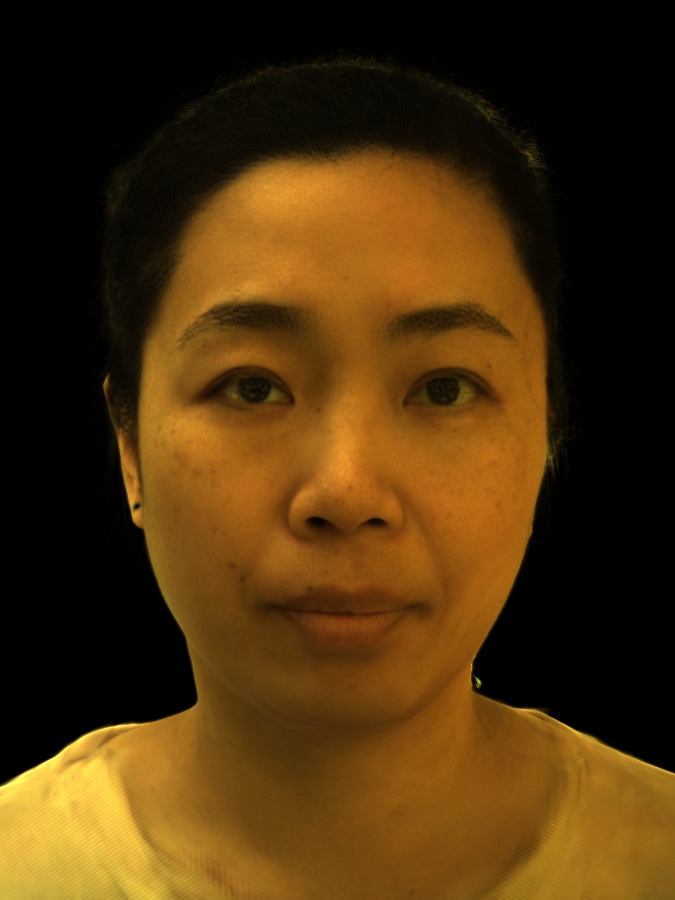}
        \includegraphics[width=\onetenthfigurewidth\linewidth]{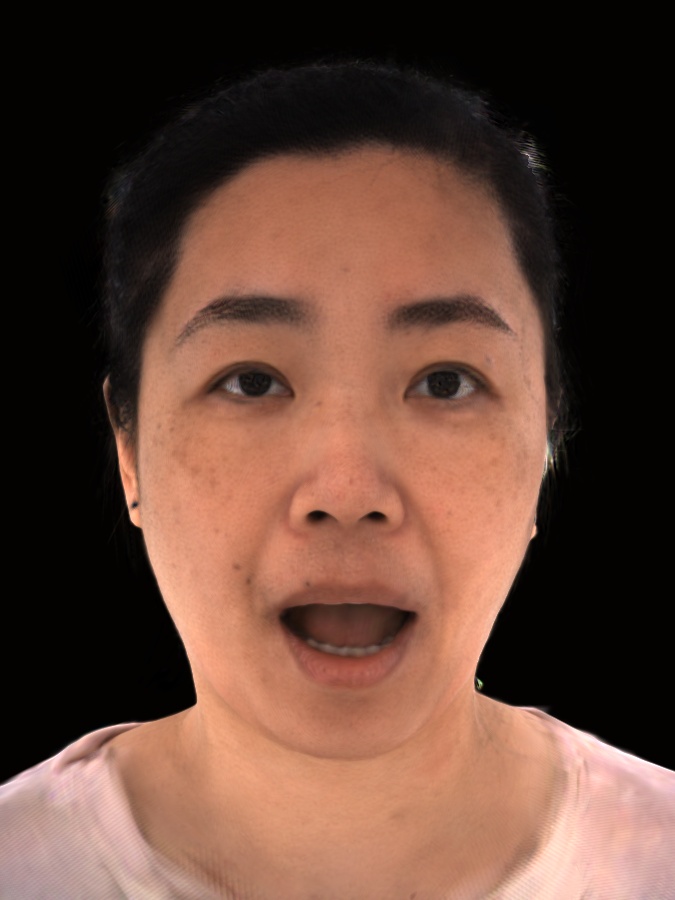}%
        \includegraphics[width=\onetenthfigurewidth\linewidth]{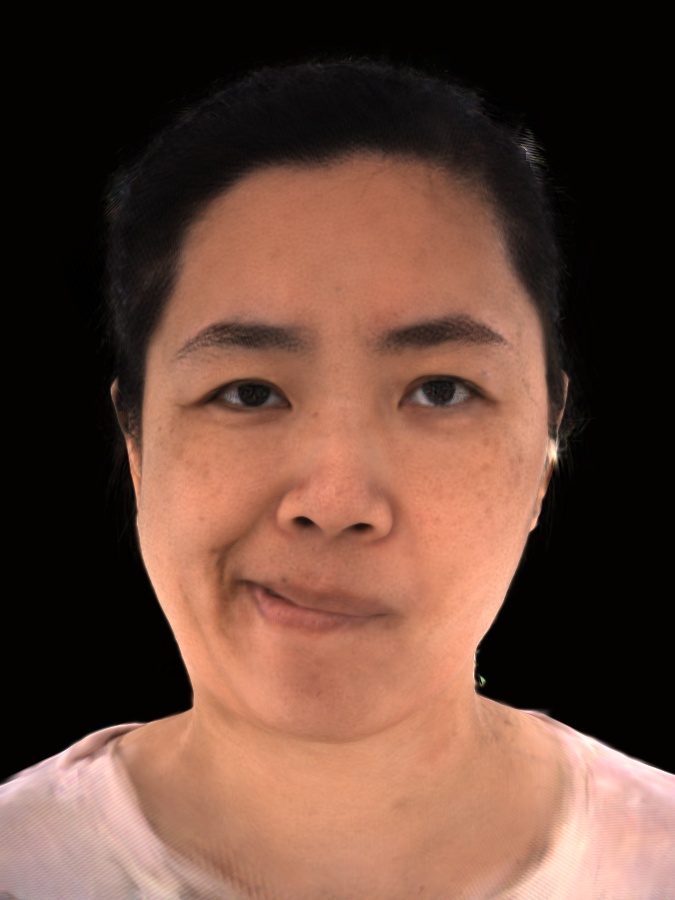}%
        \includegraphics[width=\onetenthfigurewidth\linewidth]{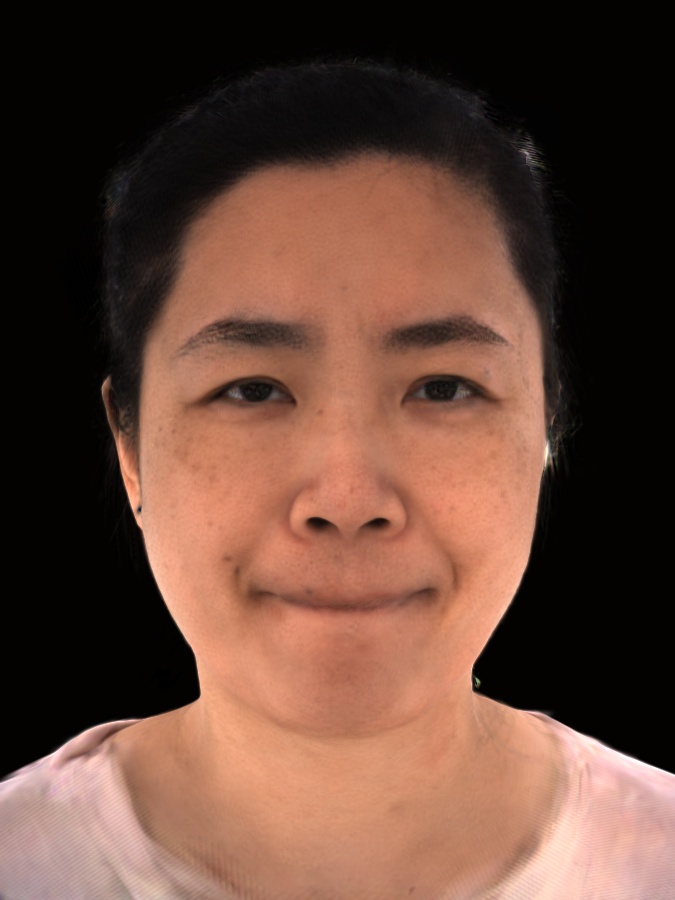}
    \end{minipage}

    %\vspace{5pt}
    \begin{minipage}{\linewidth}
        \centering
        \begin{minipage}[t]{\onetenthfigurewidth\linewidth}
            \centering
            \subfloat{\small Reference}
        \end{minipage}
        % \begin{minipage}[t]{\onetenthfigurewidth\linewidth}
        %     \centering
        %     \subfloat{ }%
        % \end{minipage}
        \begin{minipage}[t]{0.29\linewidth}
            \centering
            \subfloat{\small Reconstruction}%
        \end{minipage}
        % \begin{minipage}[t]{\onetenthfigurewidth\linewidth}
        %     \centering
        %     \subfloat{ }
        % \end{minipage}
        % \begin{minipage}[t]{\onetenthfigurewidth\linewidth}
        %     \centering
        %     \subfloat{ }%
        % \end{minipage}
        \begin{minipage}[t]{0.29\linewidth}
            \centering
            \subfloat{\small Relighting}%
        \end{minipage}
        % \begin{minipage}[t]{\onetenthfigurewidth\linewidth}
        %     \centering
        %     \subfloat{ }
        % \end{minipage}
        % \begin{minipage}[t]{\onetenthfigurewidth\linewidth}
        %     \centering
        %     \subfloat{ }%
        % \end{minipage}
        \begin{minipage}[t]{0.29\linewidth}
            \centering
            \subfloat{\small Animation}%
        \end{minipage}
        % \begin{minipage}[t]{\onetenthfigurewidth\linewidth}
        %     \centering
        %     \subfloat{ }
        % \end{minipage}
    \end{minipage}
    \end{minipage}
    \caption{Few-shot personalization results. Given multi-view images of a subject in a static expression under unknown fixed illumination, our model enables the creation of a high-fidelity avatar that can be animated and relighted. Our model learns the relightable appearance even when the input images suffer from strong specular reflection.
    }
\label{fig:personalization}
\end{figure*}

\newcommand{\onefigureheight}{0.178}

\begin{figure*}%[t]
    \centering
    \begin{minipage}[t]{\linewidth}
        \centering
        \includegraphics[height=\onefigureheight\linewidth]{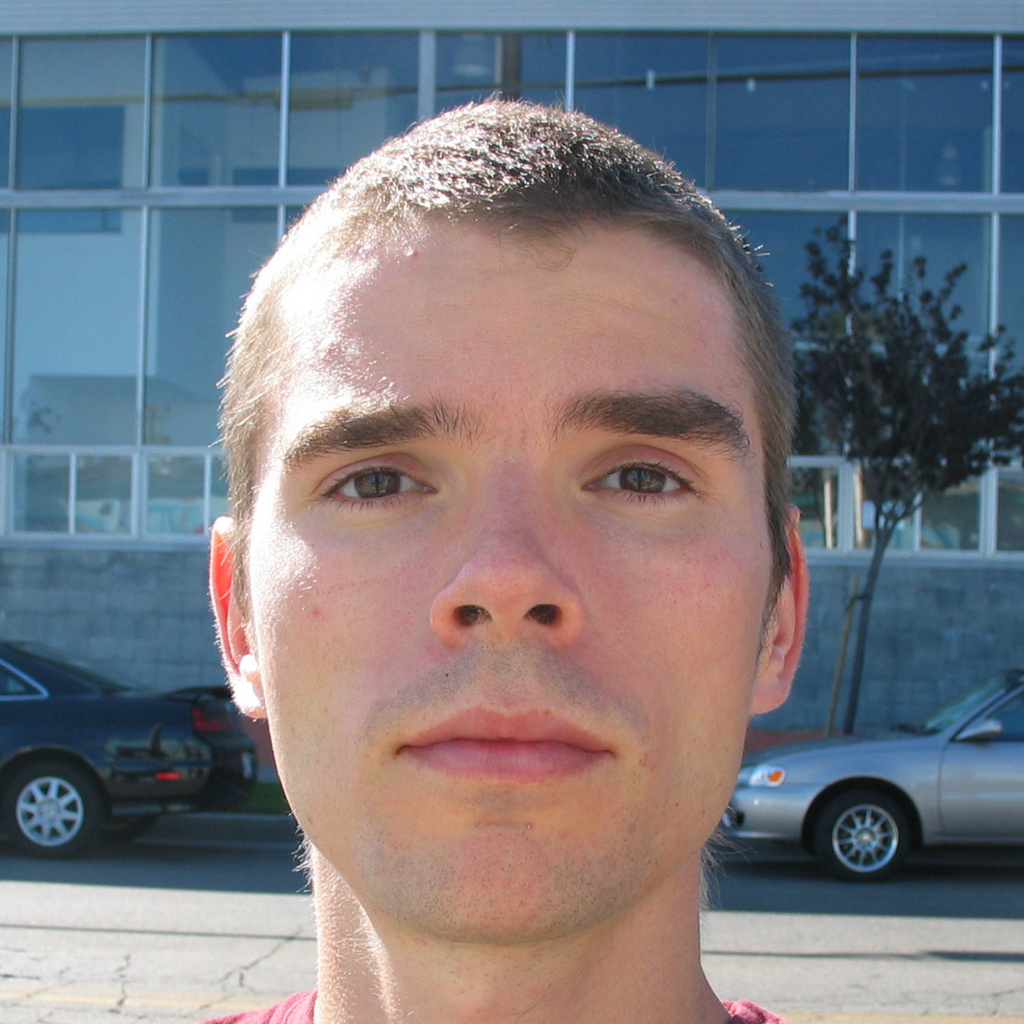}
        \includegraphics[height=\onefigureheight\linewidth]{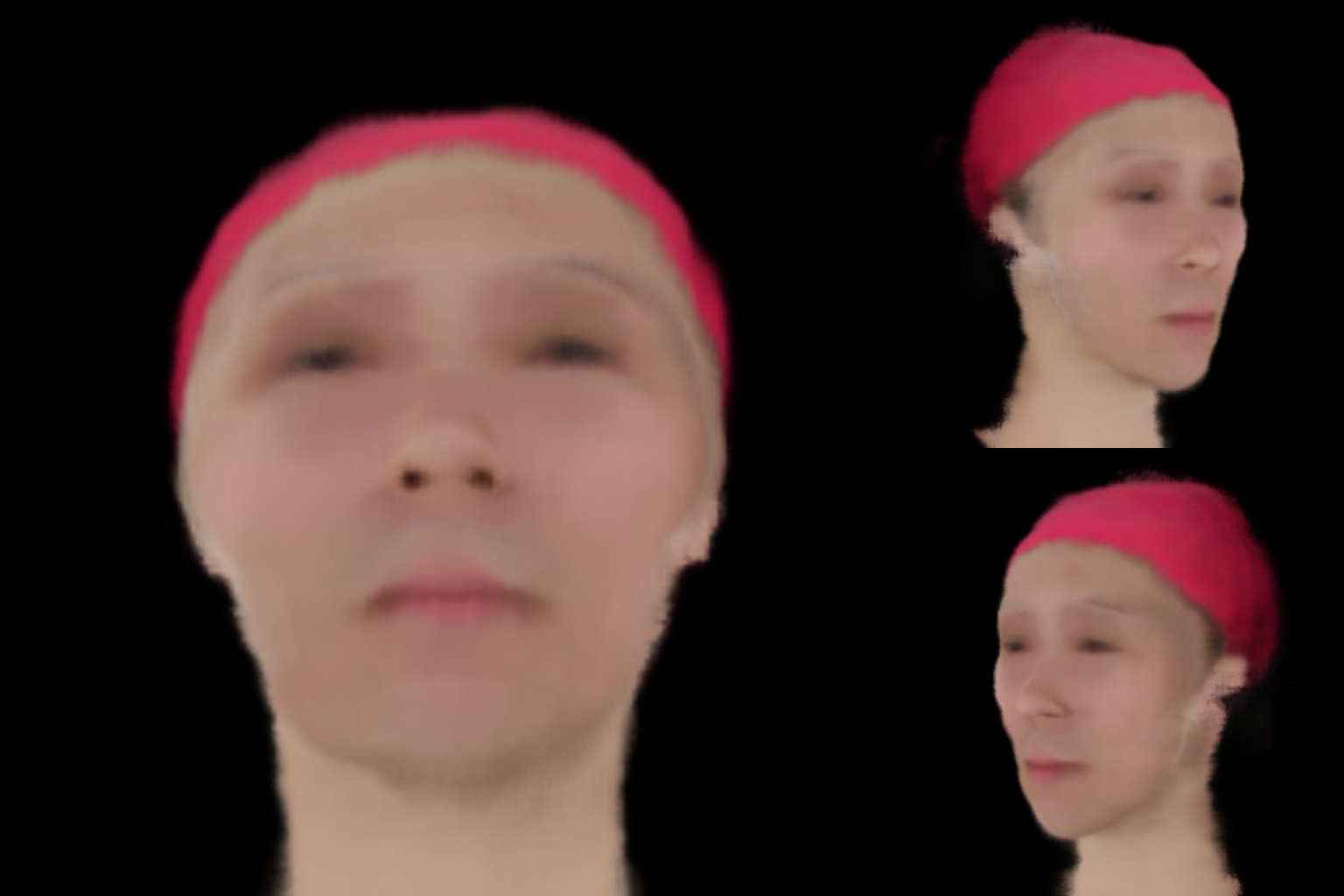}
        \includegraphics[height=\onefigureheight\linewidth]{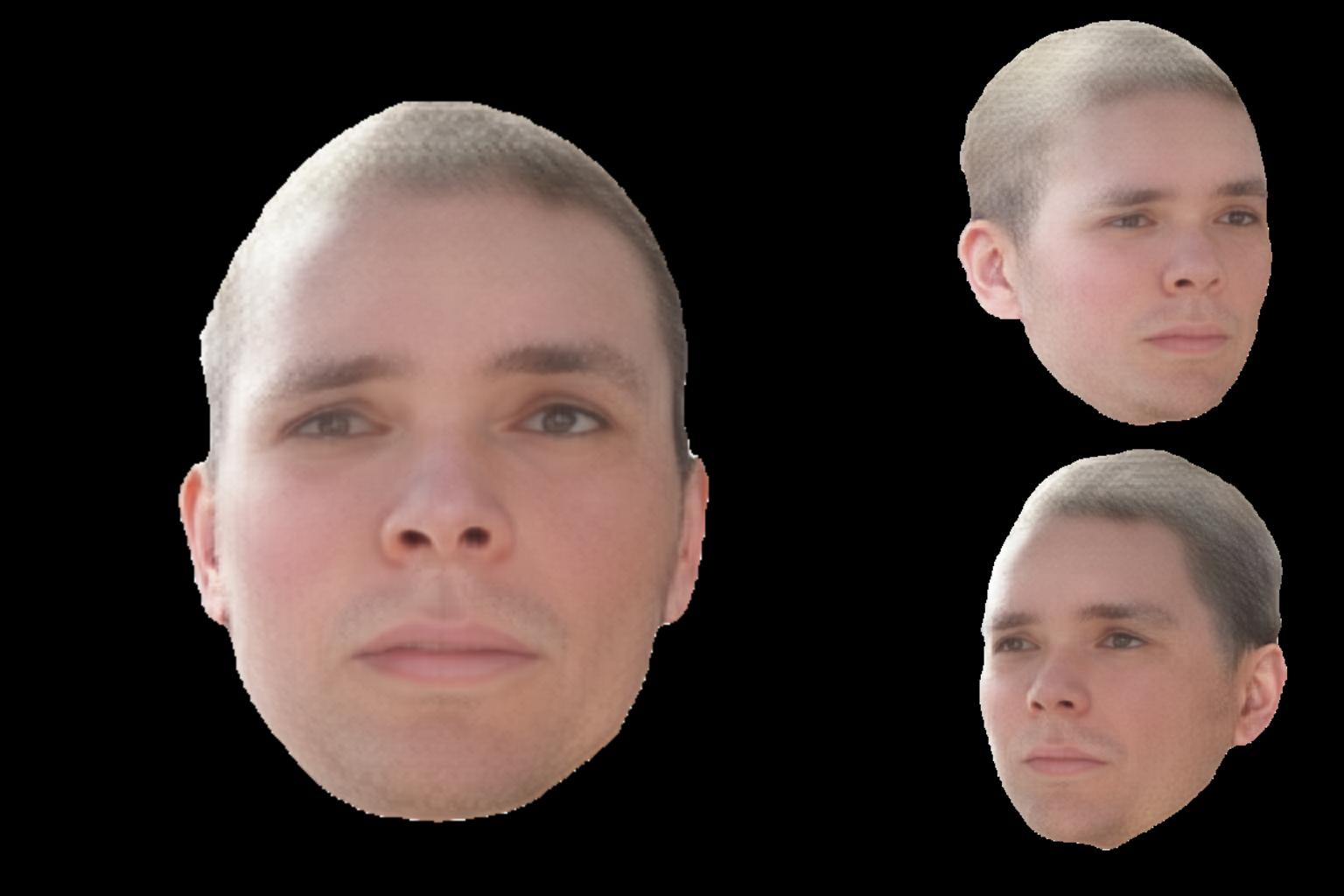}
        \includegraphics[height=\onefigureheight\linewidth]{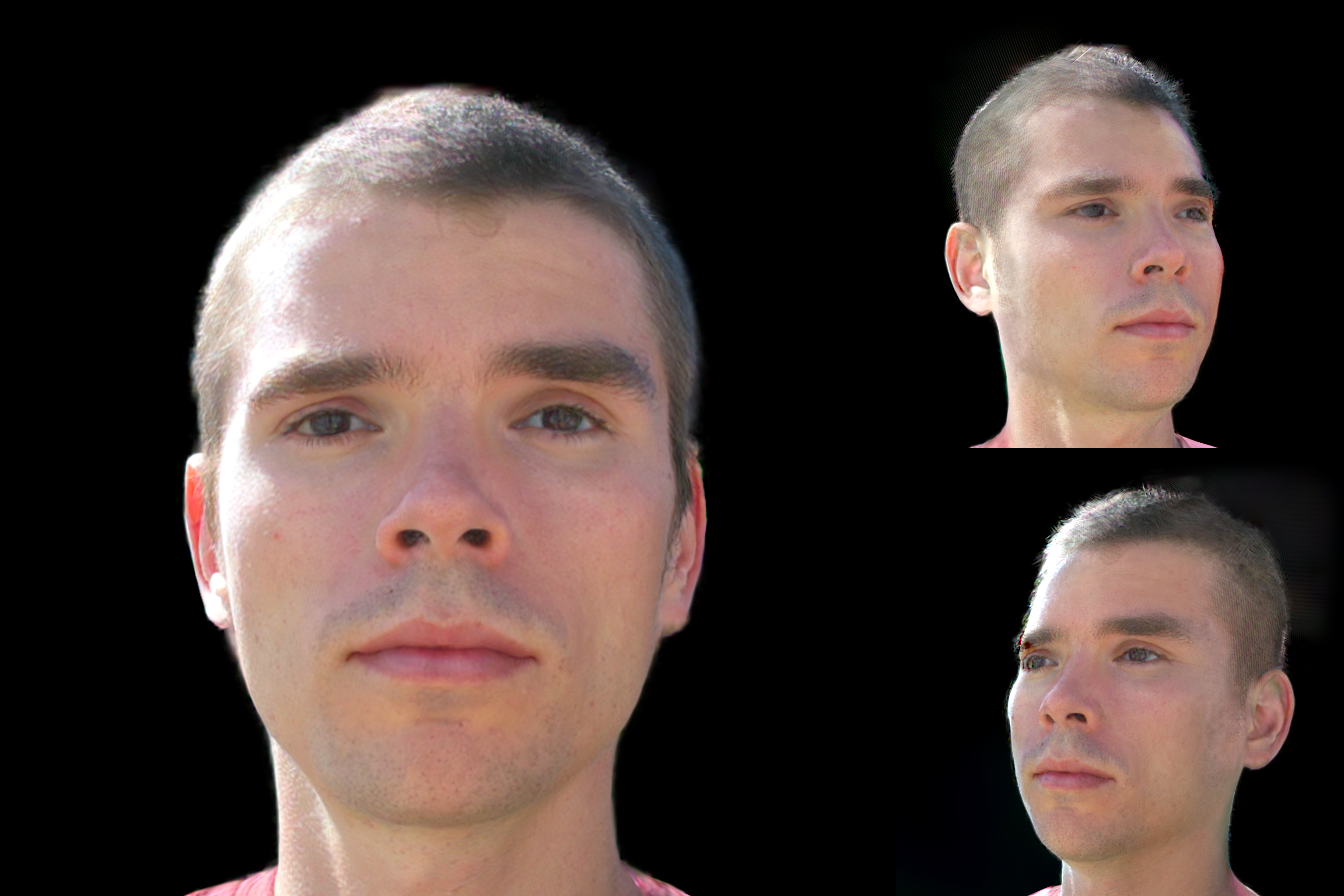}
    \end{minipage}

    \begin{minipage}[t]{\linewidth}
        \centering
        \includegraphics[height=\onefigureheight\linewidth]{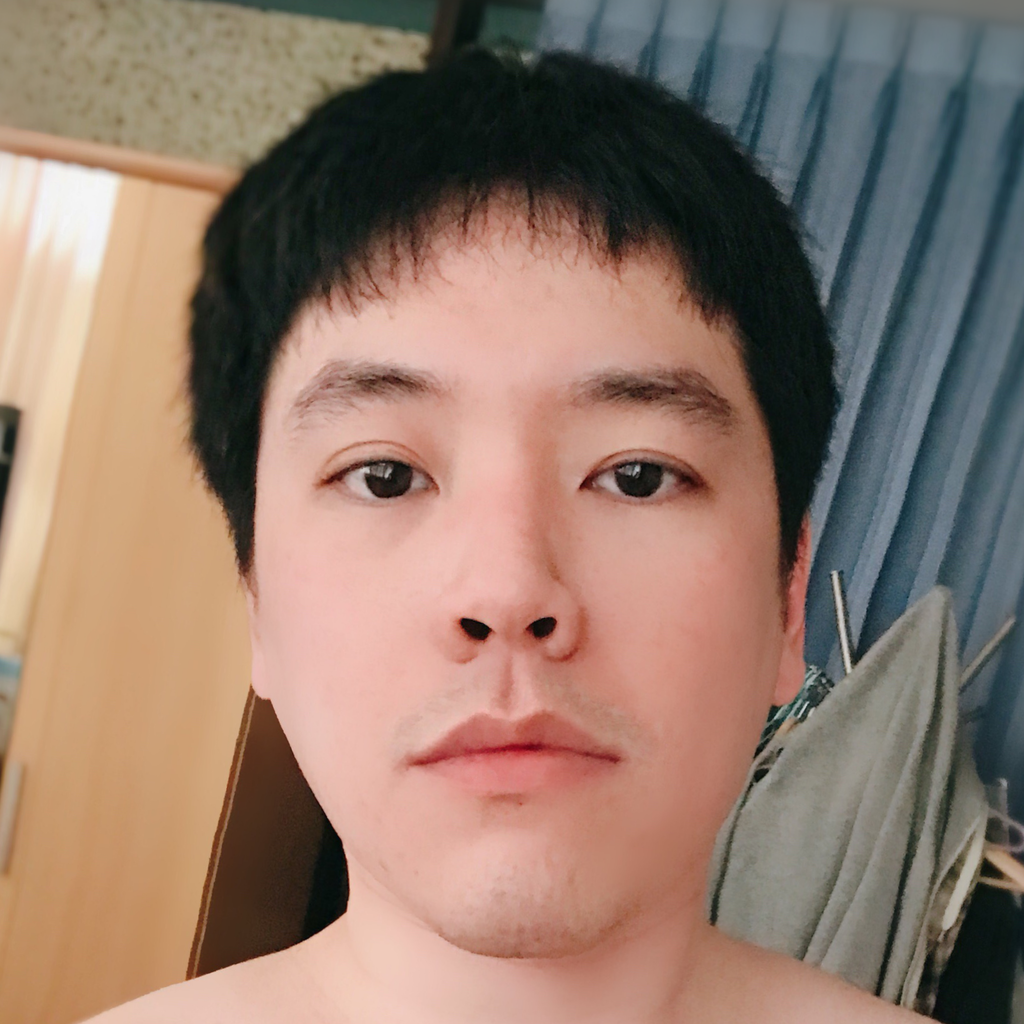}
        \includegraphics[height=\onefigureheight\linewidth]{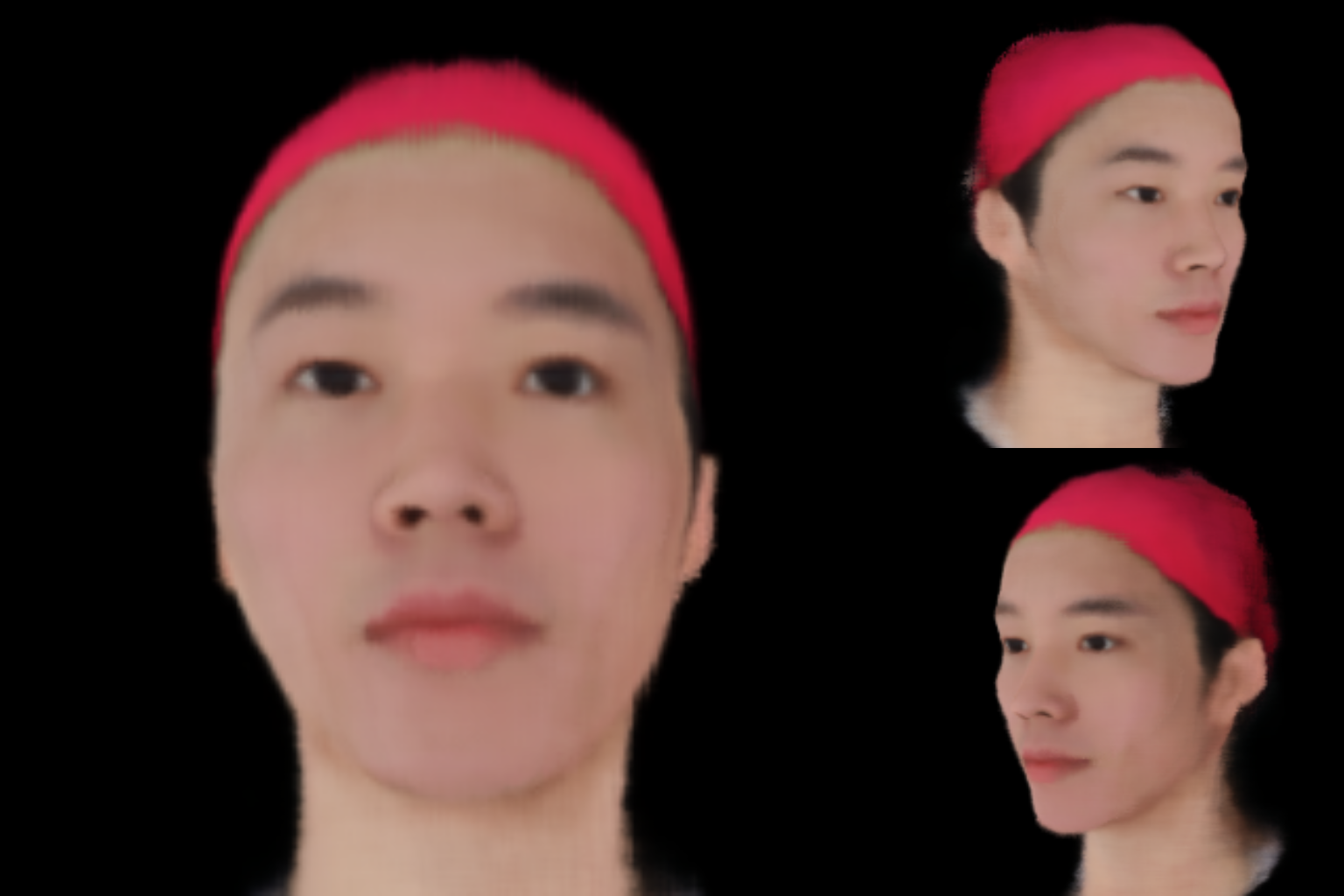}
        \includegraphics[height=\onefigureheight\linewidth]{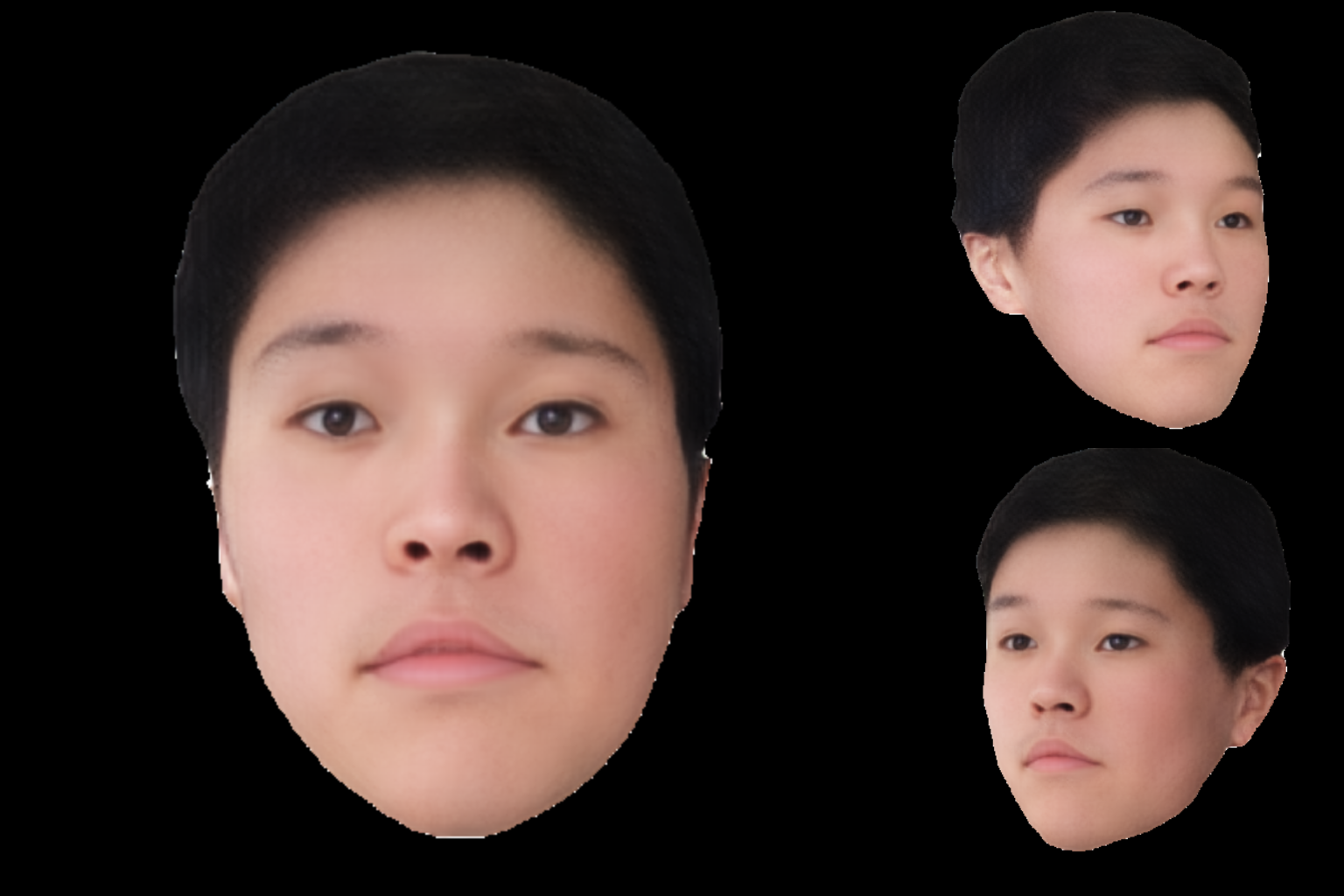}
        \includegraphics[height=\onefigureheight\linewidth]{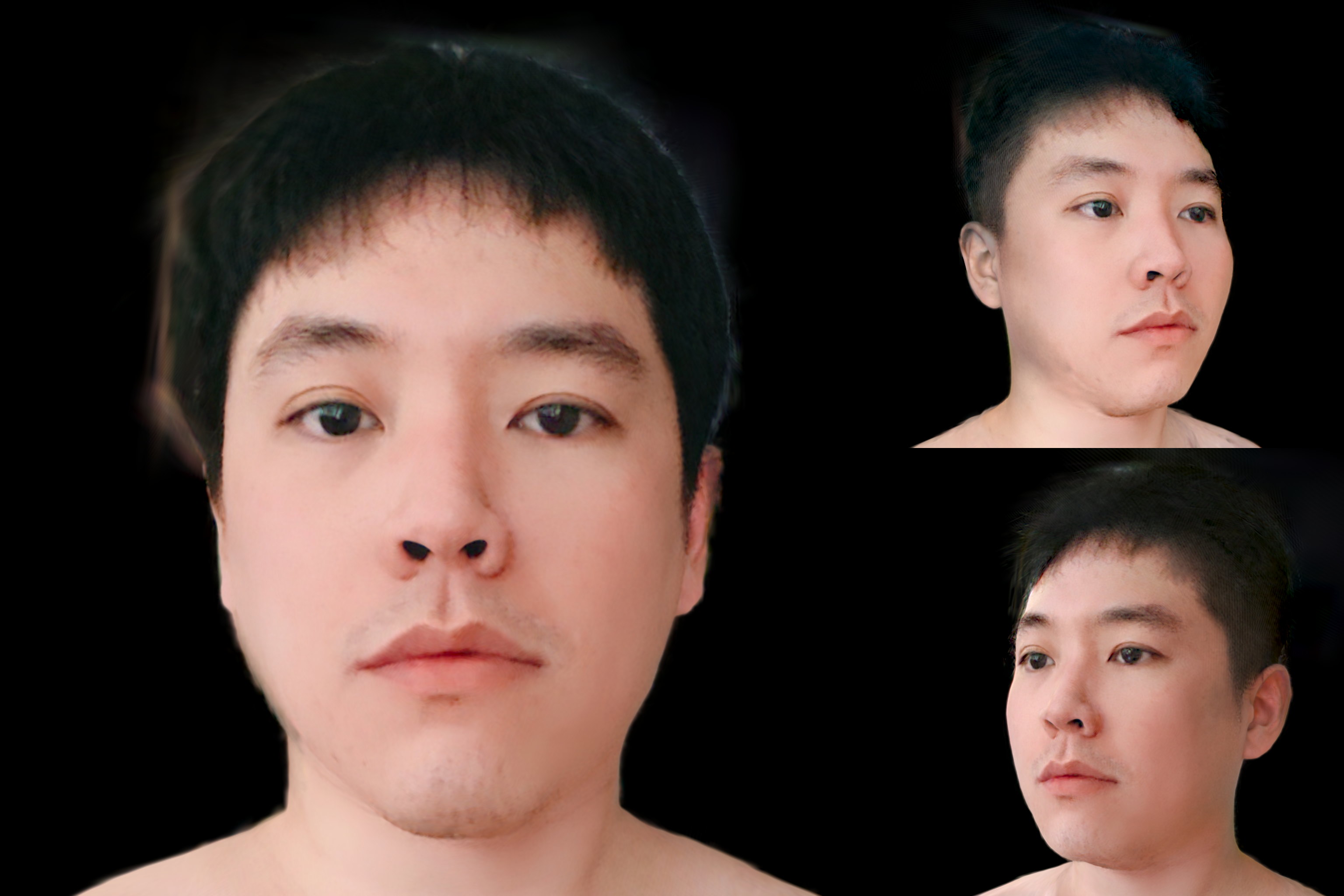}
    \end{minipage}

    \vspace{5pt}
    \begin{minipage}{\linewidth}
        \centering
        \begin{minipage}[t]{0.18\linewidth}
            \centering
            \subfloat{Input}
        \end{minipage}
        \begin{minipage}[t]{0.267\linewidth}
            \centering
            \subfloat{MoFaNeRF}
        \end{minipage}
        \begin{minipage}[t]{0.267\linewidth}
            \centering
            \subfloat{HeadNeRF}
        \end{minipage}
        \begin{minipage}[t]{0.267\linewidth}
            \centering
            \subfloat{Ours}
        \end{minipage}
    \end{minipage}
    \caption{\yl{Comparison} with existing volumetric head priors, MoFaNeRF~\cite{zhuang2022mofanerf} and HeadNeRF~\cite{hong2022headnerf}, on single-view head reconstruction. We show the fitting results in the original view as well as two different views. The reconstruction results of our method achieve \yl{significantly} better visual quality.
    }
\label{fig:cmp_fit_single}
\end{figure*}

}
{
}

\subsection{Dataset}
We capture dynamic facial performance of 254 subjects in a custom-built apparatus with synchronized multi-view cameras and controllable lighting condition. Each subject is asked to perform 21 predefined expressions, read out a paragraph, look at different directions and freely perform \yl{exaggerated} and combined expressions. We record 1800 frames for each subject at 16fps with 29 cameras surrounding the head with a resolution of $2448\times2048$. We utilize the group light pattern with basic background illumination~\cite{bi2021deep,yang2023towards} for relightable model capture. We also capture a full-on frame every four shots. \revision{The background images are recorded and alpha blended with the rendering during training similar to previous work~\cite{lombardi2021mixture,yang2023towards}.} The unprecedented dataset consists of more than 13M images in total with diverse dynamic expressions and known varying \yl{illuminations}. Notice that the expressions do not need to be aligned or registered for different identities thanks to our self-supervised training framework, allowing for flexible dynamic performance capture. The network training on the dataset takes about two weeks on eight NVIDIA V100 graphics cards.

% \ifthenelse{\equal{\arxiv}{1}}
% {
% \input{figs/fig_fit_multiview}
% \input{figs/fig_interpolation}
% }
% {}

\begin{figure}
    \centering
    \begin{minipage}{\linewidth}
        \centering
        \begin{minipage}[t]{\onethirdfigurewidth\linewidth}
            \centering
            \includegraphics[width=\linewidth]{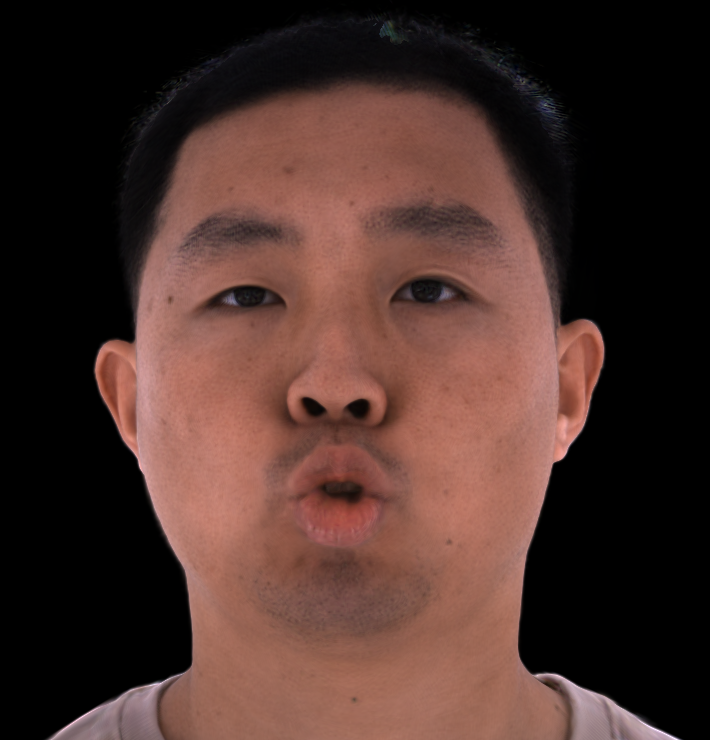}
            \subfloat{Target}
        \end{minipage}
        \begin{minipage}[t]{\onethirdfigurewidth\linewidth}
            \centering
            \includegraphics[width=\linewidth]{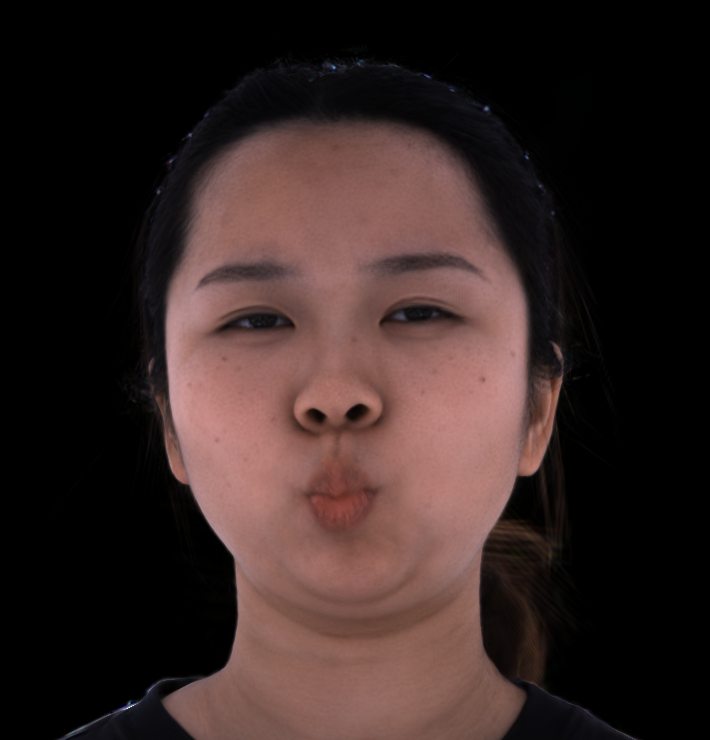}
            \subfloat{w/o ECL}
        \end{minipage}
        \begin{minipage}[t]{\onethirdfigurewidth\linewidth}
            \centering
            \includegraphics[width=\linewidth]{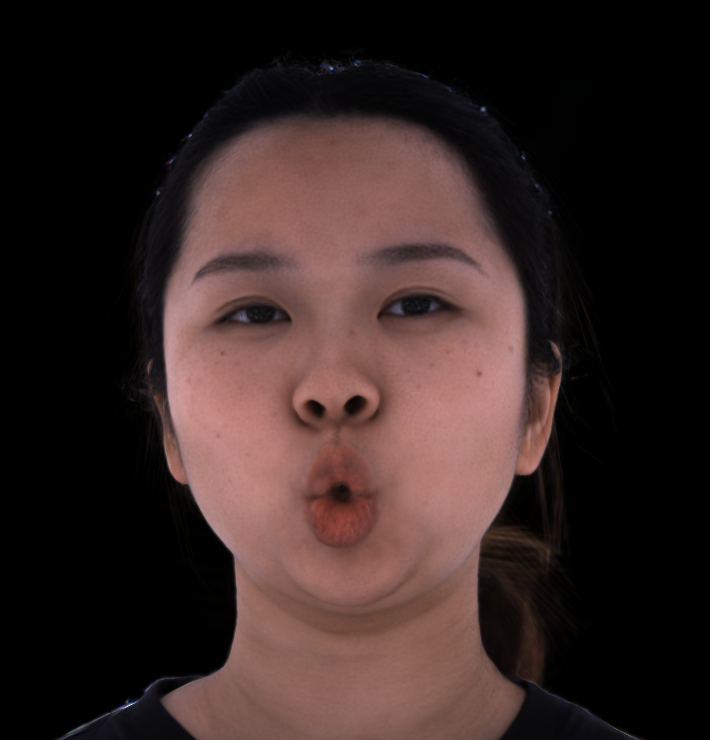}
            \subfloat{Ours}
        \end{minipage}
    \end{minipage}
    \caption{Ablation experiment about the expression consistency \yl{loss (ECL)}. By adding the expression consistency \yl{loss}, our model learns a more unified expression space of different identities.}
\label{fig:ablation_exp_cons}
\end{figure}
\begin{figure}
    \centering
    \begin{minipage}{\linewidth}
        \centering
        \begin{minipage}[t]{\onefourthfigurewidth\linewidth}
            \centering
            \includegraphics[width=\linewidth]{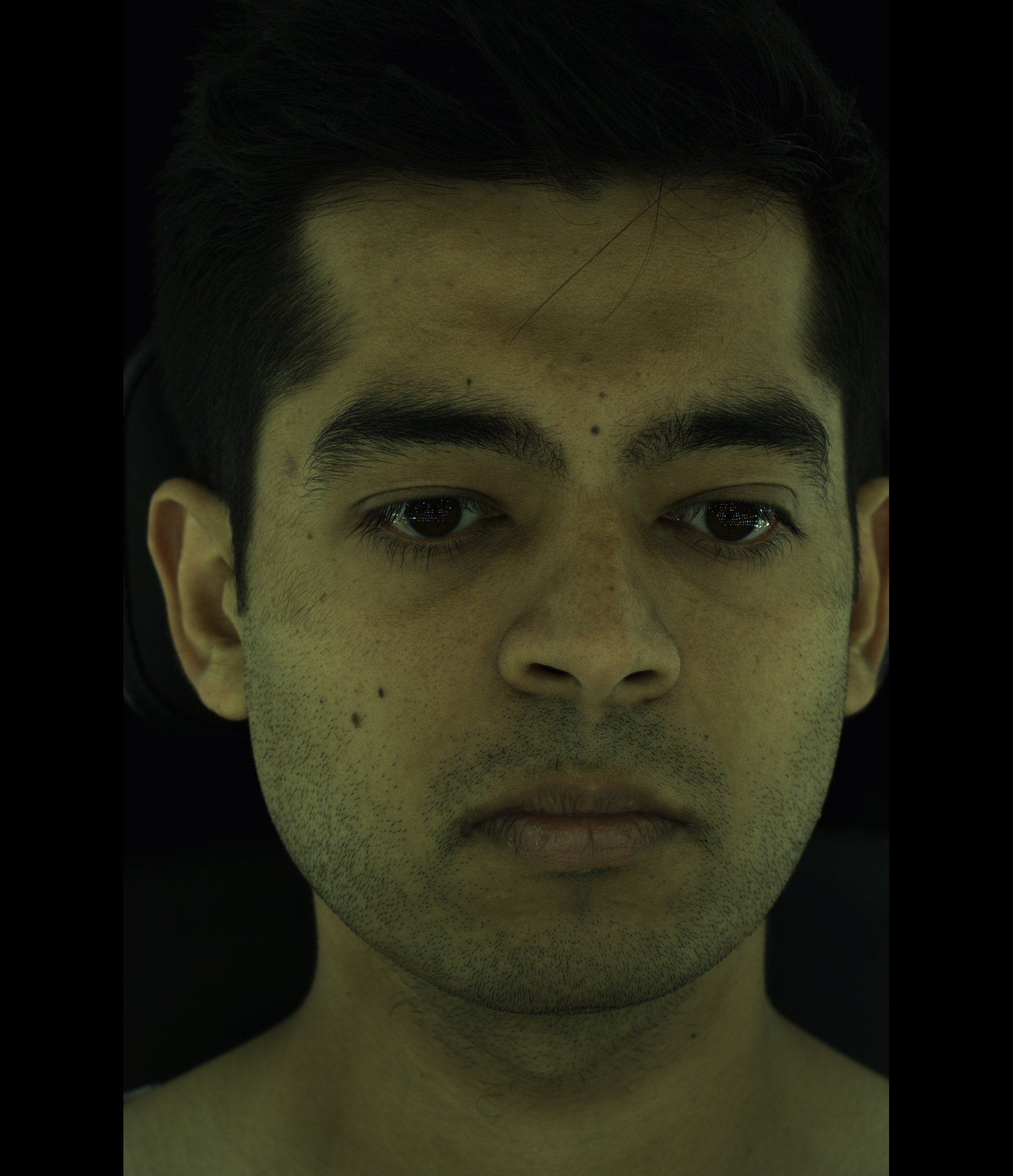}
            \subfloat{Source}
        \end{minipage}
        \begin{minipage}[t]{\onefourthfigurewidth\linewidth}
            \centering
            \includegraphics[width=\linewidth]{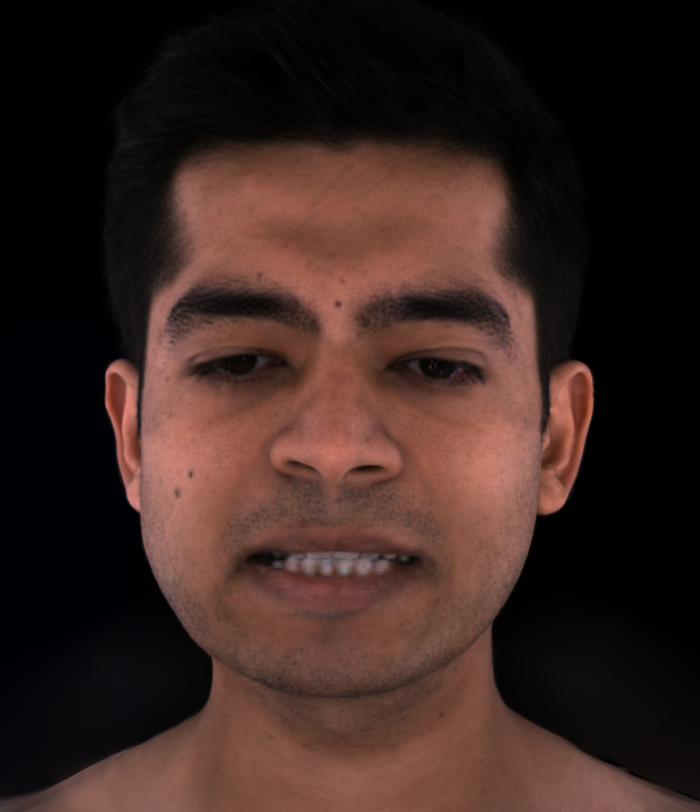}
            \subfloat{w/o LRL}
        \end{minipage}
        \begin{minipage}[t]{\onefourthfigurewidth\linewidth}
            \centering
            \includegraphics[width=\linewidth]{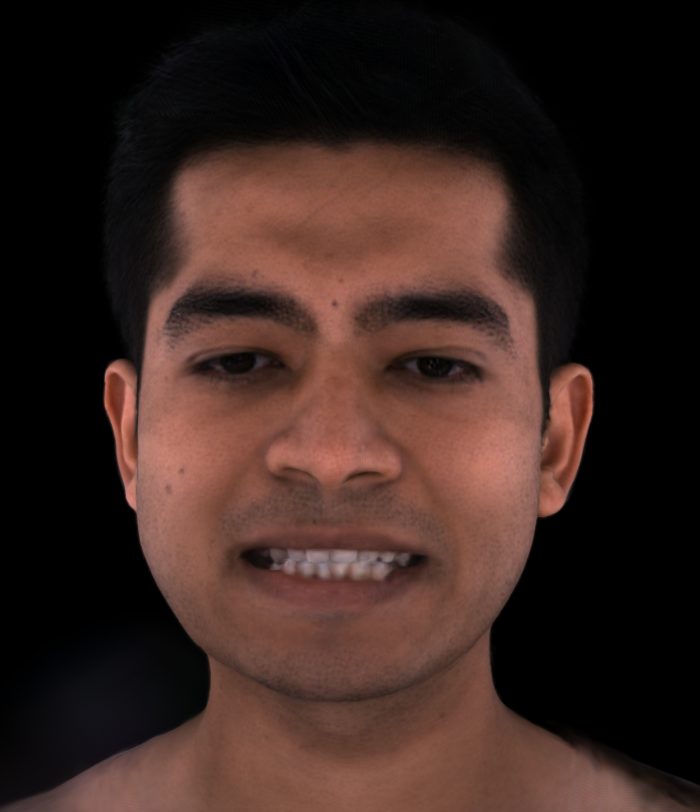}
            \subfloat{Ours}
        \end{minipage}
        \begin{minipage}[t]{\onefourthfigurewidth\linewidth}
            \centering
            \includegraphics[width=\linewidth]{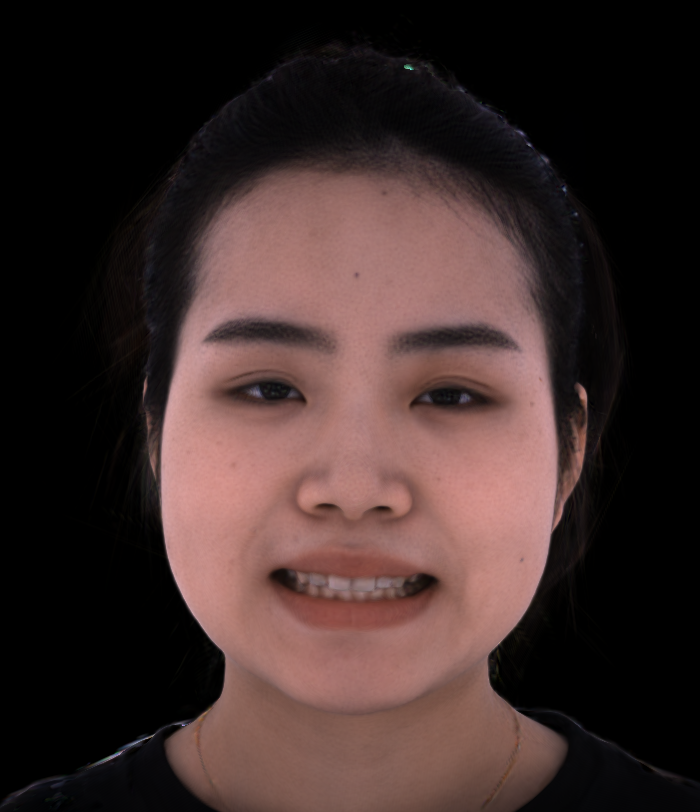}
            \subfloat{Target}
        \end{minipage}
    \end{minipage}
    \caption{Ablation study on the locality regularization loss. The locality regularization loss helps our model to keep the knowledge from the prior during fine-tuning, resulting in accurate expression transfer after personalization.}
\label{fig:ablation_ppl}
\end{figure}

\subsection{Qualitative Results}

\paragraph{Latent space evaluation}
Figure~\ref{fig:interpolation} shows interpolation between the identity \yl{codes} of different individuals while keeping expression code and lighting condition fixed. The relighting results are shown in Figure~\ref{fig:relighting}. The visualization shows that the identity, expression and illumination \yl{spaces} of our \sysname \yl{are} fully disentangled and can be freely combined, which demonstrates the effectiveness of our self-supervised training framework.

% \begin{table}[]
% \centering
% \caption{Quantitative evaluation results.}
% \begin{tabular}{{l|lll}}
% \hline
% Method   & MAE $\downarrow$   & SSIM $\uparrow$  & LPIPS $\downarrow$ \\ \hline
% MoFaNeRF~\cite{zhuang2022mofanerf} & 35.00 & 0.878 & 0.199 \\
% HeadNeRF~\cite{hong2022headnerf} & 14.07 & 0.905 & 0.163 \\
% Ours     & \textbf{4.45}  & \textbf{0.933} & \textbf{0.106} \\ \hline
% \end{tabular}
% \label{tab:cmp_fit_single}%
% \end{table}
\begin{table}[]
\centering
\caption{\revision{Quantitative evaluation results.}}
\resizebox{\columnwidth}{!}{%
\begin{tabular}{l|ccc|ccc}
\hline
         & \multicolumn{3}{c|}{Novel view synthesis} & \multicolumn{3}{c}{Single-view reconstruction} \\ \hline
Method   & MAE $\downarrow$         & SSIM $\uparrow$        & LPIPS $\downarrow$      & MAE $\downarrow$         & SSIM $\uparrow$       & LPIPS $\downarrow$      \\ \hline
MoFaNeRF & 28.3         & 0.829        & 0.247       & 35.00        & 0.878       & 0.199       \\
HeadNeRF & 17.19        & 0.892        & 0.211       & 14.07        & 0.905       & 0.163       \\
Ours     & \textbf{5.29}         & \textbf{0.950}        & \textbf{0.120}       & \textbf{4.45}         & \textbf{0.933}       & \textbf{0.106}       \\ \hline
\end{tabular}
\label{tab:cmp_fit_single}%
}
\end{table}

\paragraph{Avatar personalization}
Our model can be utilized to create personalized avatars from few-shot captures. Figure~\ref{fig:personalization} shows that \sysname generates high-fidelity relightable and animatable avatars by fitting multi-view images from various sources. The first two rows are from the Multiface Dataset~\cite{wuu2022multiface}, which are collected in a multi-view capture system with 40 and 146 cameras, respectively. We use the images of one frame as input. The middle two rows are from \cite{yang2023towards} with 23 input views. The last two rows are captured in our studio. The personalization is performed given only three images. Please refer to our accompanying video for the corresponding animations results.

\subsection{Comparisons}
% We compare our method with existing publicly available parametric models~\cite{hong2022headnerf,zhuang2022mofanerf} by fitting to a single in-the-wild image from FFHQ dataset~\cite{karras2019style} for volumetric head reconstruction. The qualitative results are shown in Figure~\ref{fig:cmp_fit_single}, while the average quantitative fitting error on 100 images is reported in Table~\ref{tab:cmp_fit_single}. The quantitative numbers are calculated only in the shared regions of different methods. Our method can achieve high-quality reconstruction from even a single in-the-wild image, while other baselines produce blurred results.

\revision{
We compare our method with existing publicly available parametric head models, \ie, HeadNeRF~\cite{hong2022headnerf} and MoFaNeRF~\cite{zhuang2022mofanerf}.

\paragraph{Novel view synthesis.}
We conduct both qualitative and quantitative evaluations on the task of novel view synthesis using the Multiface Dataset~\cite{wuu2022multiface}.
The parametric models are fitted to multi-view images from a single frame of each subject, and performance is measured against five held out views.
The visual comparison is provided in Figure~\ref{fig:cmp_viewinterp}, with detailed quantitative results presented in Table~\ref{tab:cmp_fit_single}.

\paragraph{Single-view reconstruction.}
Figure~\ref{fig:cmp_fit_single} shows qualitative comparison results for single-view head reconstruction. The average quantitative fitting error on 100 in-the-wild images from the FFHQ dataset~\cite{karras2019style} is reported in Table~\ref{tab:cmp_fit_single}.
%The numbers are calculated only in the shared regions of different methods.
These calculations are confined to the regions shared by different methods. 
Our method can achieve high-quality reconstruction results from a single in-the-wild image, while other baselines produce less accurate results.
}%revision

\subsection{Ablation Study}

\paragraph{Expression consistency constraint.}
We conduct \yl{an} ablation study to evaluate the effectiveness of the expression consistency loss \yl{(ECL)}. The experimental model is trained without expression consistency loss while \yl{keeping} other parts identical. Then we set the expression code as the one corresponding to the target image to evaluate the expression consistency across identities. The results in Figure~\ref{fig:ablation_exp_cons} \yl{show that our method learns more consistent expressions for different identities compared with the model w/o ECL. }

\paragraph{Locality regularization loss.} We attempt to remove our proposed locality regularization loss \yl{(LRL)}. The results \yl{are} shown in Figure~\ref{fig:ablation_ppl}. We perform personalization using 39 images of a single frame from the Multiface Dataset. The personalized avatar is animated by the expression code corresponding to the target expression. The fine-tuning \yl{deteriorates} the prior model without the locality regularization loss, which results in inaccurate animation results.

\section{Conclusion}
We present \sysname, the first volumetric morphable head model that enables continuous control over expression, identity, and global illumination. Through \yl{an} elaborately designed training framework, \sysname is capable of disentangling complex attributes from multi-view sequences of casually varied expressions and lightings in a self-supervised manner. Once trained, \sysname can serve as a powerful prior for various reconstruction tasks. Combined with the novel fitting technique we propose, \sysname requires only minimal observations to accurately fit a specific portrait and generate an \yl{animatable and relightable} avatar with real-time rendering.  We believe this work will have a profound impact on the development of the field.

\begin{acks}
We thank Xiaoqiang Liu for being our capture subject, the authors of \cite{zhuang2022mofanerf} and \cite{hong2022headnerf} for releasing their source code and pretrained models, as well as anonymous reviewers for their insightful feedback.
\end{acks}

\ifthenelse{\equal{\arxiv}{1}}
{
\appendix
\section{Implementation Details}

We provide the values of hyperparameters in Table~\ref{tab:hyperparameter}. The detailed neural network architecture is shown in Figure~\ref{fig:networks}. The network is initially trained without the expression consistency loss $\expressionloss$ for efficiency. Subsequently, $\expressionloss$ is incorporated and the network continues training until it converges.

\begin{figure}
    \centering
    \includegraphics[width=1.0\linewidth]{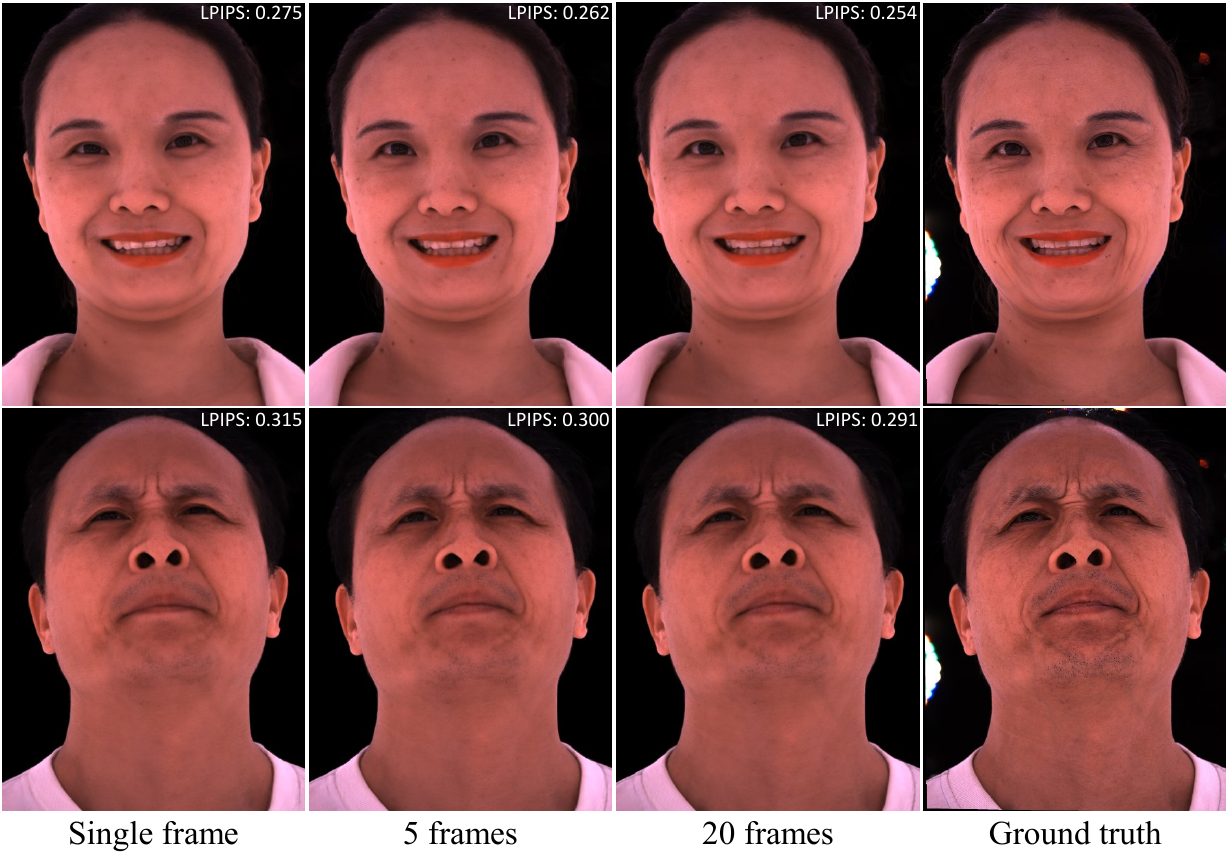}
    \caption{The assessment of animation quality is depicted in the left three columns, illustrating the animated results of the personalized avatar derived from varying numbers of frames. The accuracy of expression amplitude improves, and additional identity-specific wrinkles emerge as the number of reference images increases. Quantitative metrics for each animation are displayed at the top-right corner of the corresponding image.}
\label{fig:cmp_animation}
\end{figure}

\begin{figure*}
\centering
    \includegraphics[width=\linewidth]{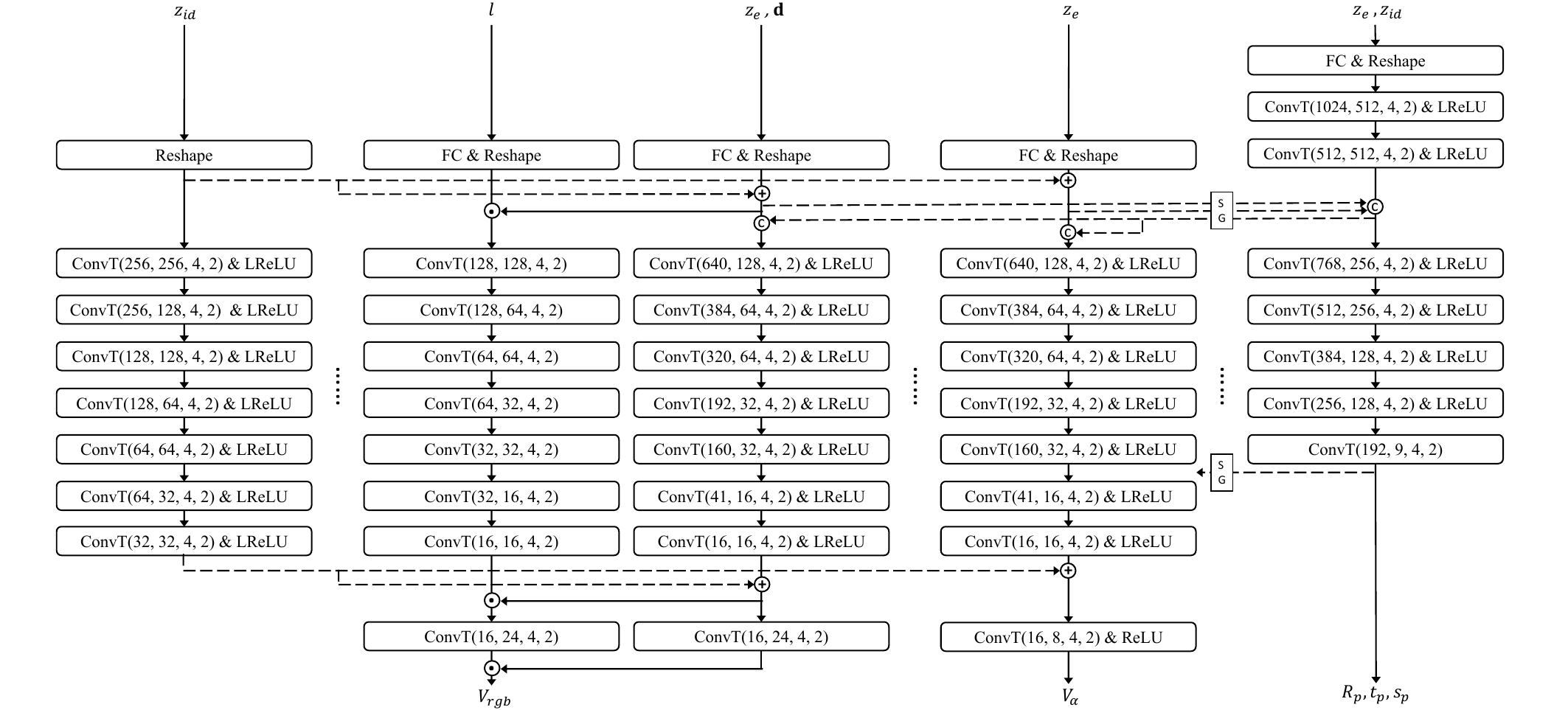}
    %\vspace{-0.05in}
    \caption{Detailed neural network architecture of \sysname. The transposed convolutional layer is represented as $ConvT(input\ channels, output\ channels, kernel size, stride)$. The inter-connections between intermediate layers are omitted to avoid clutter. The expression encoder $\expressionencoder$, transformation encoder $\transformencoder$, and the mesh decoder $\meshdecoder$ are identical to that in \cite{yang2023towards}.}
\label{fig:networks}
\end{figure*}
%\section{Networks Architectures}

\begin{table}[]
\centering
\caption{Values of our hyperparameters.}
\begin{tabular}{cr|cr|cr}
\hline
Parameter & Value & Parameter & Value & Parameter & Value \\ \hline
$N_{l}$     & 356  & $N_{prim}$     & 16384 & ${\lambda}_{VGG}$         & 0.1     \\
${\lambda}_{KLD}$    & 0.01   & ${\lambda}_{vol}$    & 0.01 & ${\lambda}_{scale}$    & 0.01  \\
${\lambda}_{id}$     & 0.0001    & ${\lambda}_{imgexp}$    & 0.2  & ${\lambda}_{parexp}$    & 0.02 \\ 
${\lambda}_{exp}$    & 0.01    & ${\lambda}_{LR}$    & 0.1   \\ \hline
\end{tabular}
\label{tab:hyperparameter}%
\end{table}

\begin{figure}
    \centering
    \includegraphics[width=1.0\linewidth]{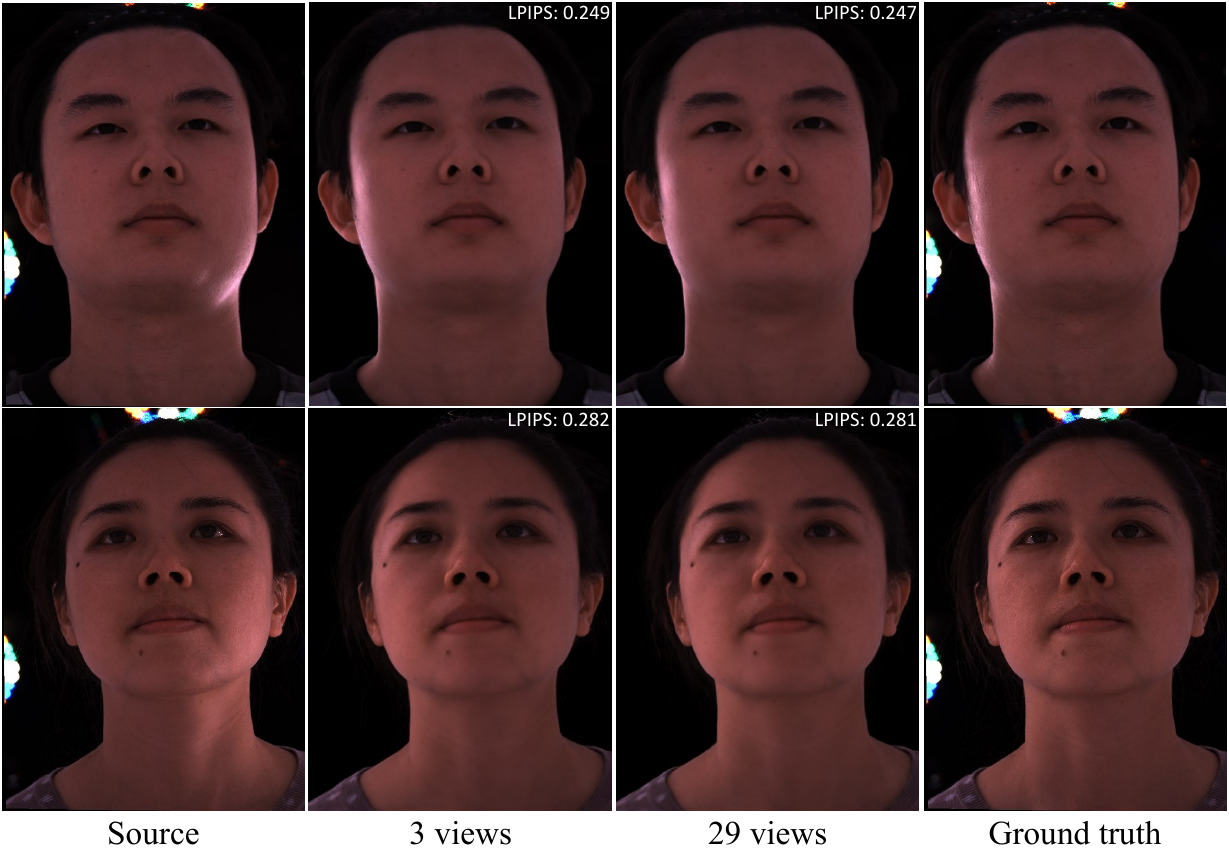}
    \caption{The evaluation of relighting performance is showcased, where \sysname is adapted to different numbers of views captured under the original source lighting, as displayed in the left column. The model demonstrates the capability to achieve accurate relighting outcomes even with a minimal set of views.}
\label{fig:cmp_relighting}
\end{figure}

\begin{figure}
    \centering
    \begin{minipage}[t]{\onefourthfigurewidth\linewidth}
        \centering
        \includegraphics[width=\linewidth]{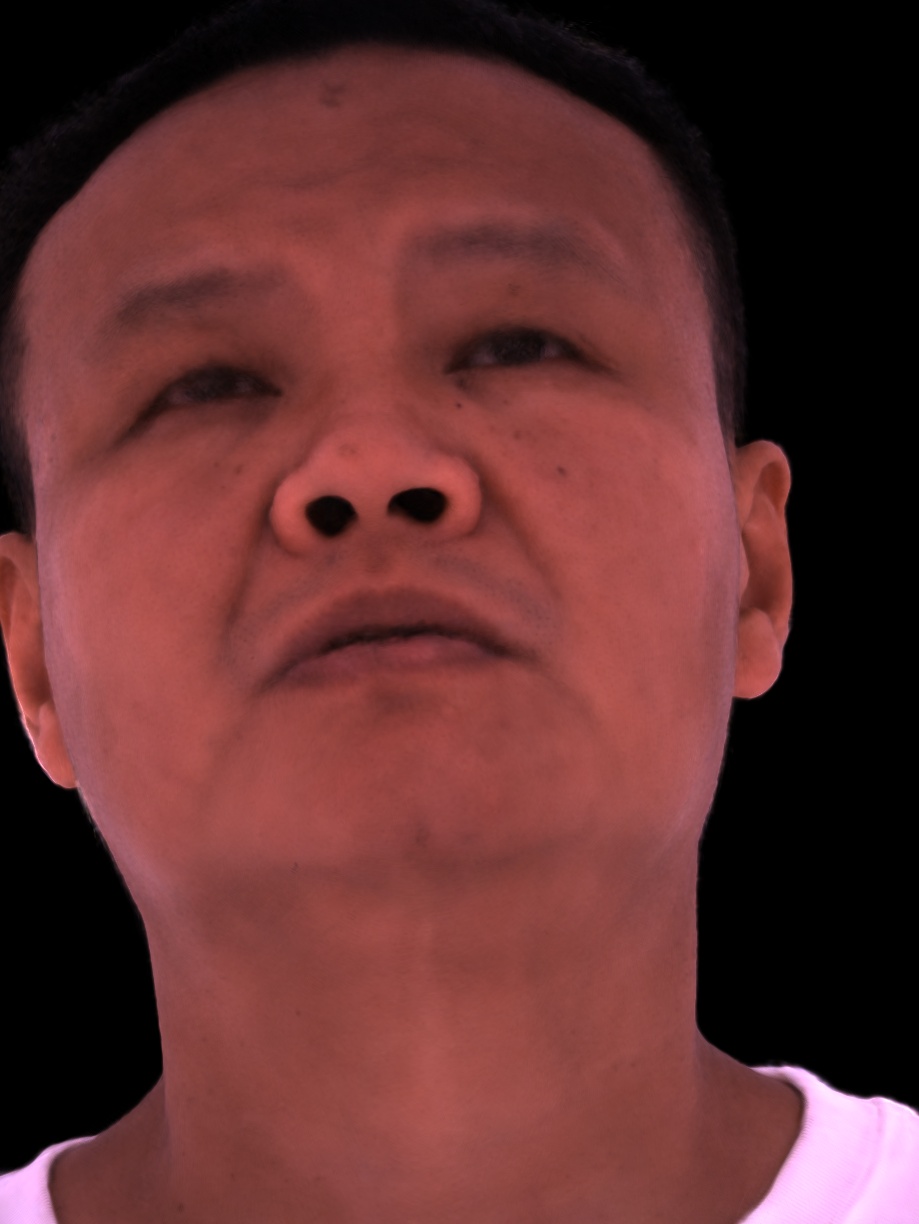}
        \includegraphics[width=\linewidth]{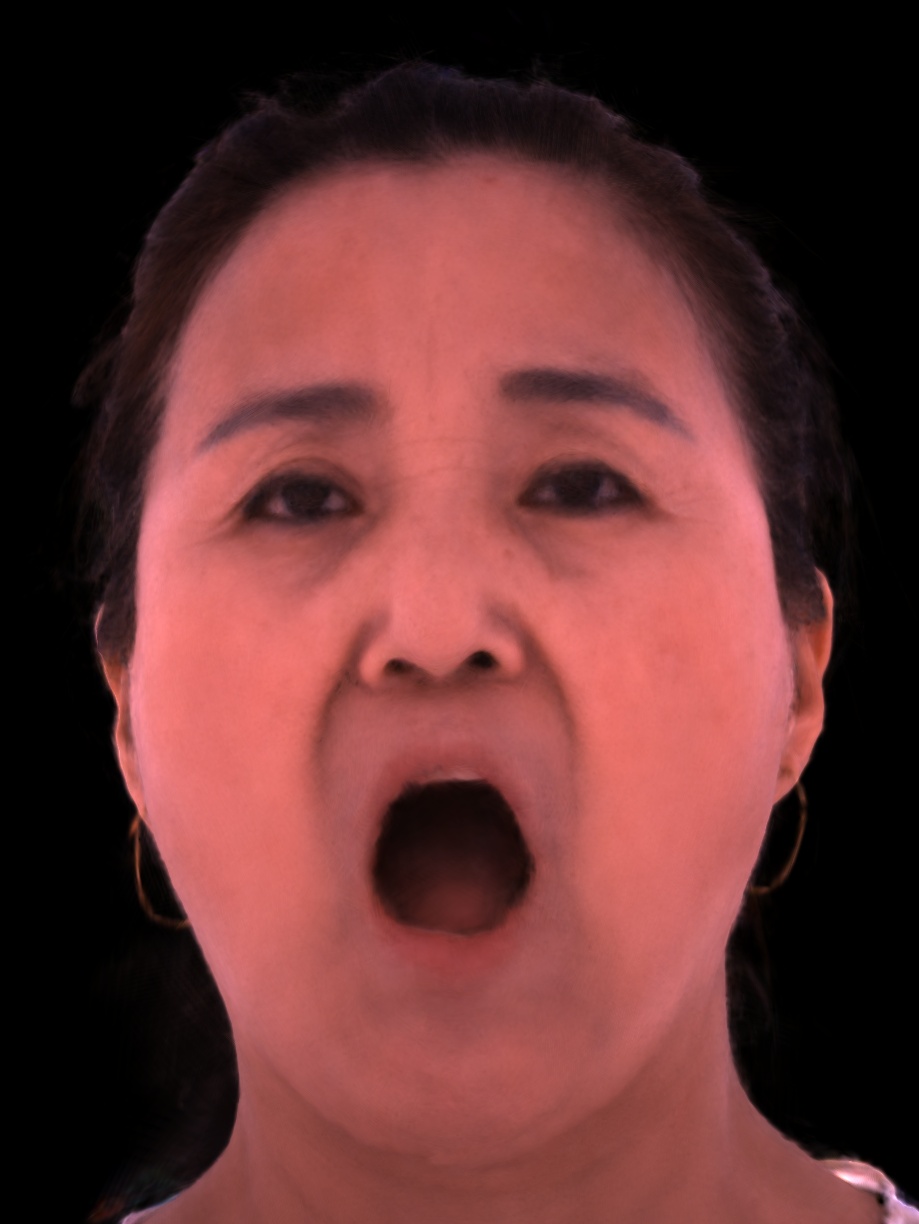}
        \subfloat{w/o connection}
    \end{minipage}
    \begin{minipage}[t]{\onefourthfigurewidth\linewidth}
        \centering
        \includegraphics[width=\linewidth]{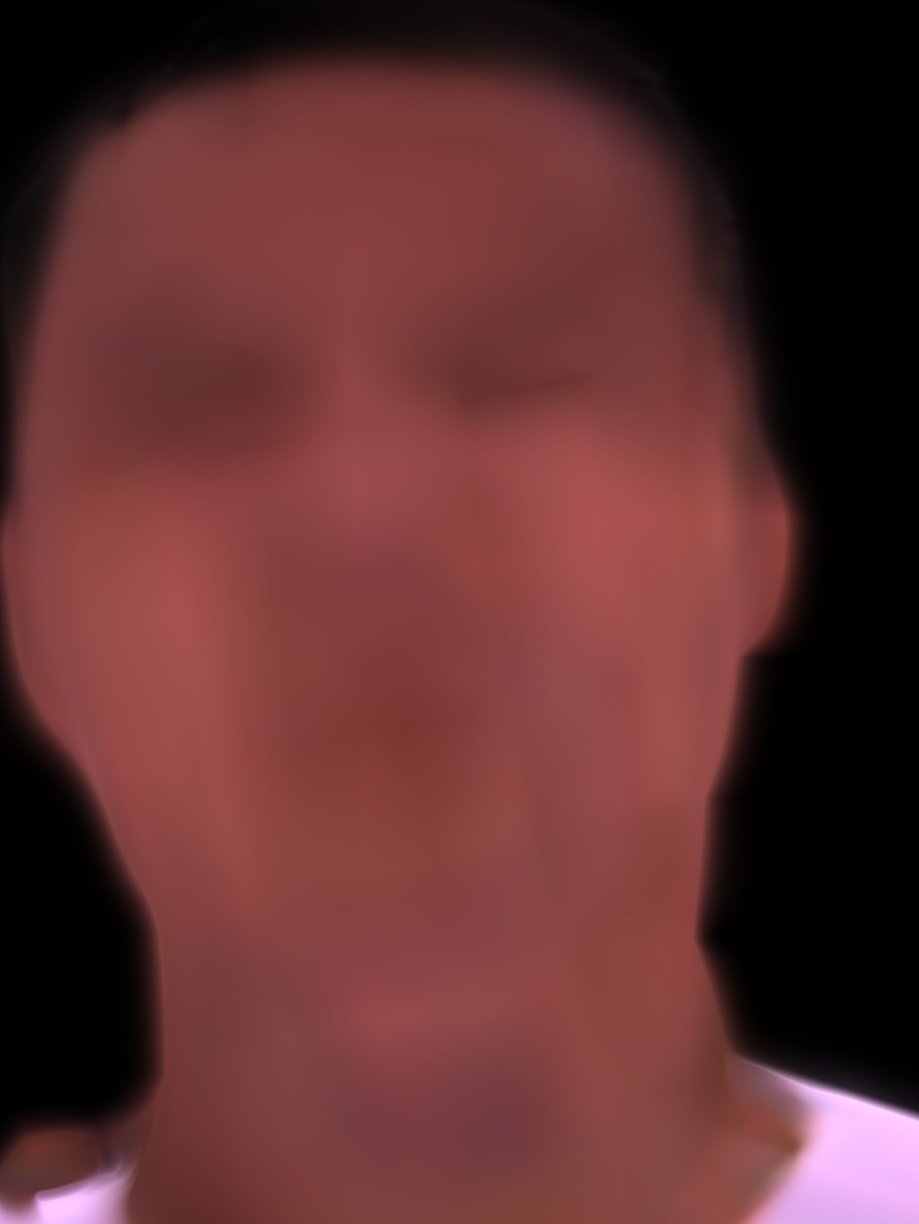}
        \includegraphics[width=\linewidth]{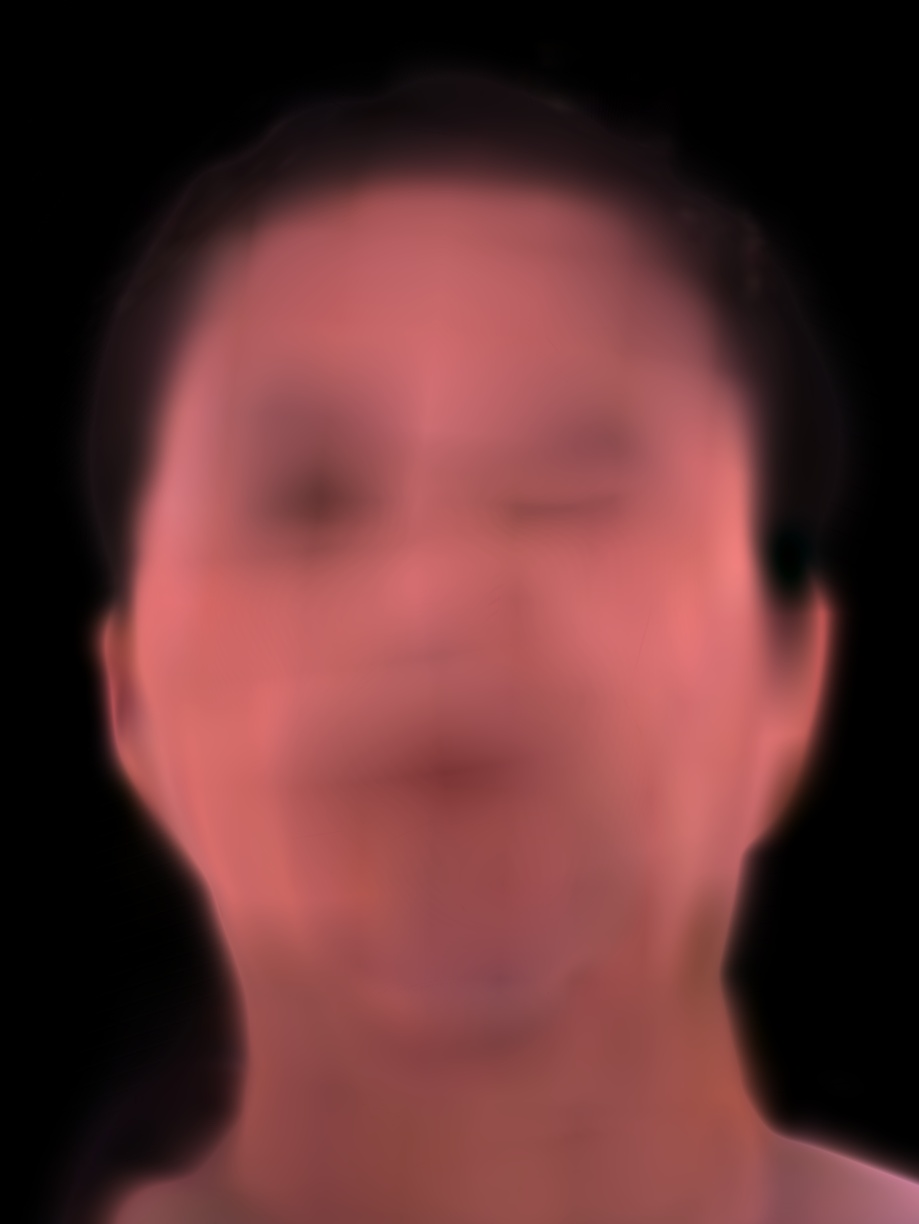}
        \subfloat{w/o SG}
    \end{minipage}
    \begin{minipage}[t]{\onefourthfigurewidth\linewidth}
        \centering
        \includegraphics[width=\linewidth]{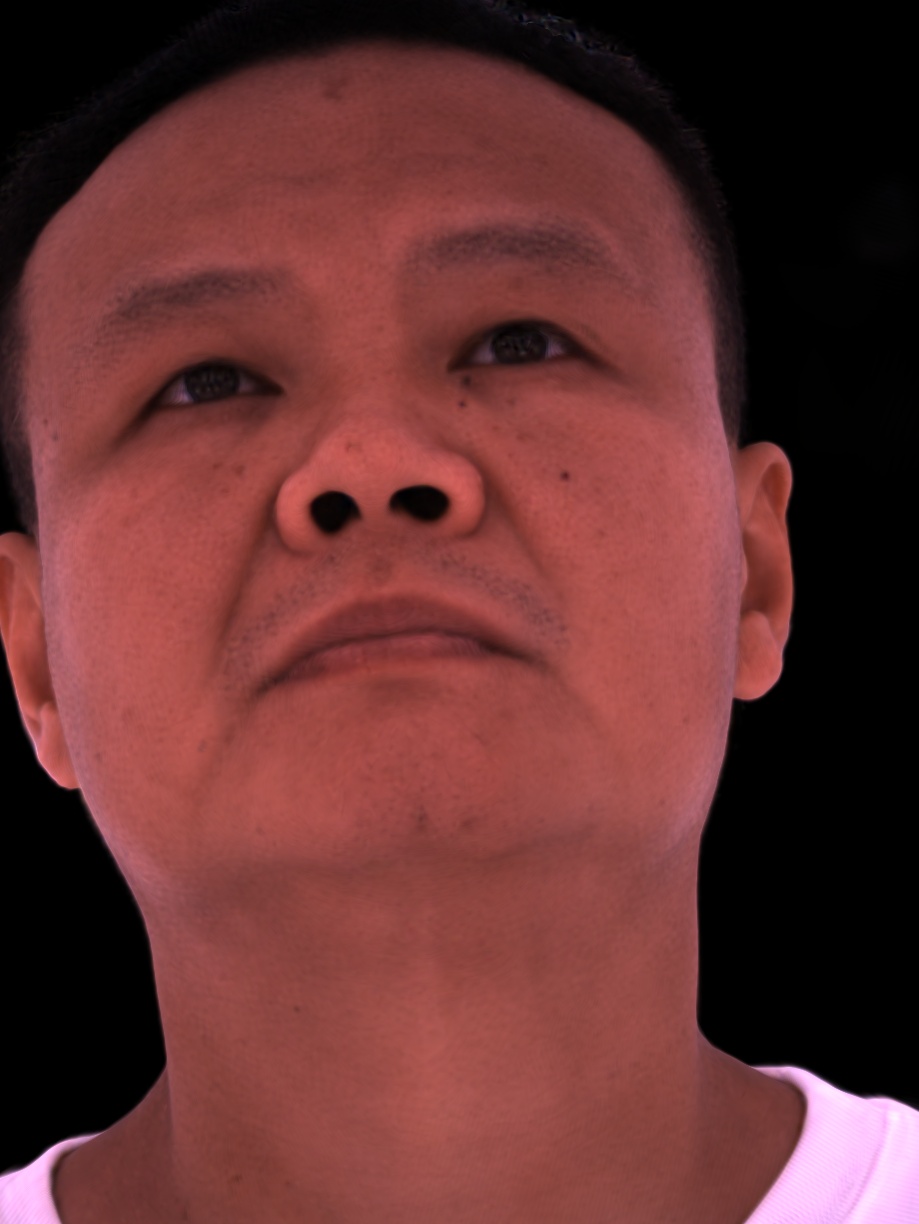}
        \includegraphics[width=\linewidth]{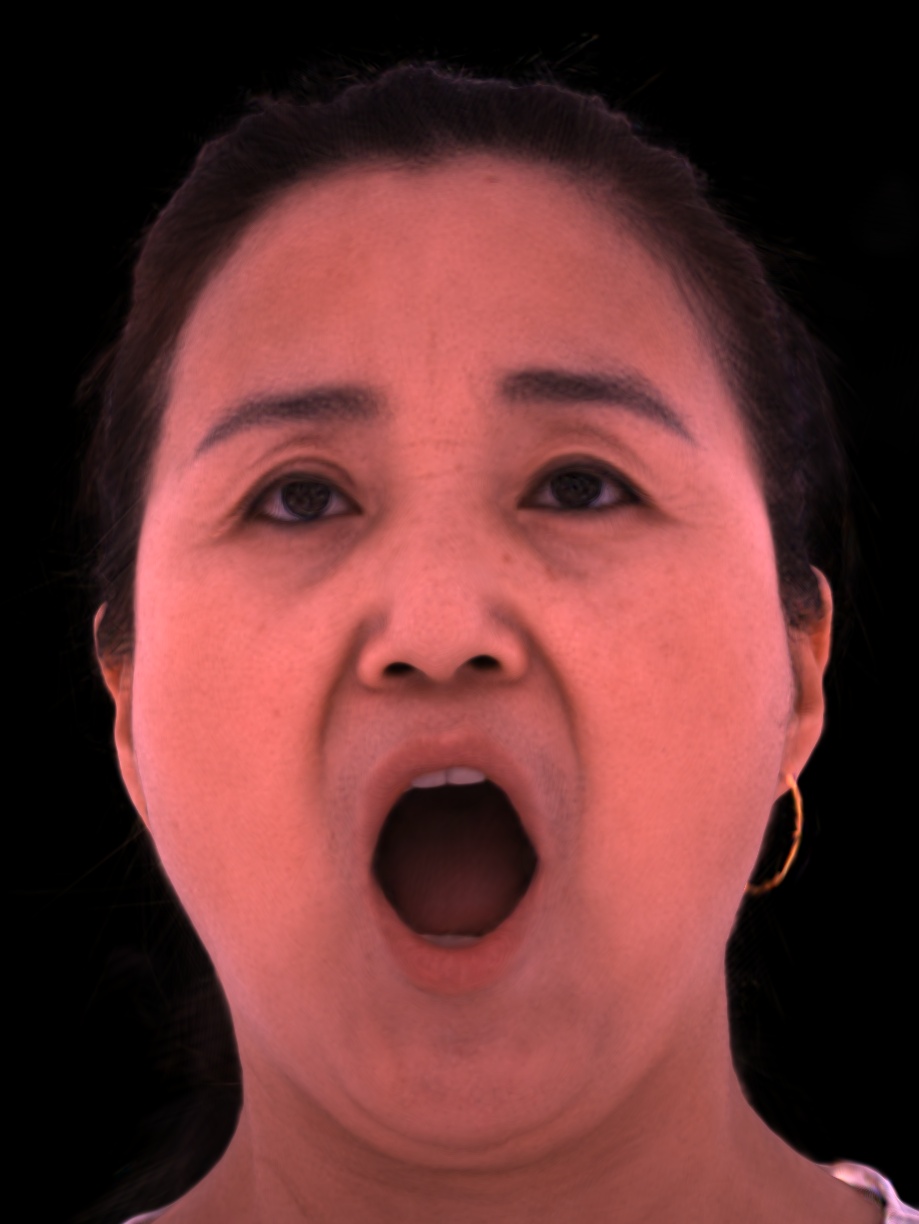}
        \subfloat{Ours}
    \end{minipage}
    \begin{minipage}[t]{\onefourthfigurewidth\linewidth}
        \centering
        \includegraphics[width=\linewidth]{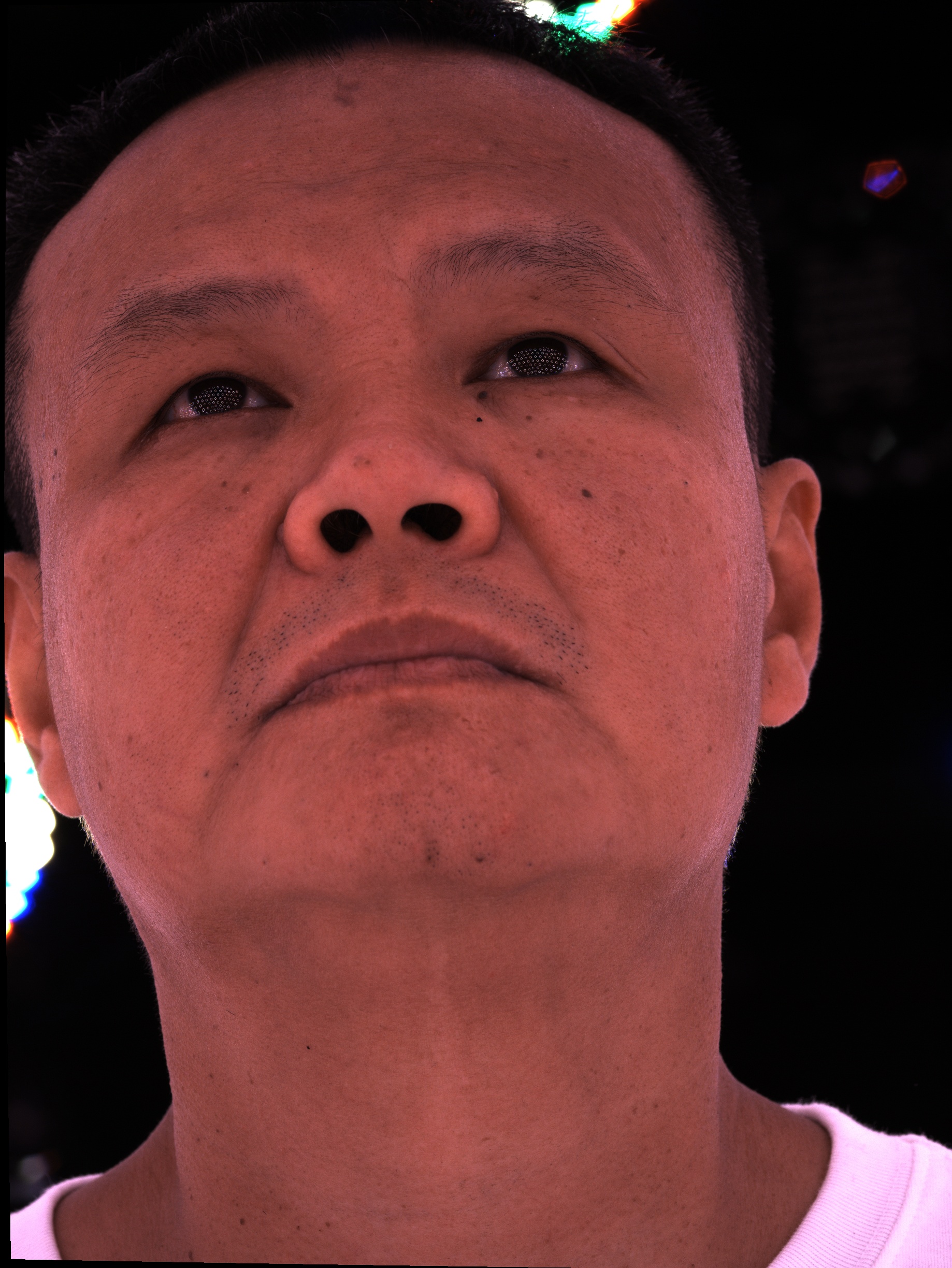}
        \includegraphics[width=\linewidth]{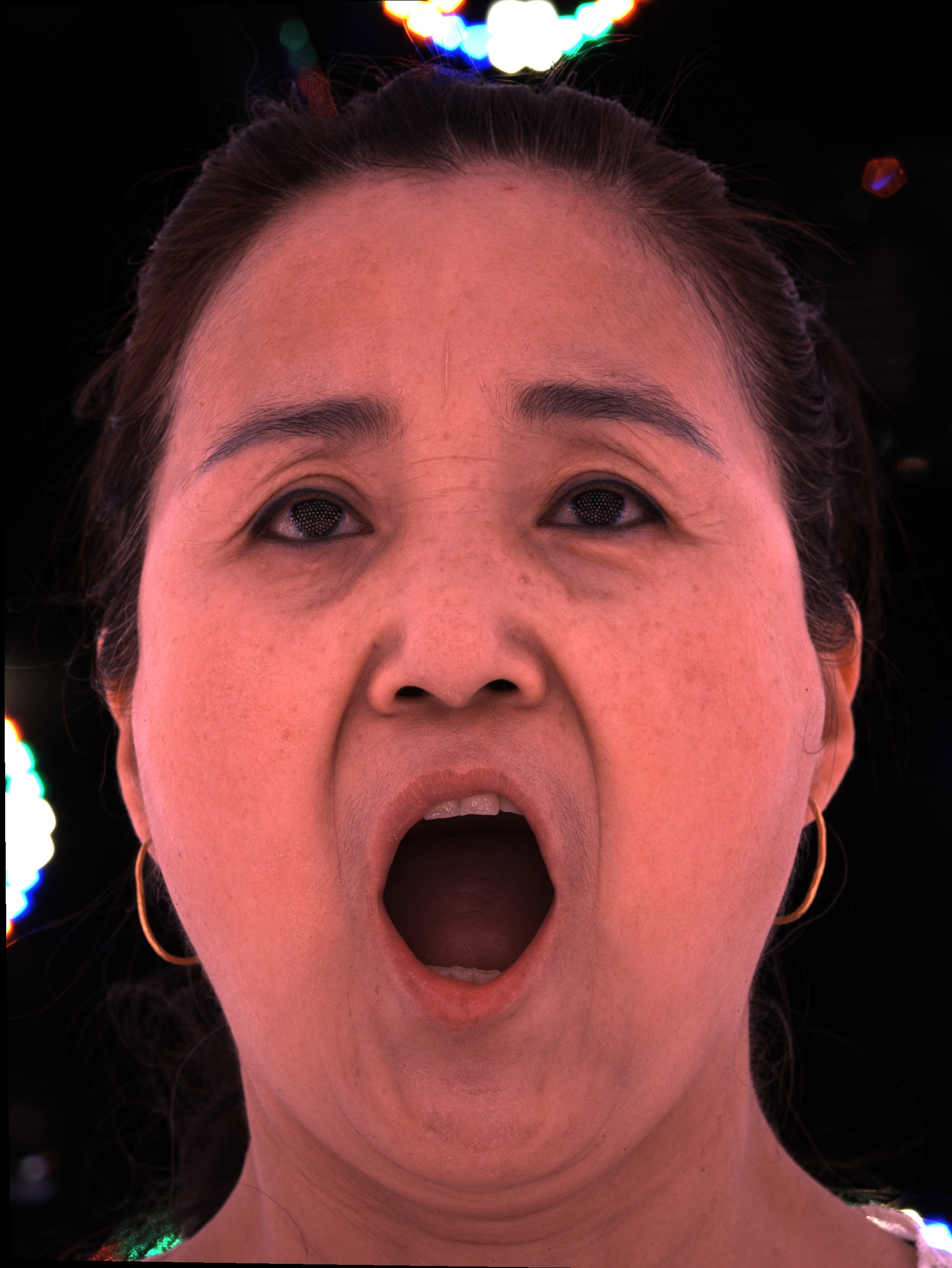}
        \subfloat{Ground truth}
    \end{minipage}
    %\caption{The trained hybrid mesh-volumetric avatar enjoys nearly consistent topology.}
    \caption{Ablation study on the detach-concatenation operation. The rendered images get blurred as the number of identities increases without the inter-connection. Moreover, without detaching, the training process fails to converge. The detach-concatenation leads to clearer and more accurate renderings in the final output.}
\label{fig:ablation_detach_cat}
\end{figure}

\begin{figure}
    \centering
    \begin{minipage}[t]{\onethirdfigurewidth\linewidth}
        \centering
        \includegraphics[width=\linewidth]{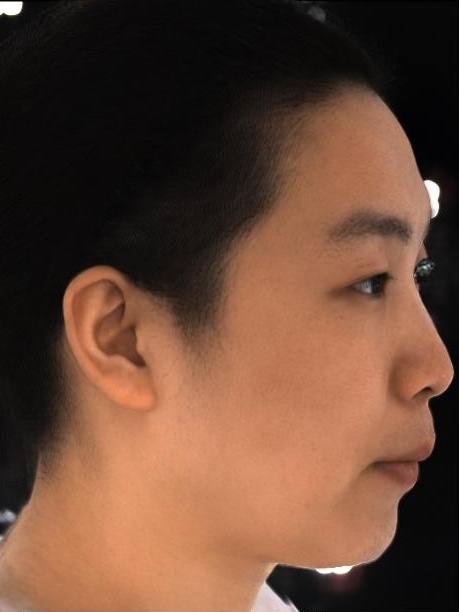}
        \includegraphics[width=\linewidth]{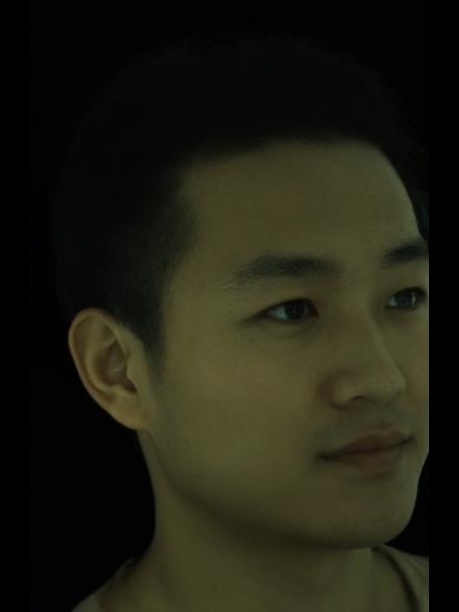}
        \subfloat{w/o fine-tuning}
    \end{minipage}
    \begin{minipage}[t]{\onethirdfigurewidth\linewidth}
        \centering
        \includegraphics[width=\linewidth]{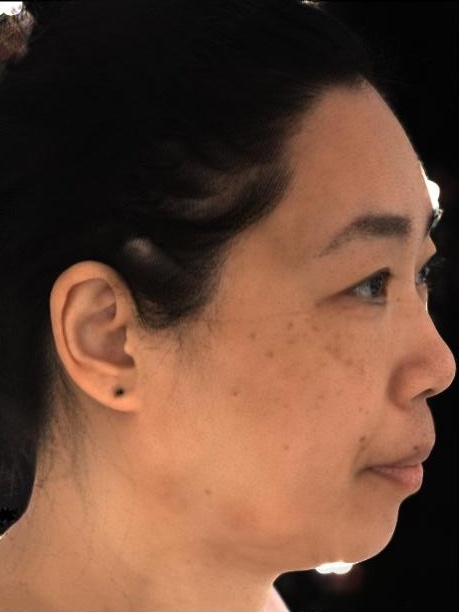}
        \includegraphics[width=\linewidth]{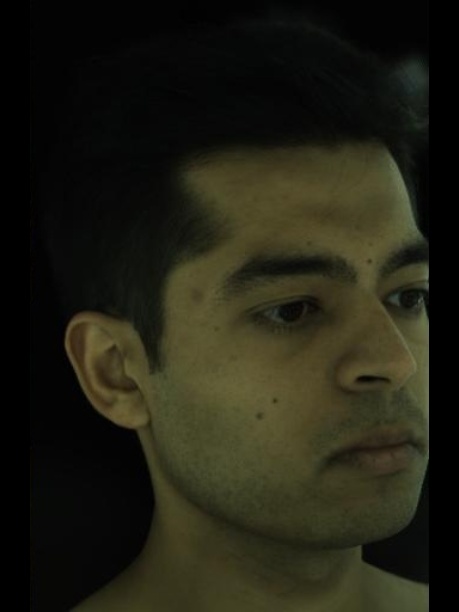}
        \subfloat{Ours}
    \end{minipage}
    \begin{minipage}[t]{\onethirdfigurewidth\linewidth}
        \centering
        \includegraphics[width=\linewidth]{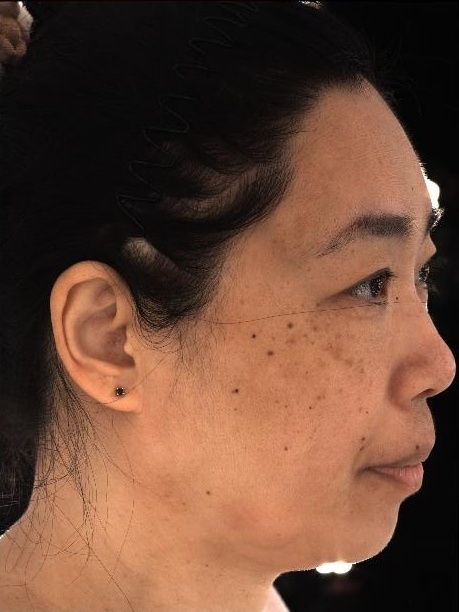}
        \includegraphics[width=\linewidth]{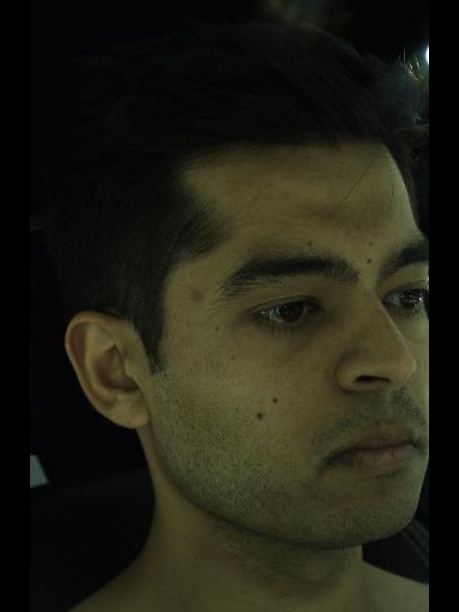}
        \subfloat{Ground truth}
    \end{minipage}
    \caption{The ablation study highlights that fine-tuning dramatically improves the quality of reconstructions, especially for identities that deviate from the initial training set.}
\label{fig:ablation_finetune}
\end{figure}

\begin{figure}
    \centering
    \begin{minipage}{\linewidth}
        \centering
        \begin{minipage}[t]{\onethirdfigurewidth\linewidth}
            \centering
            \includegraphics[width=\linewidth]{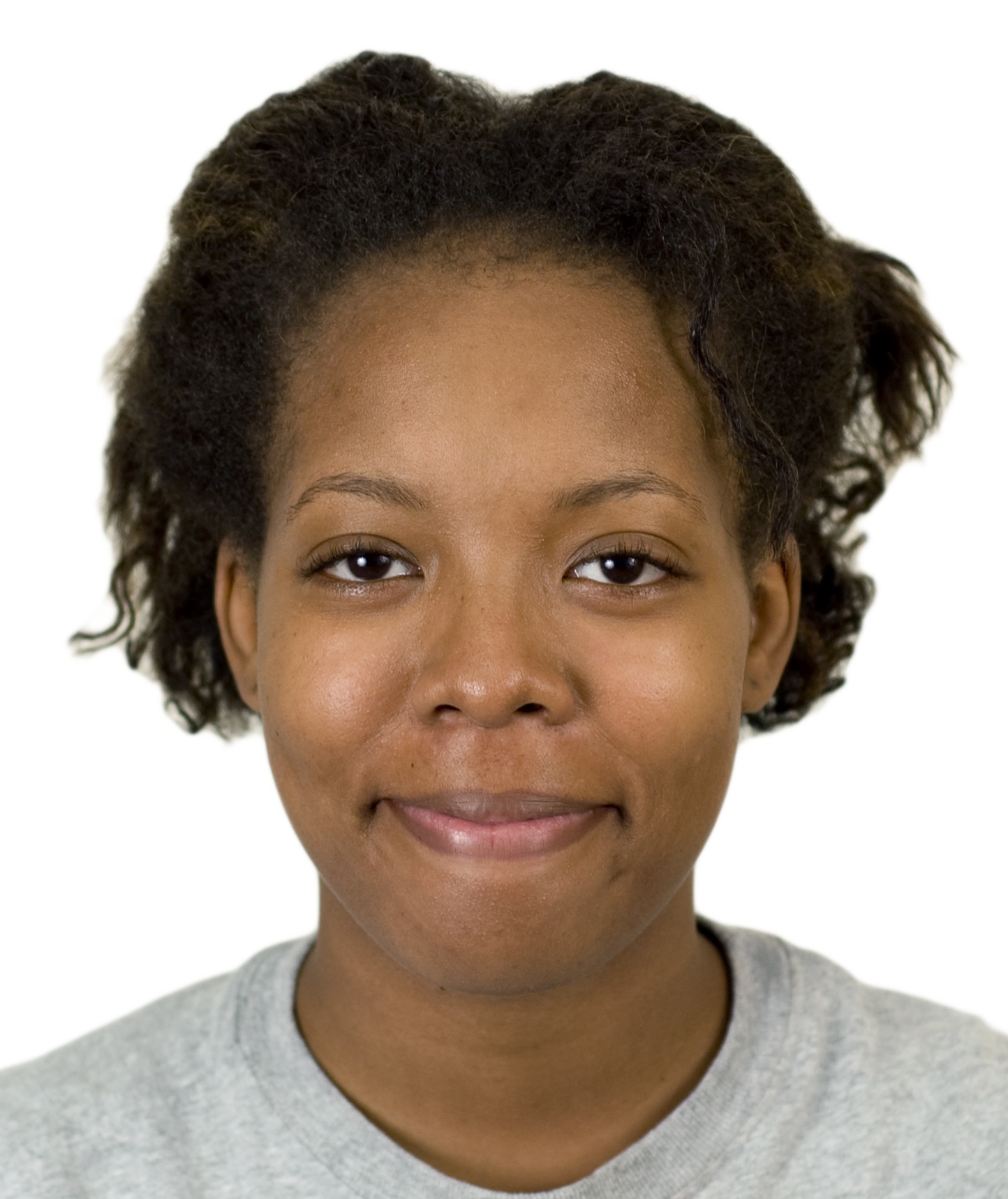}
            \subfloat{Input}
        \end{minipage}
        \begin{minipage}[t]{\onethirdfigurewidth\linewidth}
            \centering
            \includegraphics[width=\linewidth]{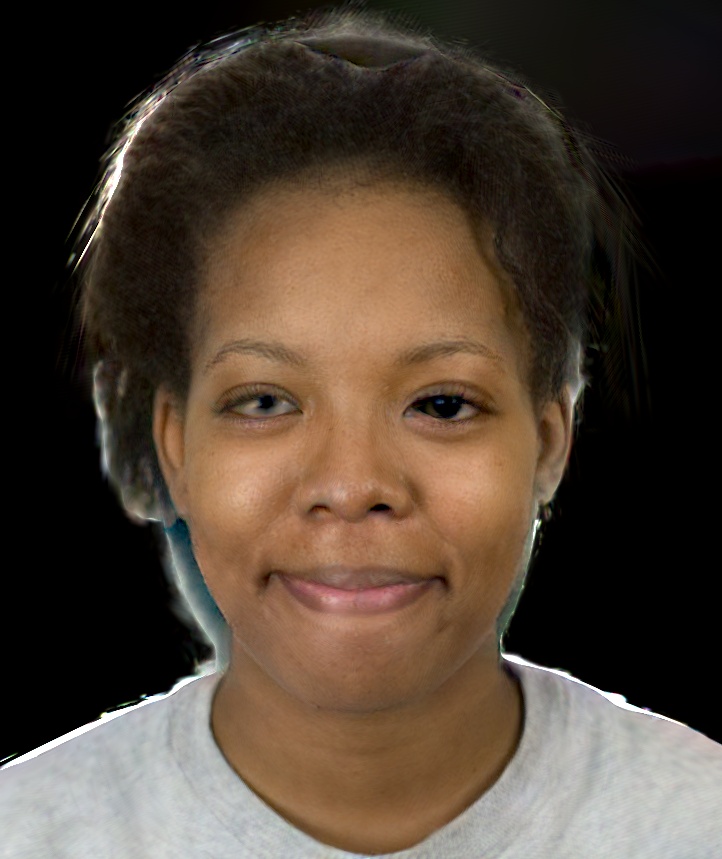}
            \subfloat{Reconstruction}
        \end{minipage}
        \begin{minipage}[t]{\onethirdfigurewidth\linewidth}
            \centering
            \includegraphics[width=\linewidth]{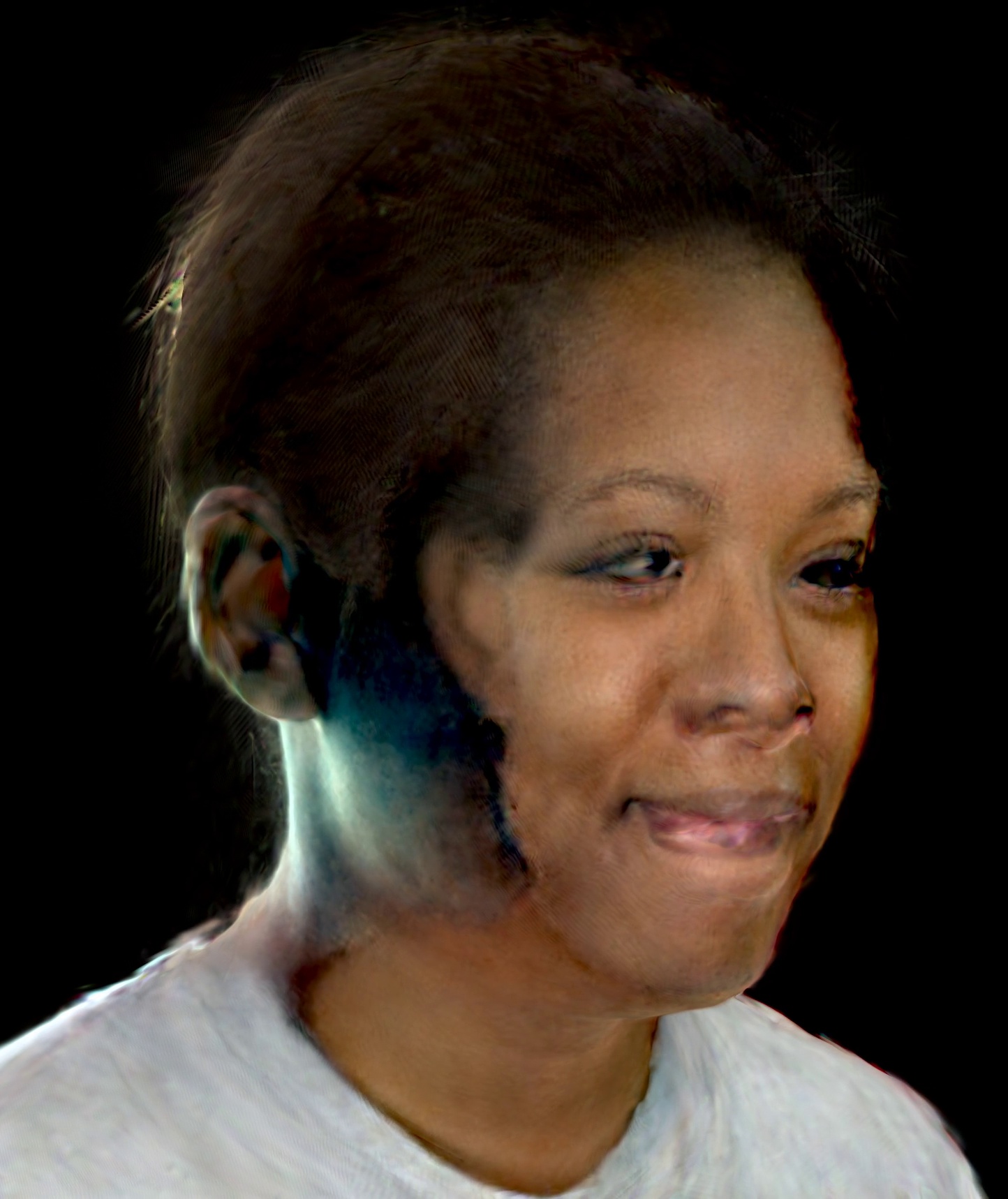}
            \subfloat{Left view}
        \end{minipage}
    \end{minipage}
    \caption{Failure case of single image reconstruction. Optimization trapped in local minimum, resulting from an inaccurate initial alignment of the image. The input image is from the Chicago Face Database~\cite{ma2015chicago}.}
\label{fig:failure}
\end{figure}

\section{Additional Experiments}

\subsection{Evaluation of Personalization}

\paragraph{Animation.} We evaluate the animation quality of the few-shot personalized avatar with held-out identities from the captured dataset. The fitting is performed on different numbers of frames in varying expressions and other frames are used to drive the avatar. As shown in Figure~\ref{fig:cmp_animation}, the animation resembles to the ground truth expression, and personal idiosyncrasies are more distinct when the number of reference expressions increases.

\paragraph{Relighting.} The relighting results of the personalized avatar from different numbers of views are compared to the ground truth images captured under known illumination. Figure~\ref{fig:cmp_relighting} shows the results. Our \sysname achieves accurate relighting from as few as three view inputs (frontal, left, right). The relighting accuracy is slightly improved with more input views, which demonstrates the robustness of our personalization framework.

\subsection{Additional Ablations}

\paragraph{Detach-concatenation operation.}
We conduct experiments to evaluate the effectiveness of the detach-concatenation operation.  We ablate two alternative choices: 
\begin{enumerate}%[leftmargin=*]
    \item w/o connection: We remove the inter-connection between the transformation decoder $\transformdecoder$ and other two decoders.
    \item w/o SG: The stop gradient operation is removed and the inter-connection falls back to vanilla feature concatenation.
\end{enumerate}
The comparison is shown in Figure~\ref{fig:ablation_detach_cat}. Our approach yields the best reconstruction results.

\paragraph{Fine-tuning stage.}
We show the rendering results before and after the fine-tuning in Figure~\ref{fig:ablation_finetune}. The fine-tuning stage significantly improves the faithfulness of the personalized avatar.

\section{Limitations}

Although our results are encouraging, future work will need to address certain limitations. 

First, the capabilities of our model are constrained by the training data; for instance, it struggles to model effectively when the input includes long hair or obstructions. Future work could explore strategies for fine-tuning on a large amount of in-the-wild data to improve the generalization ability. 

Similar to traditional 3DMMs, our optimization-based reconstruction method is relatively sensitive to initial values, especially for single-image reconstruction. A failure case is shown in Figure~\ref{fig:failure}. Besides, there is still blurring and an identity shift when animating and relighting a novel identity from limited observations, which can be alleviated by adding more reference images. Subsequent work could explore better personalization methods. 

Additionally, as we adopt the lighting model of \cite{yang2023towards}, our method cannot achieve near-field and high-frequency relighting. Lastly, the current strategy of fine-tuning for each individual may also limit the widespread application of our technology.
}
{
}

% Bibliography
\bibliographystyle{ACM-Reference-Format}
\bibliography{bib_ra}

\newpage

% figures-only pages
\ifthenelse{\equal{\arxiv}{0}}
{

}
{
}

\end{document}